%% file: main.tex
\documentclass[acmtog]{acmart}
\acmSubmissionID{1917}

\input{preamble}

\acmJournal{TOG}
\acmVolume{44}
\acmNumber{6}
\acmArticle{236}
\acmYear{2025}
\acmMonth{12}

\setcopyright{acmlicensed}

\acmDOI{10.1145/3763341}

\begin{document}
\title{MALeR: Improving Compositional Fidelity in Layout-Guided Generation}

\author{Shivank Saxena}
\orcid{0009-0009-4964-2454}
\affiliation{%
  \institution{CVIT, IIIT Hyderabad}
  \country{India}}
\email{shivank.saxena@research.iiit.ac.in}

\author{Dhruv Srivastava}
\orcid{0000-0002-6789-4390}
\affiliation{%
  \institution{CVIT, IIIT Hyderabad and Adobe Research}
  \country{India}}
\email{dsrivastava@adbobe.com}

\author{Makarand Tapaswi}
\orcid{}
\affiliation{%
  \institution{CVIT, IIIT Hyderabad}
  \country{India}}
\email{makarand.tapaswi@iiit.ac.in}

\renewcommand\shortauthors{Saxena, S. et al.}

\begin{abstract}

\input{sec/0_abstract}
\end{abstract}

%

%
%


\begin{teaserfigure}
\input{figures/teaser}

\end{teaserfigure}

\maketitle

\input{sec/1_intro}
\input{sec/2_relatedwork}
\input{sec/3_method}

\input{sec/4_experiments}

\input{sec/5_conclusion}
\input{sec/6_ack}

\bibliographystyle{ACM-Reference-Format}
\bibliography{bibliography/longstrings, bibliography/main}

\clearpage

\input{sec/7_figurepages}

\end{document}

%% file: preamble.tex
\usepackage{graphicx}
\usepackage{amsmath}
\usepackage{mathtools}
\usepackage{booktabs}
\usepackage{tabularx}
\usepackage{rotating}
\usepackage{multirow}
\usepackage{booktabs}
\usepackage{lipsum}
\usepackage{enumitem}
\usepackage{pifont}
\usepackage{soul}
\usepackage{mdframed}
\usepackage{siunitx}
\usepackage{balance}
\usepackage[font=small]{caption}
\usepackage{subcaption}
\usepackage{wrapfig}

\usepackage[utf8]{inputenc} 
\usepackage[T1]{fontenc}    
\usepackage{hyperref}       
\usepackage{url}            
\usepackage{amsfonts}       
\usepackage{nicefrac}       
\usepackage{microtype}      
\usepackage[dvipsnames, table]{xcolor}
\usepackage[most]{tcolorbox}
\usepackage[capitalize]{cleveref}
\usepackage{arydshln}

\newcommand{\modelshort}{MALeR}
\newcommand{\modellong}{Masked Attribute-aware Latent Regularization}

\newcommand{\cmark}{\ding{51}} 

\renewcommand{\paragraph}[1]{\vspace{1mm}\textit{#1}}

\newcommand{\bx}{\mathbf{x}}
\newcommand{\by}{\mathbf{y}}
\newcommand{\bz}{\mathbf{z}}

\newcommand{\bI}{\mathbf{I}}

\newcommand{\MA}{\mathcal{A}}
\newcommand{\MN}{\mathcal{N}}

\newcommand{\MD}{\mathcal{D}}

\newcommand{\MS}{\mathcal{S}}

\newcommand{\ME}{\mathcal{E}}

\newcommand{\MB}{\mathcal{B}}
\newcommand{\ML}{\mathcal{L}}

\newcommand{\real}[1]{\mathbb{R}^{#1}}

\newcommand{\Liou}{\ML_\text{iou}}
\newcommand{\Lmask}{\ML_\text{mask}}
\newcommand{\lambdamask}{\lambda_\text{mask}}
\newcommand{\Lkl}{\ML_\text{KL}}
\newcommand{\lambdakl}{\lambda_\text{KL}}
\newcommand{\Latt}{\ML_\text{att}}
\newcommand{\lambdaatt}{\lambda_\text{att}}
\newcommand{\Lsim}{\ML_\text{sim}}
\newcommand{\lambdasim}{\lambda_\text{sim}}
\newcommand{\Ldis}{\ML_\text{dis}}
\newcommand{\lambdadis}{\lambda_\text{dis}}

\usepackage[ruled]{algorithm2e} 

\SetAlFnt{\small}
\SetAlCapFnt{\small}
\SetAlCapNameFnt{\small}
\SetAlCapHSkip{0pt}

\usepackage{xspace}
\makeatletter
\DeclareRobustCommand\onedot{\futurelet\@let@token\@onedot}
\def\@onedot{\ifx\@let@token.\else.\null\fi\xspace}

\def\eg{\emph{e.g}\onedot}

 \def\vs{\emph{vs}\onedot}

\makeatother

\crefname{figure}{Fig.}{Figs.}
\Crefname{figure}{Figure}{Figures}
\crefname{section}{Sec.}{Secs.}
\Crefname{section}{Section}{Sections}
\crefname{table}{Tab.}{Tabs.}
\Crefname{table}{Table}{Tables}


%% file: sec/0_abstract.tex
Recent advances in text-to-image models have enabled a new era of creative and controllable image generation. 
However, generating compositional scenes with multiple subjects and attributes remains a significant challenge.
To enhance user control over subject placement, several layout-guided methods have been proposed.
However, these methods face numerous challenges, particularly in compositional scenes.
Unintended subjects often appear outside the layouts, 
generated images can be out-of-distribution and contain unnatural artifacts, or 
attributes bleed across subjects, leading to incorrect visual outputs.
In this work, we propose \modelshort{}, a method that addresses each of these challenges.
Given a text prompt and corresponding layouts, our method prevents subjects from appearing outside the given layouts while being in-distribution.
Additionally, we propose a masked, attribute-aware binding mechanism that prevents attribute leakage, enabling accurate rendering of subjects with multiple attributes, even in complex compositional scenes.
Qualitative and quantitative evaluation demonstrates that our method achieves superior performance in compositional accuracy, generation consistency, and attribute binding compared to previous work.
\modelshort{} is particularly adept at generating images of scenes with multiple subjects and multiple attributes per subject.
Project page: \url{https://katha-ai.github.io/projects/maler/}.

%% file: figures/teaser.tex
\centering
{\small A studio photo of a
\textbf{\color{Gray}white marble lion}, a
\textbf{\color{Goldenrod!90!black}yellow gold swan},
and a
\textbf{black obsidian cat}.} \\
\includegraphics[width=0.195\linewidth]{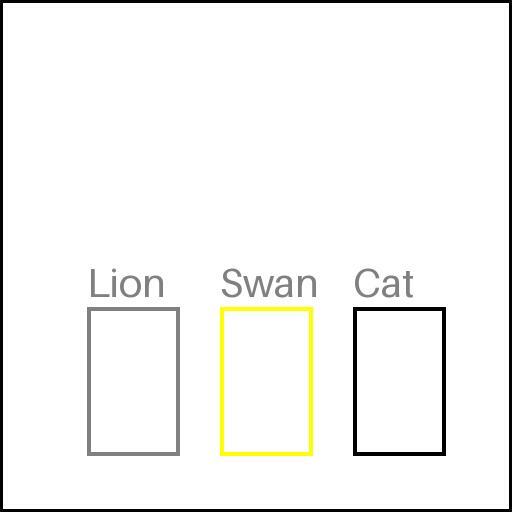} \hfill
\includegraphics[width=0.195\linewidth]{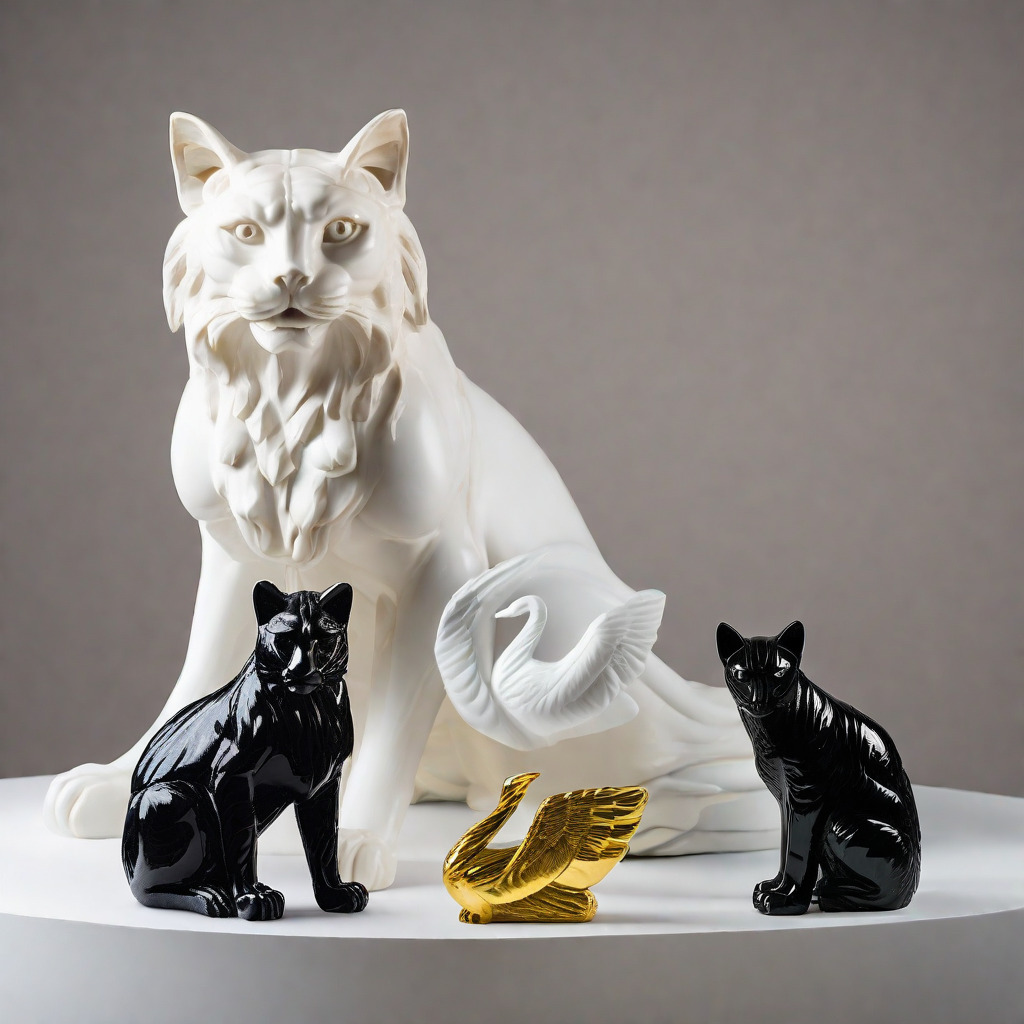} \hfill
\includegraphics[width=0.195\linewidth]{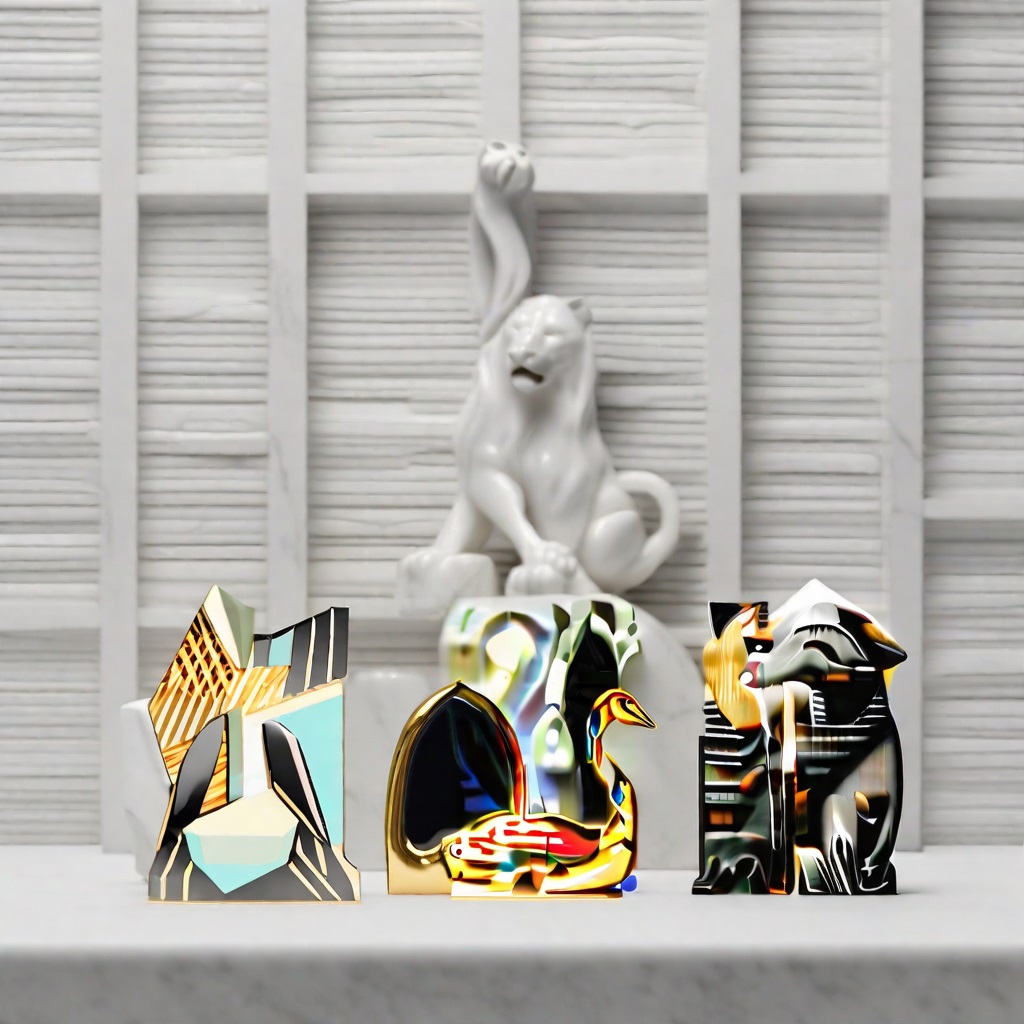} \hfill
\includegraphics[width=0.195\linewidth]{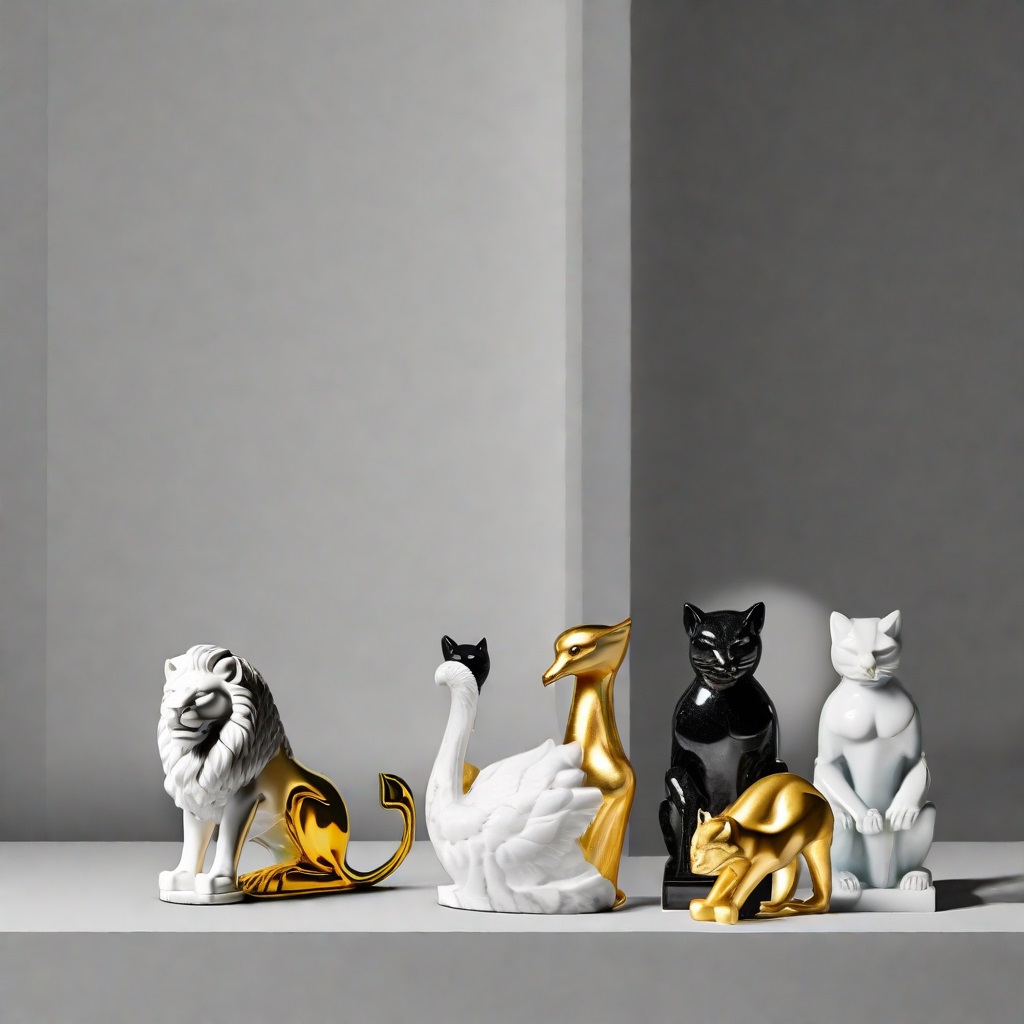} \hfill
\includegraphics[width=0.195\linewidth]{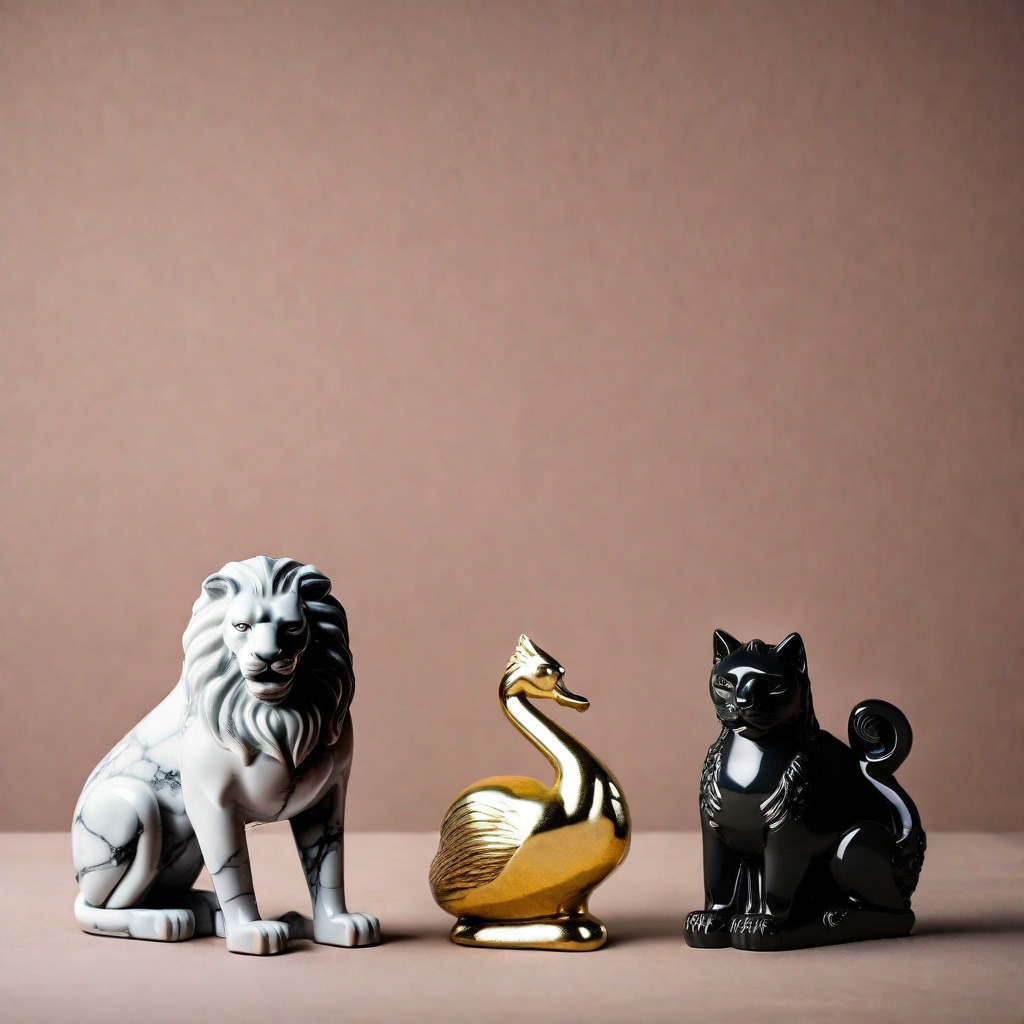}

\begin{tabularx}{\linewidth}{*{5}{>{\centering\arraybackslash}X}}
\centering
\small{Layout guidance} &
\small{(a)~Background semantic leakage} &
\small{(b)~Out-of-distribution} &
\small{(c)~Incorrect attribute binding} &
\small{(d)~\textbf{\modelshort{}} (Ours)} \\
\end{tabularx}
\vspace{-4mm}
\caption{
We present typical challenges of modern layout-guided text-to-image generation methods.
From left-to-right, we first present the layout-guidance prompt.
Next, we present some challenges:
(a)~\textit{Background semantic leakage} where additional subjects appear outside the intended region;
(b)~\textit{Out-of-distribution image generation} with cracked images and/or erroneous textures; and
(c)~\textit{Incorrect attribute binding} with too many subjects.
(d)~\modelshort{}, our approach, is able to solve these challenges and generate an accurate multi-subject multi-attribute image.
}

\Description[Set of images that present typical challenges on modern layout-guided text-to-image generation methods and an image from our approach that solves these issues.]{
We present typical challenges of modern layout-guided text-to-image generation methods.
From left-to-right, we first present the layout-guidance prompt.
Next, we present some challenges:
(a)~\textit{Background semantic leakage} where additional subjects appear outside the intended region;
(b)~\textit{Out-of-distribution generation} with erroneous animal textures; and
(c)~\textit{Incorrect attribute binding} with too many subjects.
(d)~\modelshort{}, our approach, is able to solve these challenges and generate an accurate multi-subject multi-attribute image.
}
\label{fig:teaser}

%% file: sec/1_intro.tex
\vspace{-2mm}
\section{Introduction}
\label{sec:intro}

Diffusion models have achieved remarkable success in generating high quality and realistic images from text prompts~\cite{dhariwal2021diffusion, esser2021taming, esser2024scaling, rombach2022sd, saharia2022imagen,  podell2024sdxl, ho2020ddpm, kingma2021variational, nichol2021improved, song2021score}.
These models often use techniques such as classifier-free guidance~\cite{ho2021cfg} for text-conditioned image generation. 
However, models struggle with complex text prompts involving multiple subjects and attributes~\cite{chefer2023attend, dahary2024yourself, rassin2023lingbind}.
Common failure modes include
catastrophic neglect (skipped subjects), 
creation of extra subjects (\eg~two dogs when asked to generate one), and
incorrect subject-attribute associations.

To address catastrophic neglect and improve prompt adherence, \citet{chefer2023attend} introduced a cross-attention-guided excitation mechanism that
optimizes the latent state of the diffusion model during inference.
They maximize the cross-attention between subject tokens and image patches, ensuring that each subject in the prompt exerts sufficient influence on the generated image.
Follow-up works (\eg~\cite{agarwal2023astar, li2023divide, guo2024initno})
adopt similar latent optimization paradigms,
often using variants of cross-attention-based losses.
However, the optimization process can drive the latents out-of-distribution resulting in incoherent generations.
Thus, generating multiple samples with different random initializations is common to achieve the desired scene composition.

To enhance user control, recent works condition text-to-image models through the use of visual layouts to indicate spatial location of subjects.
Grounding inputs such as
bounding boxes~\cite{li2023gligen},
depth maps~\cite{zhang2023adding, huang2023composer},
semantic maps, or scribbles~\cite{zhang2023adding, wang2024instancediffusion, lv2024place, huang2023composer},
improve control and guide the generation process.
While semantic and depth maps provide fine-grained spatial information, they are often impractical for users to manually construct or edit.
On the other hand, bounding boxes offer a simple and intuitive alternative to specify desired subject locations.

However, such layout-guided methods inherit the same fundamental limitations observed in text-to-image generation.
They are susceptible to distributional drift due to latent optimization, and additionally face the challenge of ensuring adherence to the provided layout.
While some methods attempt to mitigate these issues by manipulation of cross-attention maps during the sampling process~\cite{chefer2023attend, chen2024training, feng2023training},
recent approaches~\cite{phung2024grounded, dahary2024yourself} also incorporate self-attention maps into the guidance process.
To some extent, this helps, but creates new challenges.
For example, layout-guided methods suffer a problem of \textit{background semantic leakage}, where unintended (additional) subjects appear outside or near the specified bounding boxes (\cref{fig:teaser}a).
Moreover, the constraints also make the generations brittle resulting in \textit{out-of-distribution images} characterized by texture-less subjects, tiling or cracking, and other unnatural artifacts (\cref{fig:teaser}b).
Interestingly, while some random state initializations achieve close to desirable compositions, many lead to failures---necessitating tens of attempts to obtain a good image.

Additionally, crafting diverse compositional scenes requires precise attribute binding between subjects. 
Prior works~\cite{jiang2024pac, meral2024conform, li2023divide, rassin2023lingbind, feng2023training} associate attributes with their corresponding subjects by leveraging cross-attention-based losses or similarity measures.
However, these methods are limited to handling one or two subjects and often struggle in complex compositions resulting in
\textit{attribute leakage} across subjects and background regions, or
\textit{subject blending}, where individual identities become visually entangled (\cref{fig:teaser}c).
This highlights the need for a training-free layout-guided generation framework that not only enforces spatial alignment, but also supports robust attribute binding across multiple subjects---particularly in complex, multi-attribute scenes.

In this work, we propose \modellong{} (\modelshort{}, \textit{n. painter} in German) to address all three challenges.
We introduce a \textit{masked latent regularization} strategy to prevent background semantic leakage during latent optimization.
Specifically, we discourage the emergence of subject-like patterns outside the designated bounding boxes by anchoring background latents close to their original values.
Second, we perform \textit{in-distribution latent alignment} to prevent out-of-distribution images during latent optimization.
More concretely, during early denoising steps, we encourage the optimized latents to remain close to the prior Gaussian distribution through an alignment term based on KL-divergence.
Third, we propose a novel \textit{subject-attribute association} loss that encourages similarity (dissimilarity) between masked regions of cross-attention maps of paired (unpaired) nouns and adjectives.
This formulation not only enables accurate attribute binding across multiple subjects, but also supports \textit{multiple attributes} to be associated with each subject, enabling generation of rich and compositional scenes with precise subject-attribute association.

The main contributions of our work are summarized next.
(i)~We identify \textit{background semantic leakage} as a limitation of current layout-guided generation methods and propose \textit{masked latent regularization} as a way to address it.
(ii)~To prevent out-of-distribution artifacts, we regularize the latents through \textit{in-distribution latent alignment}.
(iii)~We introduce a novel \textit{subject-attribute association loss} to ensure correct binding of \textit{multiple attributes} in compositional generation.
(iv)~Thorough experiments are presented using the same random seeds.
We qualitatively show that \modelshort{} succeeds at generating images for difficult prompts containing multiple attributes.
We also present quantitative comparisons against previous works on DrawBench~\cite{saharia2022imagen} and HRS~\cite{bakr2023hrs} benchmarks and establish a new state-of-the-art performance.

%% file: sec/2_relatedwork.tex
\input{figures/vertical_misc_figures}

\section{Related Work}
\label{sec:relwork}

Text-to-image (T2I) models have improved a lot~\cite{rombach2022sd, podell2024sdxl, saharia2022imagen, balaji2022ediff, sauer2024fast, ramesh2021zero}.
Driven by the success of Transformers~\cite{vaswani2017transformer, dosovitskiy2021vit}, we see the emergence of Transformer-based T2I models~\cite{peebles2023scalable, zheng2024fast, gao2023mdtv2, esser2024scaling}.
However, despite the tremendous success, generating images aligned with compositional, multi-subject text prompts remains a challenge.

\paragraph{Controllable generation.}
To improve T2I model controllability, techniques such as
prompt optimization~\cite{hao2023optimizing, hertz2022prompt, mo2024dynamic, witteveen2022investigating},
reward based tuning~\cite{xu2023imagereward}, or
inference-time latent updates~\cite{chefer2023attend}
are popular.
Subsequent works build upon latent update techniques by manipulating cross- and self-attention maps~\cite{tang2023daam, feng2023training, wu2023harnessing, li2023divide, battash2024obtaining, guo2024initno, sundaram2024cocono} during inference to minimize catastrophic neglect.
However, these methods may produce out-of-distribution and unnatural images due to latent state optimization during inference.
In addition to neglect, some T2I methods aim to address incorrect attribute binding~\cite{meral2024conform, jiang2024pac, li2023divide, rassin2023lingbind, feng2023training, feng2024ranni}, but are often limited to few subjects and struggle in compositional scenes with multiple subjects involving multiple attributes.
We propose \modelshort{}, a training-free T2I approach that addresses the above challenges of catastrophic neglect, out-of-distribution images, and incorrect attribute binding.

\paragraph{Layout-guided control with boxes.}
Several layout-guided T2I methods either train or fine-tune models or external modules~\cite{li2023gligen, zhou2024migc, wu2024ifadapter, qu2023layoutllm, zheng2023layoutdiffusion, avrahami2023spatext, yang2023reco, zhang2023adding, nichol2022glide, feng2024ranni, nie2024blob, gu2025roictrl, wang2024instancediffusion} for layout-guided generation.
However, these methods require extensive computational resources.
As an alternative, several training-free methods have emerged~\cite{balaji2022ediff, bansal2023universal, bar2023multidiffusion, chen2024training, endo2024masked, xie2023boxdiff, zhao2023loco, phung2024grounded, dahary2024yourself, xiao2024rnb}.
Among them, methods such as Layout-guidance~\cite{chen2024training}, R\&B~\cite{xiao2024rnb}, or BoxDiff~\cite{xie2023boxdiff} use the cross-attention map to enable layout guidance.
Others like Attention Refocusing~\cite{phung2024grounded} and Bounded Attention~\cite{dahary2024yourself} use cross- and self-attention maps for layout guidance through latent state optimization.
However, for multi-subject prompts, we observe that such methods exhibit background semantic leakage, generate out-of-distribution images, and spread attributes across subjects.
Our approach addresses these challenges through regularization of the latent updates and promoting correct subject-attribute pairs while demoting others. 

%% file: figures/vertical_misc_figures.tex
\begin{figure*}[t]
\centering
\small
\tabcolsep=0.16mm

\begin{tabular}{c ccccccc}

\raisebox{0.88cm}[\height][\depth]{\rotatebox{90}{Layout}} &
\includegraphics[width=0.14\textwidth]{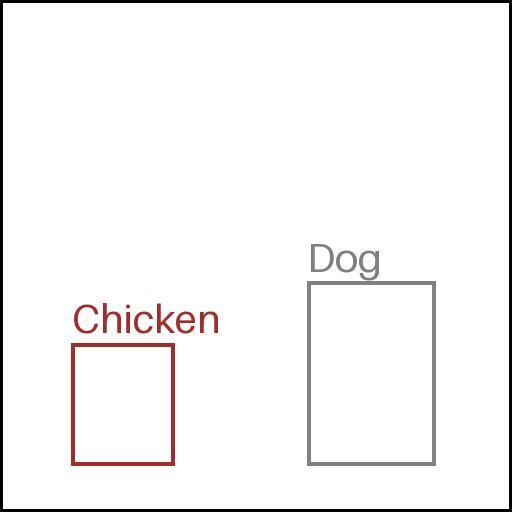} &

\includegraphics[width=0.14\textwidth]{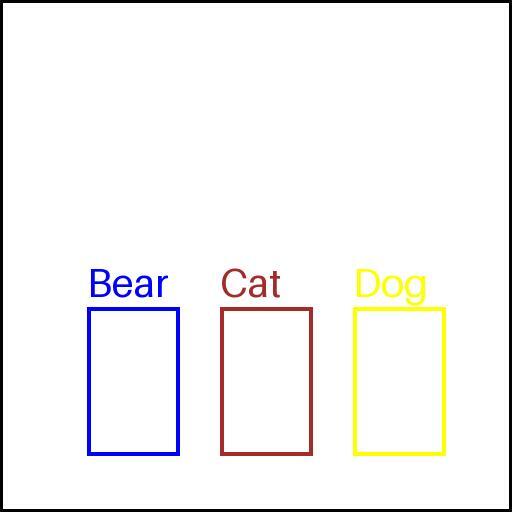} &
\includegraphics[width=0.14\textwidth]{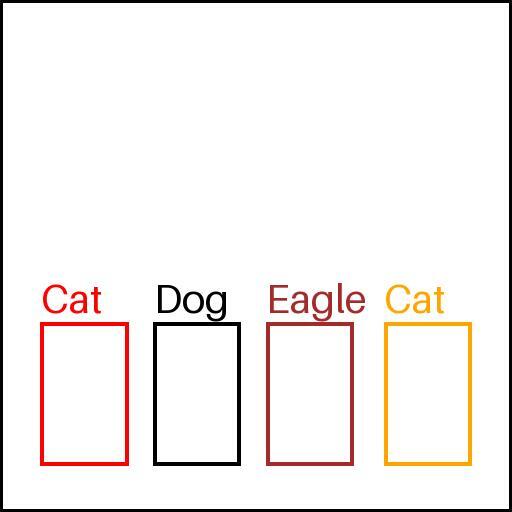} &
\includegraphics[width=0.14\textwidth]{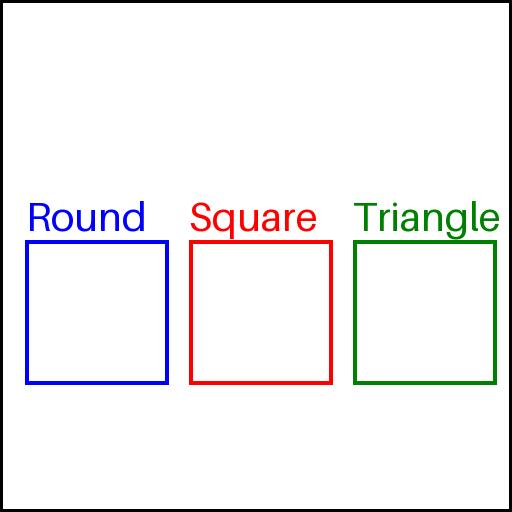} &
\includegraphics[width=0.14\textwidth]{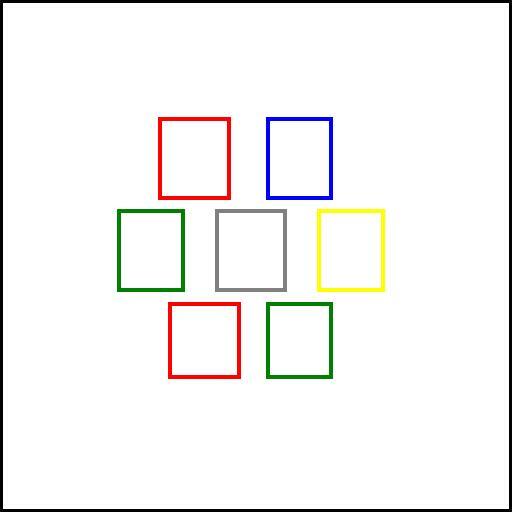} &
\includegraphics[width=0.14\textwidth]{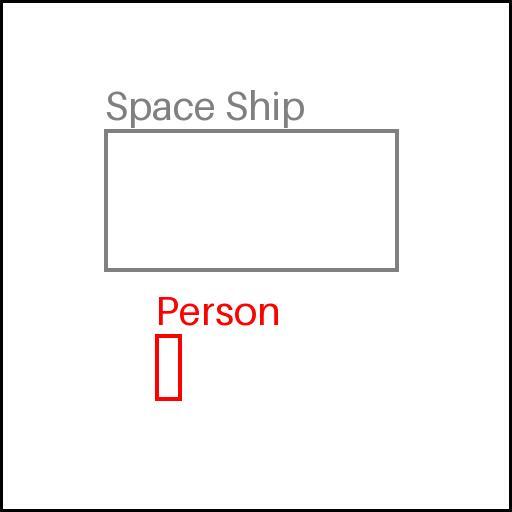} &
\includegraphics[width=0.14\textwidth]{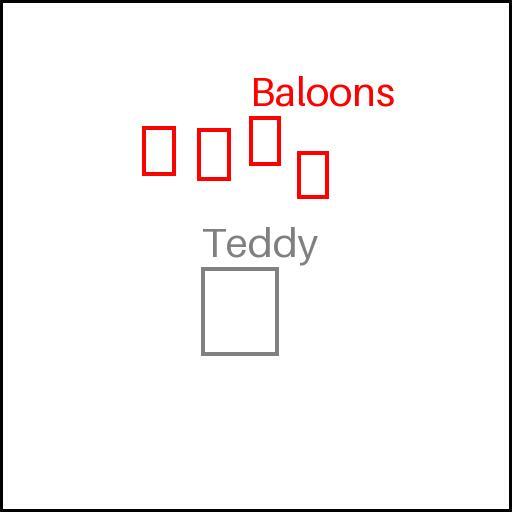} \\

\raisebox{0.15cm}[\height][\depth]{\rotatebox{90}{Bounded Attention}} &
\includegraphics[width=0.14\textwidth]{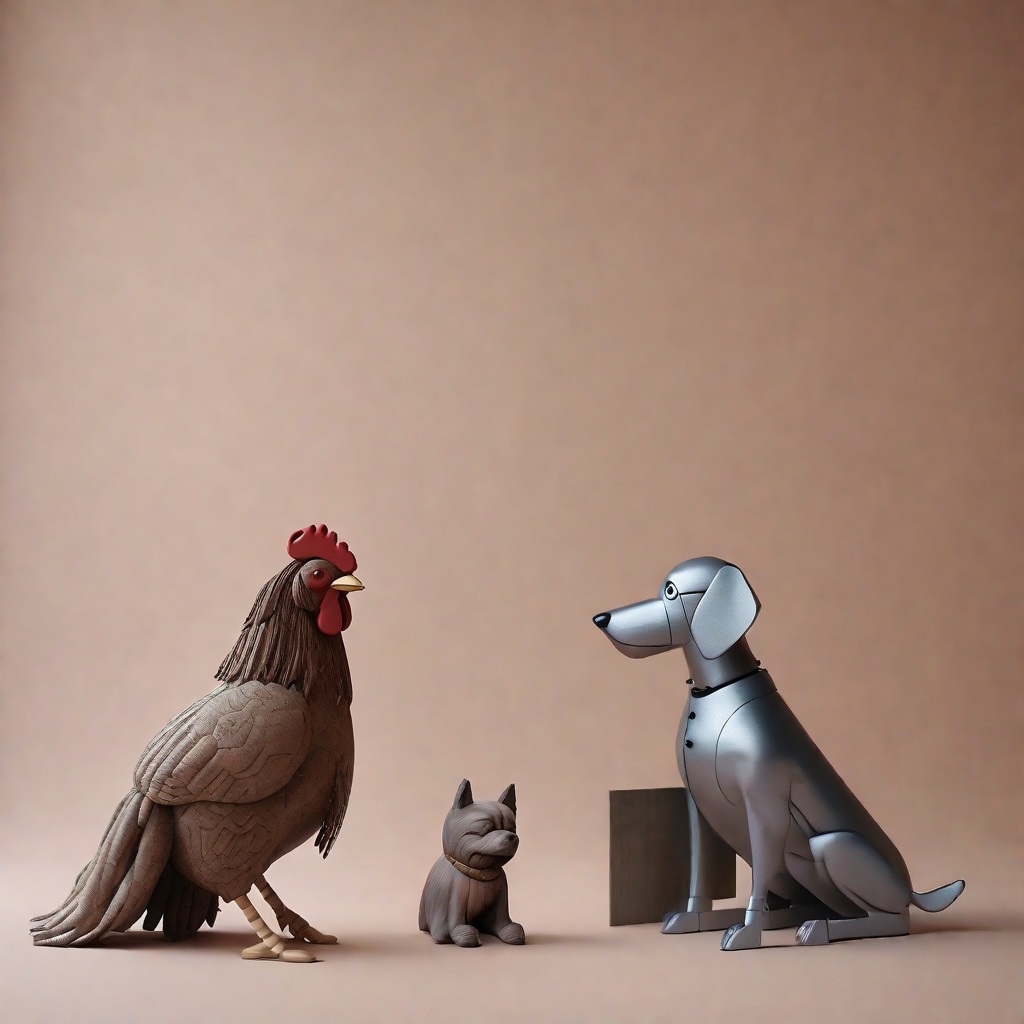} &

\includegraphics[width=0.14\textwidth]{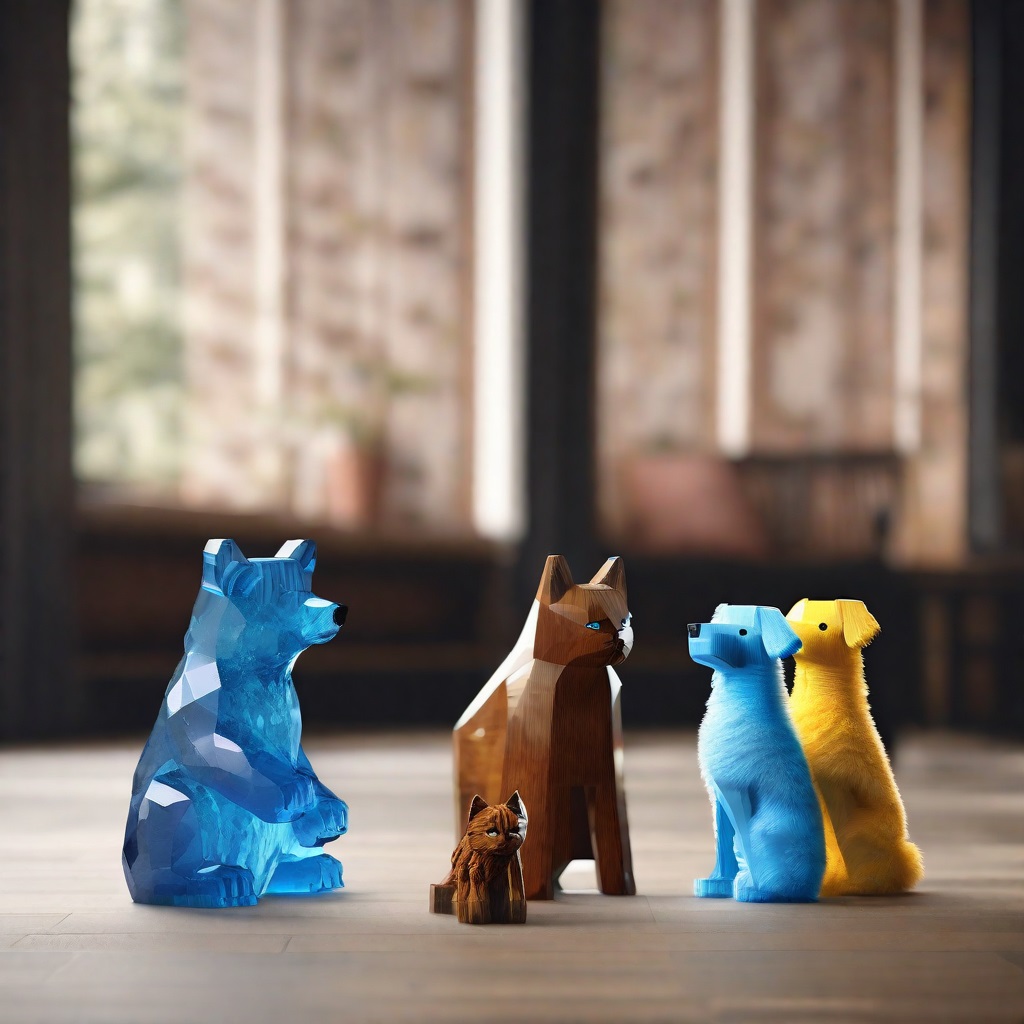} &
\includegraphics[width=0.14\textwidth]{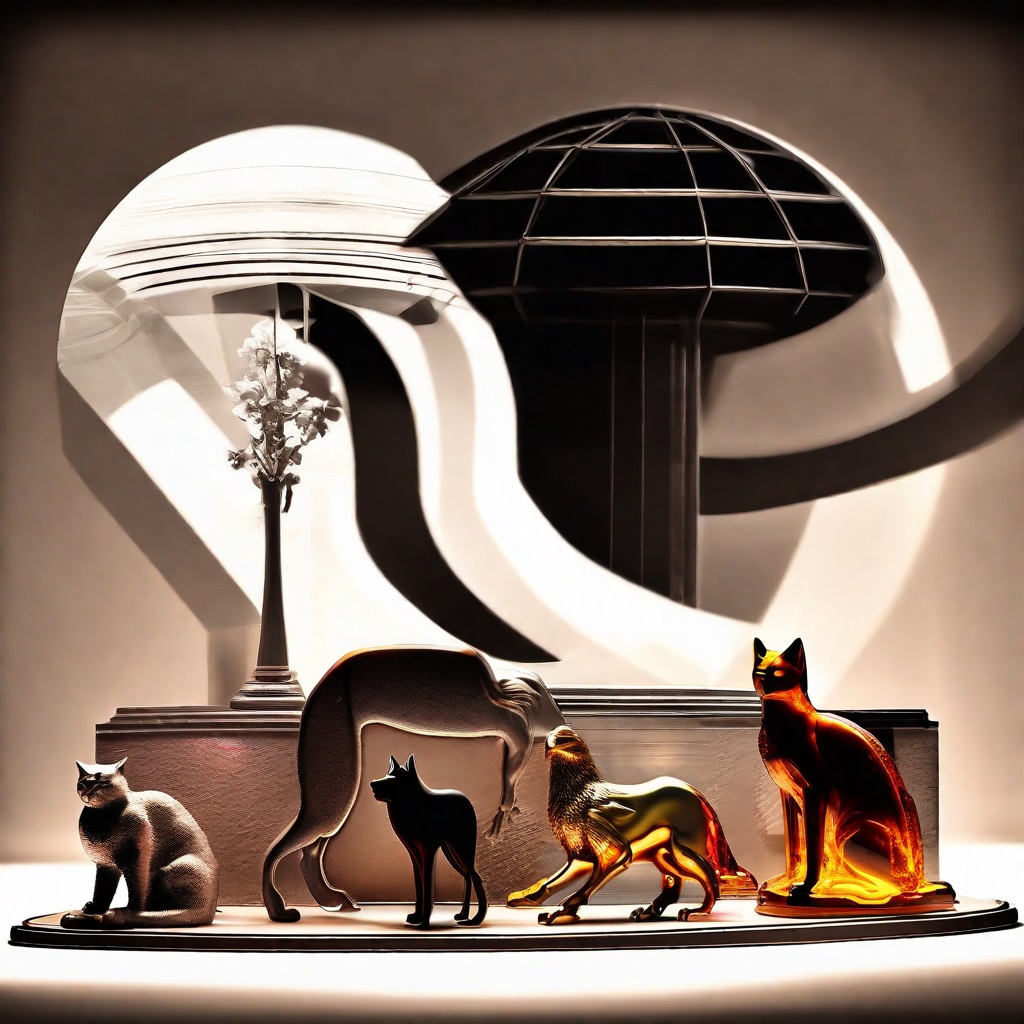} &
\includegraphics[width=0.14\textwidth]{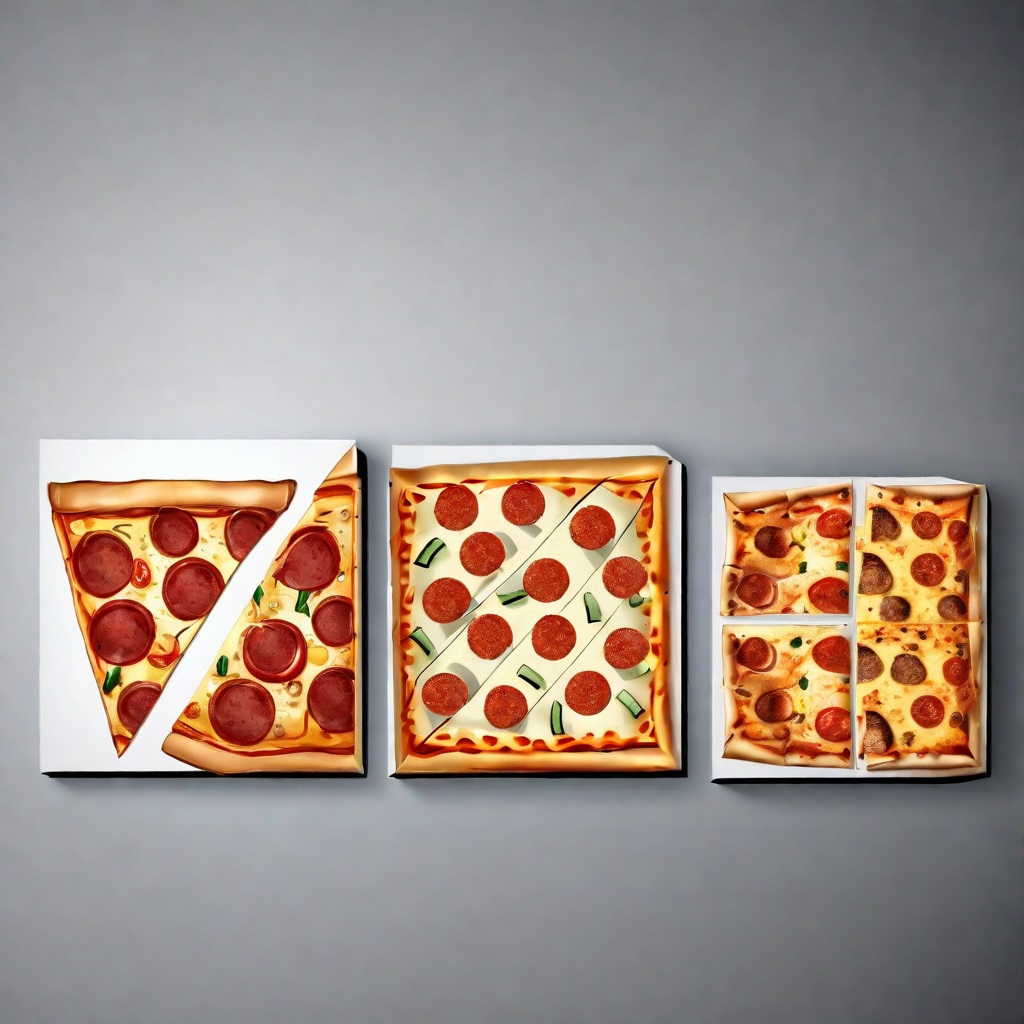} &
\includegraphics[width=0.14\textwidth]{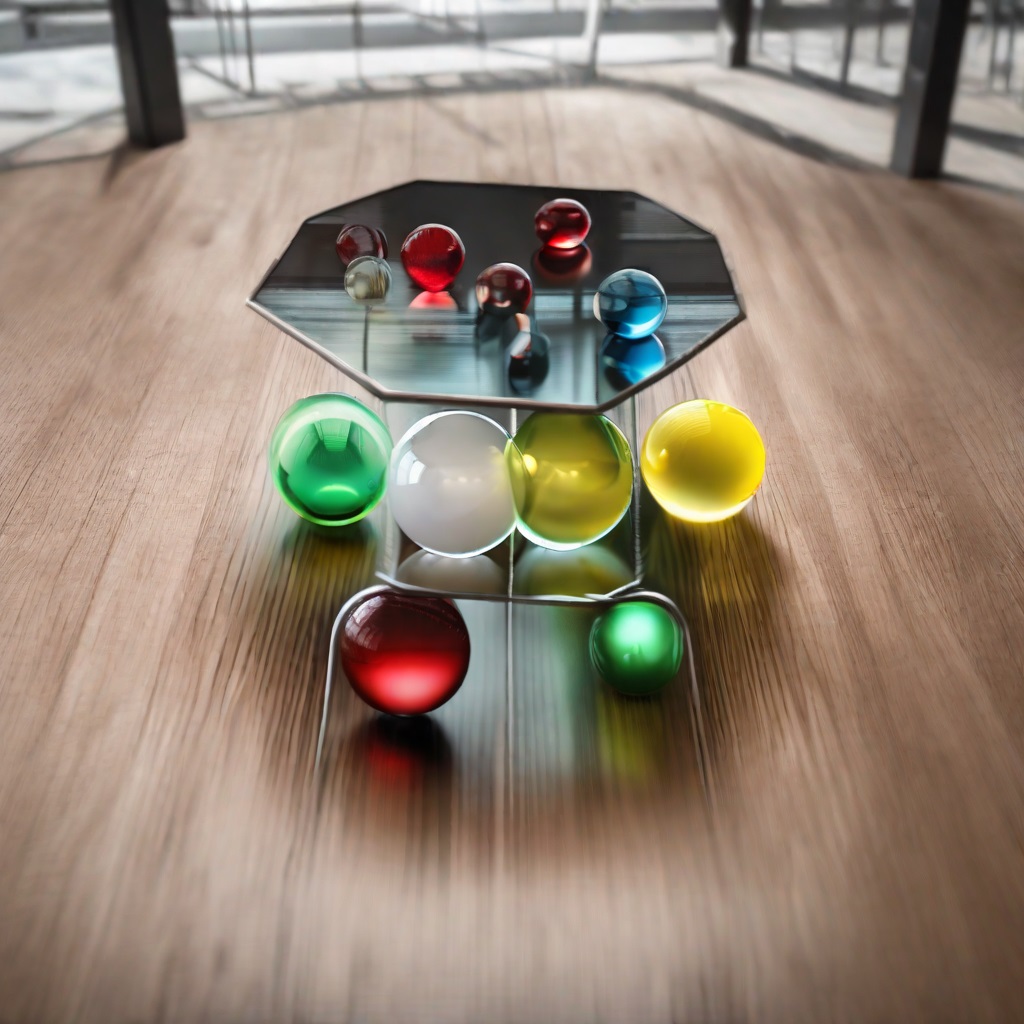} &
\includegraphics[width=0.14\textwidth]{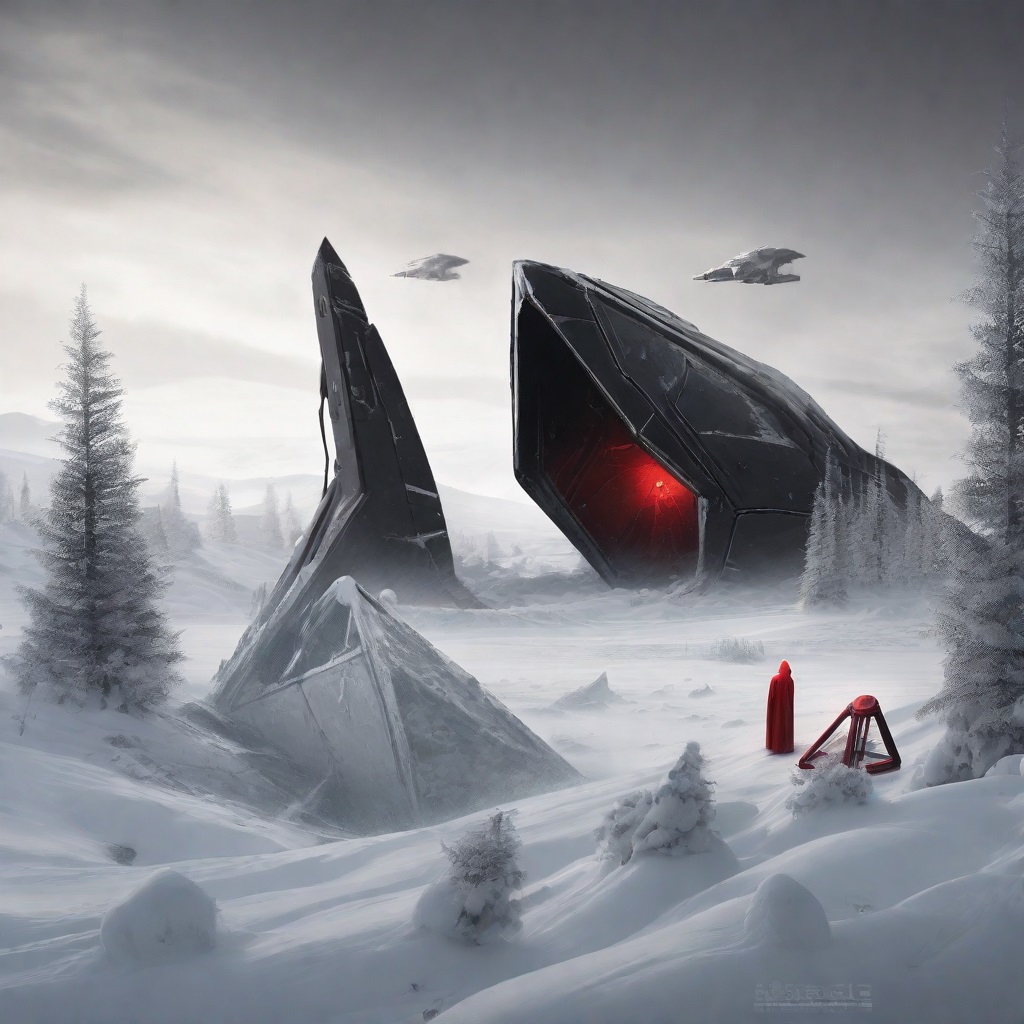} &
\includegraphics[width=0.14\textwidth]{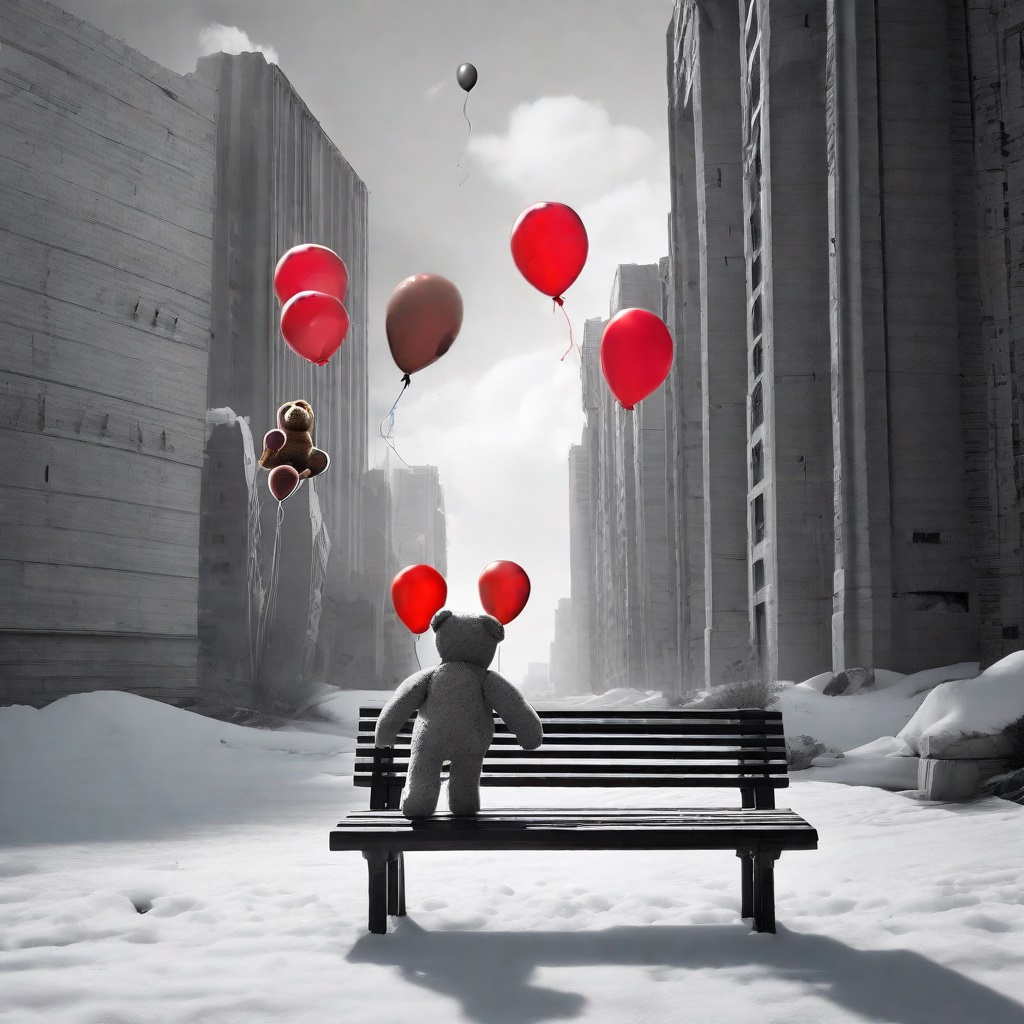} \\

\raisebox{0.48cm}[\height][\depth]{\rotatebox{90}{\modelshort{} (Ours)}} &
\includegraphics[width=0.14\textwidth]{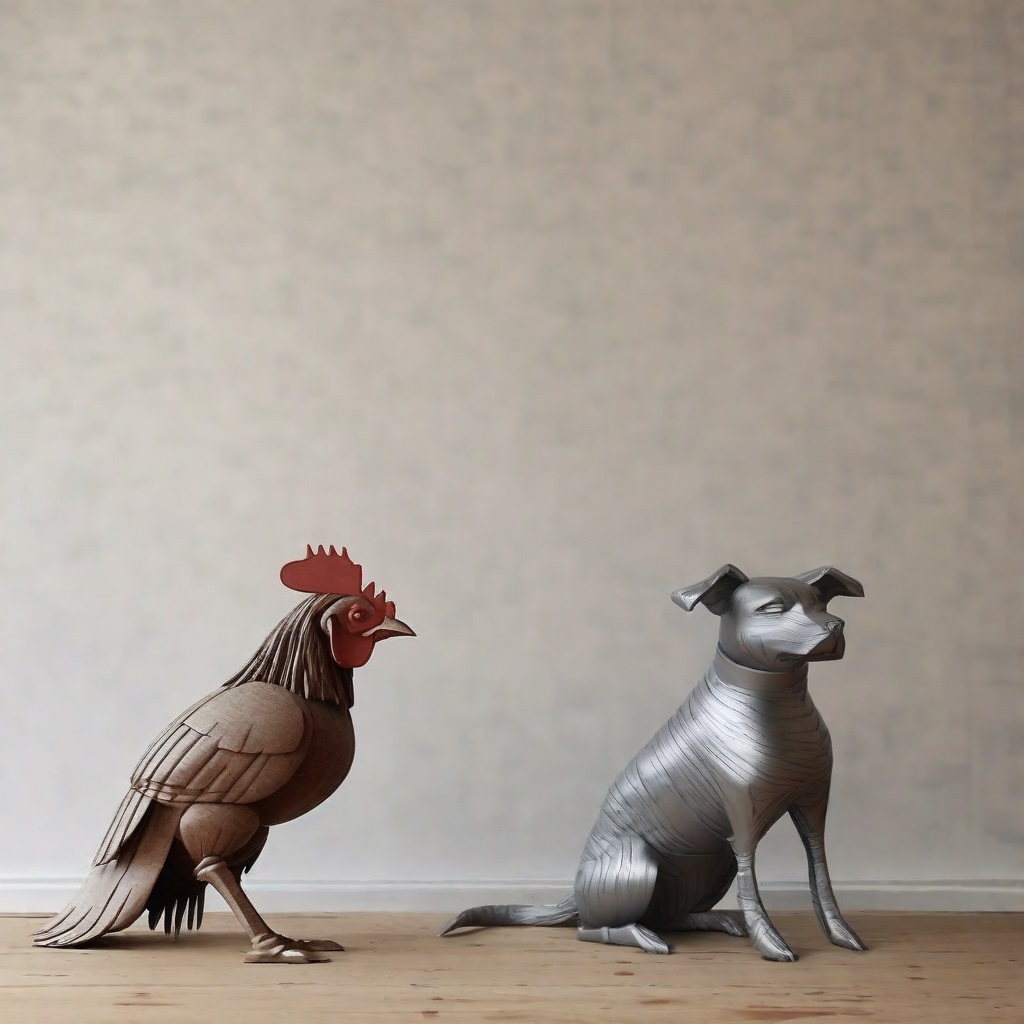} &

\includegraphics[width=0.14\textwidth]{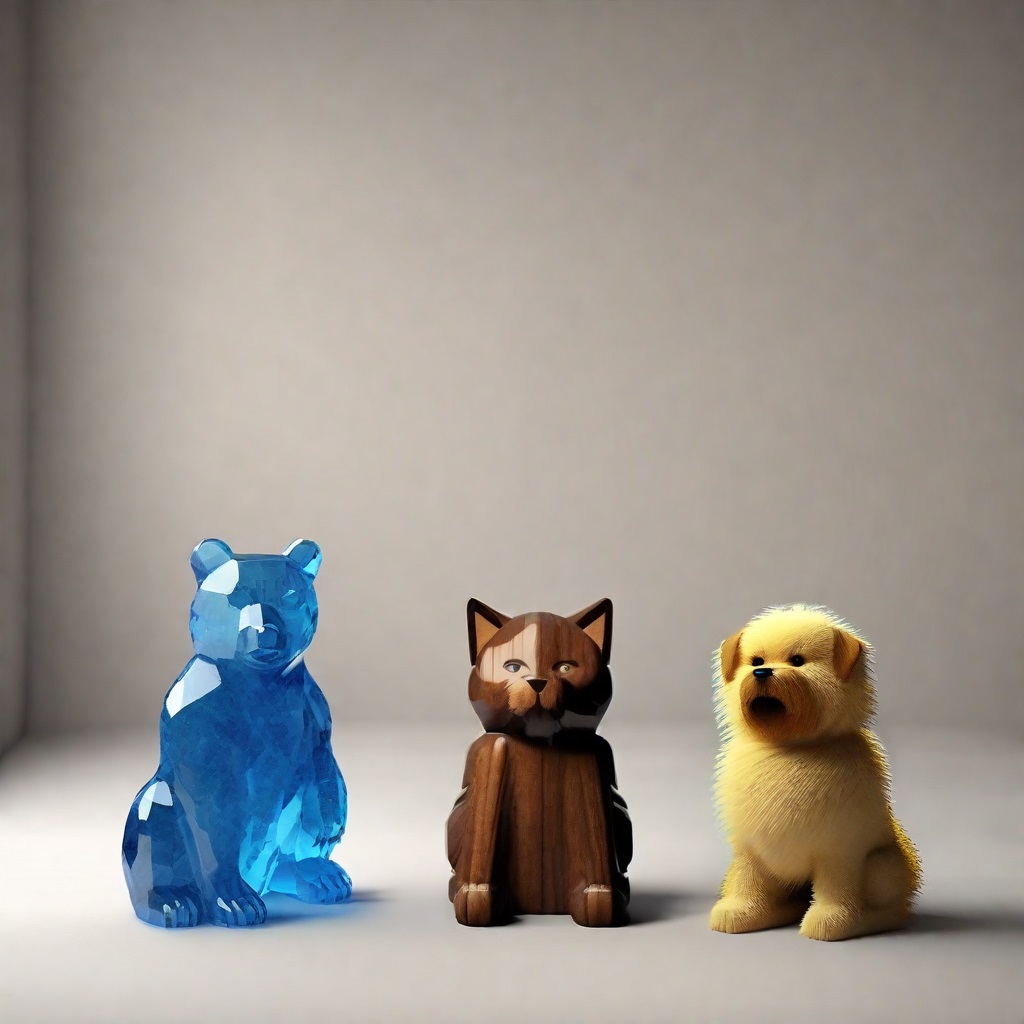} &
\includegraphics[width=0.14\textwidth]{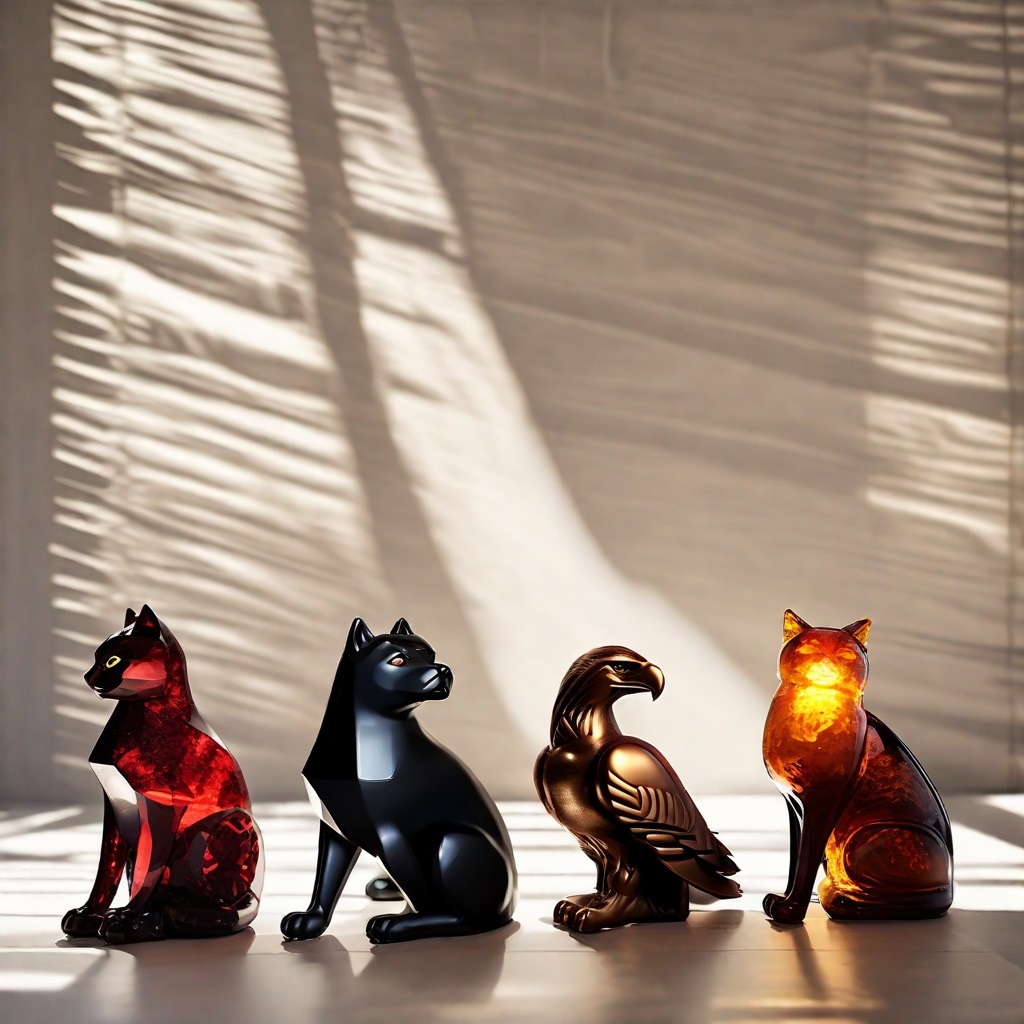} &
\includegraphics[width=0.14\textwidth]{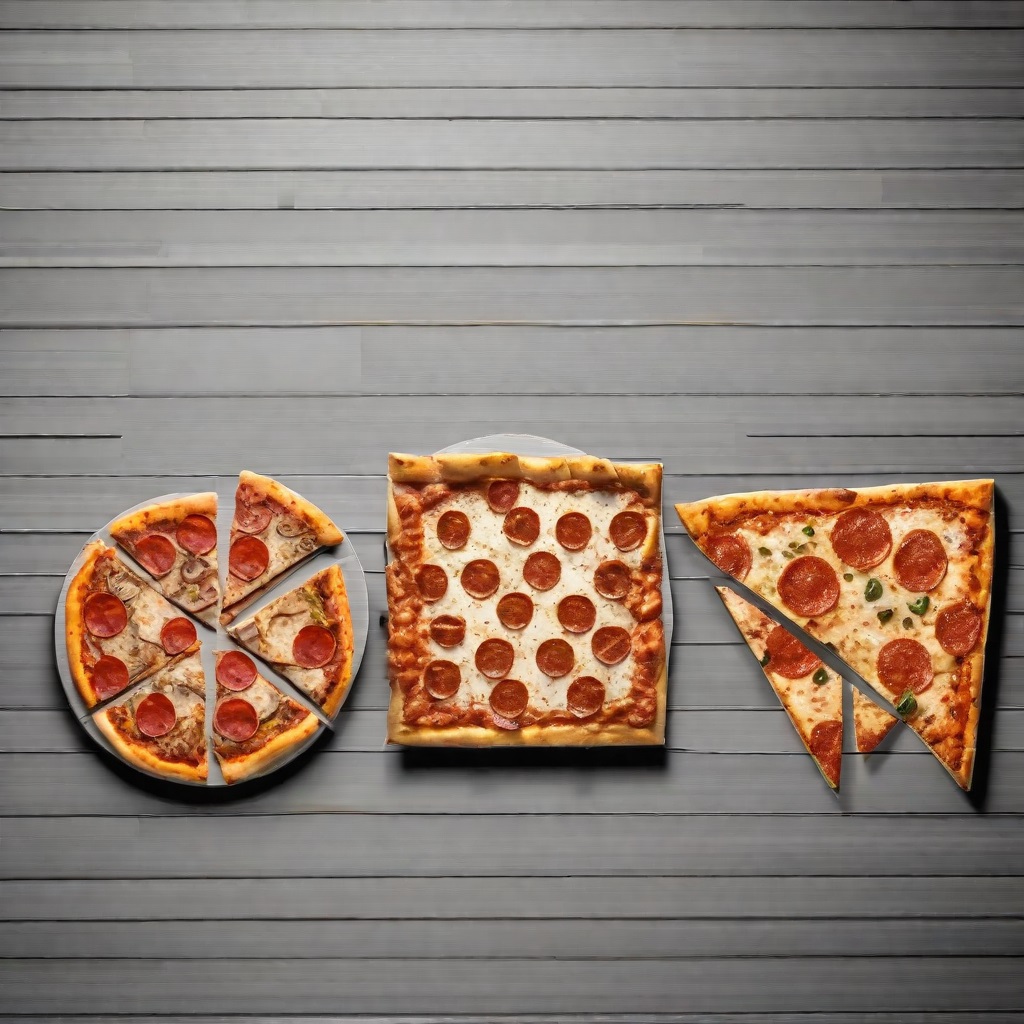} &
\includegraphics[width=0.14\textwidth]{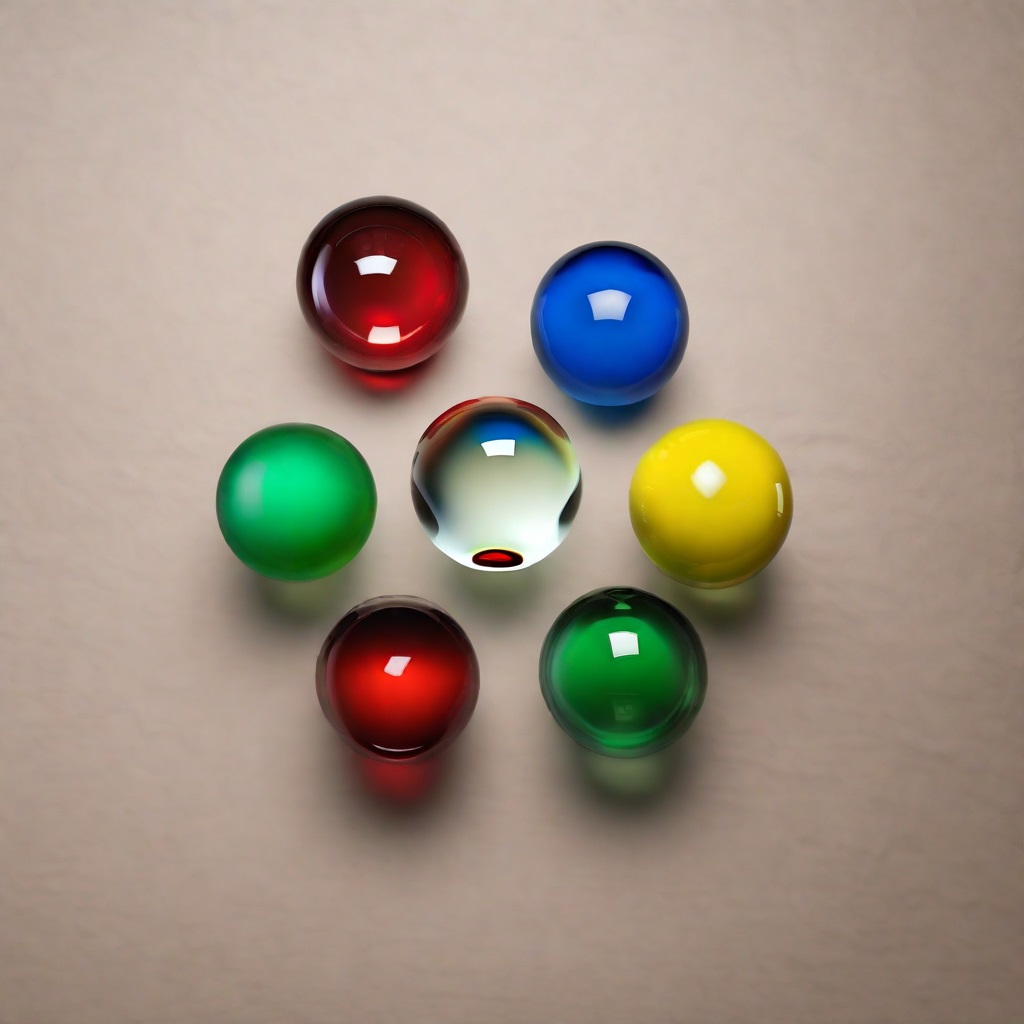} &
\includegraphics[width=0.14\textwidth]{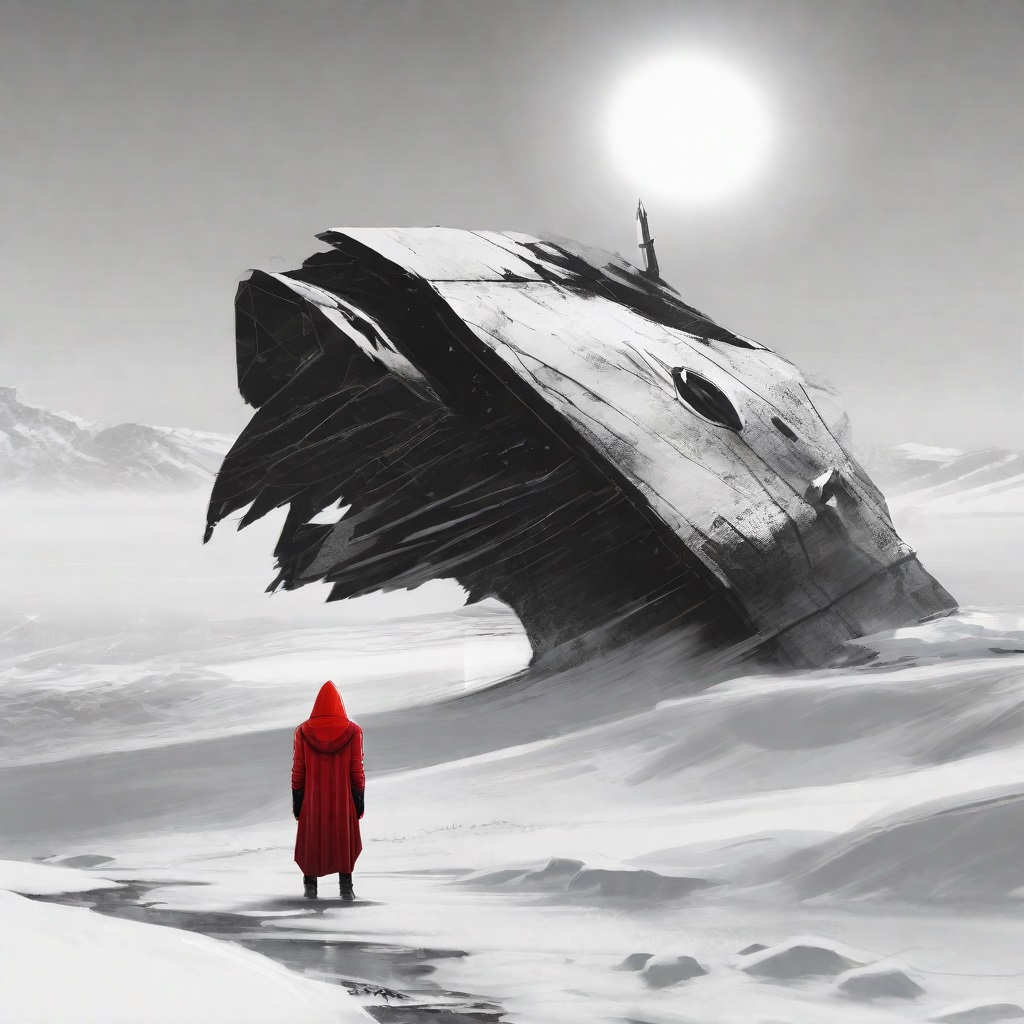} &
\includegraphics[width=0.14\textwidth]{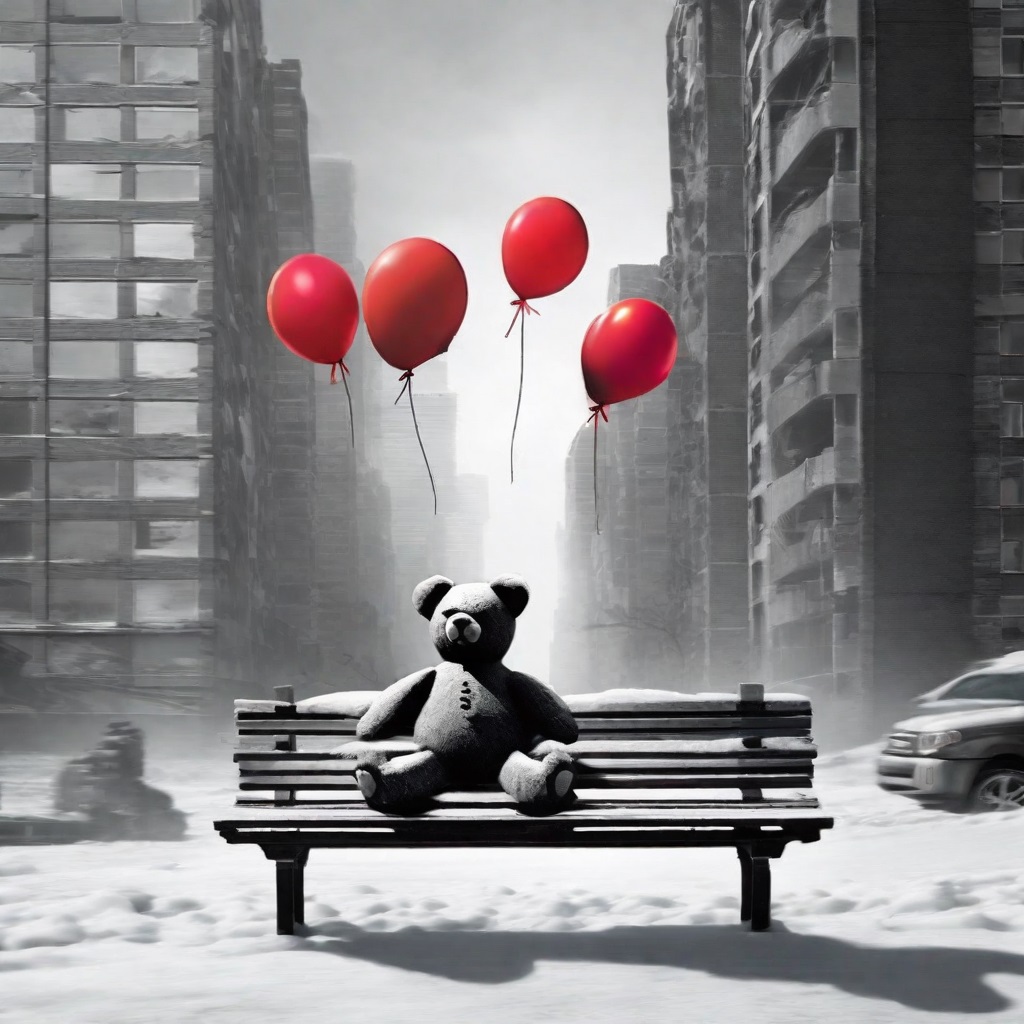} \\
\end{tabular}
\vspace{-4mm}
\caption{\modelshort{} (SDXL) outperforms Bounded Attention~\cite{dahary2024yourself} on complex prompts with multiple subjects and attributes.
Both BA and \modelshort{} use the same seed.
The prompts from L2R are:
1. A realistic photo of a \textbf{\color{brown} brown wooden chicken} and a \textbf{\color{gray} gray metallic dog.} \\
2. A realistic photo of a \textbf{\color{blue} blue crystal bear} and a \textbf{\color{brown} brown wooden cat} and a \textbf{\color{Goldenrod} yellow fluffy dog}. \\
3. A realistic photo of a \textbf{\color{red} shiny red crystal cat} and a \textbf{\color{black} black matte plastic dog} and a \textbf{\color{brown} rustic bronze eagle} and a \textbf{\color{orange} glowing amber cat}. \\
4. A \textbf{\color{blue} round pizza} and a \textbf{\color{red} square pizza} and a \textbf{\color{Green} triangle pizza}. \\
5. A realistic photo of \textbf{\color{red} two red glass sphere} and a \textbf{\color{blue} blue glass sphere} and \textbf{\color{Green} two green glass sphere} and a \textbf{\color{Goldenrod} yellow glass sphere} and a \textbf{\color{gray} white glass sphere}. \\
6. A black and white concept art of a \textbf{\color{gray} crashed spaceship} partially buried in icy landscape and a \textbf{\color{red} red hooded person} is watching it from a distance. \\
7. A black and white concept art of a destroyed apocalyptic city covered with snow and a decaying \textbf{\color{gray} teddy bear} on a bench with \textbf{\color{red} four red balloons} tied.
}
\vspace{-2mm}
\label{fig:vertical_misc_fig}
\end{figure*}

%% file: sec/3_method.tex
\section{Our Approach: \modelshort{}}
\label{sec:method}

We present the components of \modellong{} (\modelshort{}):
(i)~masked latent regularization prevents background semantic leakage (\cref{subsec:method:masked_latent}),
(ii)~in-distribution latent alignment avoids artifacts (\cref{subsec:method:in_dist}), and
(iii)~subject-attribute association improves binding (\cref{subsec:method:attribute_bind}).
First, we re-visit the fundamentals of inference-time latent optimization in layout-guided T2I models.

\subsection{Preliminaries}
\label{subsec:method:preliminaries}

Different from pixel space diffusion~\cite{ho2020ddpm}, a Latent Diffusion Model (LDM)~\cite{rombach2022sd} operates in the latent embedding space.
It has an autoencoder that encodes an image $\bx$ to the latent space $\bz = \ME(\bx)$ and reconstructs the image through a decoder $\hat{\bx} = \MD(\bz)$.
During training, noise is gradually added to the original latent state $\bz_0$ to obtain $\bz_t$.
During inference, a UNet~\cite{ronneberger2015unet}, equipped with self- and cross-attention layers acts as a denoiser.
Starting from random noise $\bz_T \sim \MN(0, \bI)$, the denoiser estimates and removes the noise $\hat{\epsilon}_t = \phi(\bz_t, t, \by)$ conditioned on the time step $t$ and the text prompt $\by$.
Specifically, we adopt the DDIM~\cite{song2021ddim} update mechanism.

\paragraph{Attention layers.}
At each layer of the UNet, the prompt embedding $\by$ is injected through cross-attention layers to steer the image towards the prompt.
Specifically, spatial UNet features $\phi(\bz_t)$ are projected to obtain queries $Q = W_q \phi(\bz_t)$, while keys $K = W_k \by$ and values $V = W_v \by$ are obtained using the prompt embedding.
The \textit{cross}-attention map at $t$ is calculated as
$A_t^c = \text{softmax}(Q K^T)$.
While the cross-attention maps $A_t^c \in \real{hw \times n}$ capture the relationship between the latent of spatial dimensions $hw$ against $n$ prompt tokens, the self-attention maps $A_t^s \in \real{hw \times hw}$ depict the spatial correspondence between latents.

\paragraph{Latent optimization.}
A popular approach to improve inference-time prompt adherence of LDMs is to optimize the latent $\bz_t$ at each time step during sampling:
\begin{equation}
\label{eq:latent_update}
\bz_t' \xleftarrow{} \bz_t - \alpha_t \cdot \nabla_{\bz_t} \ML,
\end{equation}
where $\alpha_t$ is the update rate.
$\ML$ defines the desired objective,
\eg~\citet{chefer2023attend} minimize catastrophic neglect by ensuring high attention of at least one latent region to each subject token $s_i \in \MS$:
\begin{equation}
\label{eq:attend_excite}
\mathcal{L} = \max_{s_i} (1 - \max_{hw}( A_t^c(s_i) )),
\end{equation}
where $A_t^c(s_i) \in \real{hw}$ represents the spatial cross-attention to token $s_i$ of the prompt.

\paragraph{Layout-guided generation.}
Subjects in the prompt are associated with spatial guidance in the form of bounding boxes by pairing token indices with boxes $(s_i, b_i)$.
Latent optimization encourages subjects to be present inside the bounding boxes~\cite{phung2024grounded, dahary2024yourself}.
Specifically, the latent updates minimize two losses:
a cross-attention loss $\ML_c$, encourages generated subjects to be present inside the corresponding bounding boxes; and
a self-attention loss $\ML_s$, prevents latent pixels from attending to irrelevant regions.
For both losses, previous works~\cite{xiao2024rnb, phung2024grounded, dahary2024yourself} use an intersection-over-union (IoU) formulation that encourages attention to focus inside the bounding box region while disregarding the rest.
For a subject token index $s_i$ with box $b_i$, this is defined as:
\begin{equation}
\label{eq:Liou}
\vspace{1mm}
\ML_{i} = 1 - \frac{\sum \hat{A}_t[b_i]}{\sum \hat{A}_t[b_i] + \gamma \sum \hat{A}_t[\bar{b_i}]},
\quad \text{and} \quad
\ML_{\text{iou}} = \sum_i \ML_{i}^2 \, ,
\vspace{1mm}
\end{equation}
where $\hat{A}_t$ is the aggregated self- or cross-attention map (heads and layers) at step $t$.
$\hat{A}_t[b_i]$ corresponds to attention within the subject's box $b_i$ and $\hat{A}_t[\bar{b_i}]$ to regions outside the box.
$\gamma$ is the number of subjects in the prompt and it amplifies the attention towards the background.
For details, we refer the reader to~\cite{dahary2024yourself} and adopt this IoU loss as our baseline.

\subsection{Masked Latent Regularization}
\label{subsec:method:masked_latent}

Layout-guided generation methods aim to create images where subjects adhere to both, the text prompts and the bounding boxes.
However, we observe that current methods exhibit \textit{background semantic leakage} where multiple subjects, not specified in the prompts (box, text, or both), appear in the image (see \cref{fig:teaser}a).
We see that this becomes frequent with increasing number of subjects.

The first few inference time steps of diffusion models largely determine the layout~\cite{hertz2022prompt}.
Subjects emerge first while fine-grained details later.
While previous works optimize the latent for 15 to 25 steps~\cite{dahary2024yourself}, we find that the object layout is already determined in the first 5 steps.
In fact, over optimization of the latent leads to out-of-distribution images.

While optimizing the latent $\bz_t$, we wish to discourage subject patterns from forming outside bounding boxes.
We do so by constraining the background latents to remain close to their original values.
Consider $\bz_t^\text{ref}$ as a detached copy of the latent before the update ($\bz_t$ in \cref{eq:latent_update}).
We create a binary mask $M$ with value 1 in regions corresponding to a bounding box and 0 elsewhere, and let
$\bar{M}$ represent the inverted background mask.
We introduce a masked regularization term that penalizes deviations and incorporate it in the latent update as:
\begin{align}
\vspace{1mm}
&\Lmask = \left\| (\bz_t^{(k)} - \bz_t^\text{ref}) \odot \bar{M} \right\|_1, \quad \text{and} \\
&\bz_t'^{(k+1)} \leftarrow{} \bz_t^{(k)} - \alpha_t \cdot \nabla_{\bz_t} \left(\Liou + \lambdamask \Lmask \right) ,
\vspace{1mm}
\end{align}
where $\lambdamask$ is a hyperparameter and we perform iterative refinement $k$ times at each time step $t$, similar to~\cite{chefer2023attend}.
This approach is illustrated in \cref{fig:method}a with highlighted masks.

\subsection{In-Distribution Latent Alignment}
\label{subsec:method:in_dist}

\begin{figure}[t]
\centering
\includegraphics[width=1\linewidth]{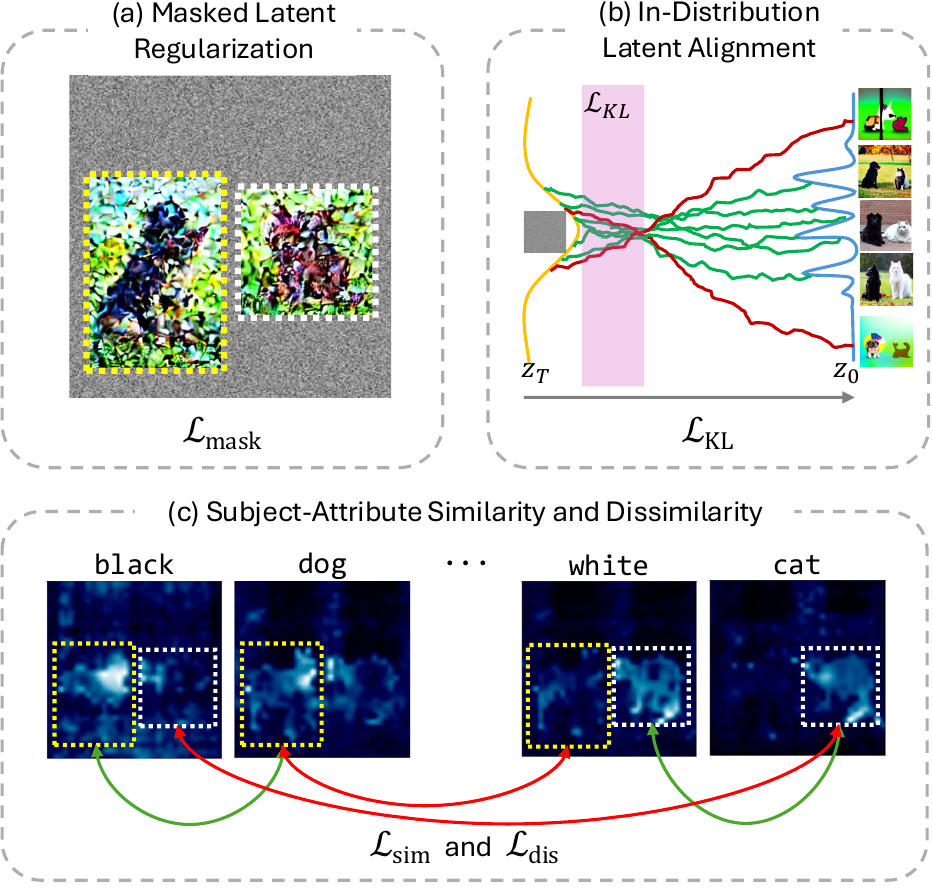}
\vspace{-6mm}
\caption{\textbf{\modelshort{} overview.}
We illustrate three key components for layout-guided compositional scene generation.
(a)~Masked latent regularization prevents background semantic leakage,
(b)~KL-based alignment keeps the latent in-distribution during optimization, and
(c)~layout-guided subject-attribute association enables accurate compositional binding.
}
\vspace{-2mm}
\label{fig:method}
\end{figure}

Latent space updates risk pushing $\bz_t$ away from its typical noise distribution, resulting in out-of-distribution images.
In layout-guided methods, the imbalance between latent updates within and outside the boxes further increases this chance.
To mitigate this problem, we align the latent back to a stable prior distribution.
In the early stages of the denoising process, the latent is close to the random noise distribution, $\bz_t \sim \MN(0, \bI)$.
Thus, we introduce an alignment term (for 5 steps) based on the Kullback-Leibler (KL) divergence~\cite{kullback1951information} and add it to the latent update loss as:
\begin{align}
&\Lkl = D_\text{KL} \left( \MN(\mu_{\bz_t}, \sigma_{\bz_t}) \| \MN(0,1) \right), \\
&\ML = \Liou + \lambdamask \Lmask + \lambdakl \Lkl.
\end{align}
$\lambdakl$ is a hyperparameter that controls the strength of the KL term.
The combined formulation not only prevents background leakage but also constrains the tendency of $\Liou$ and $\Lmask$ from pushing $\bz_t$ out-of-distribution during the early phase of optimization (see \cref{fig:method}b).
The outcome is generation of higher fidelity images without needing multiple random initializations (\cref{fig:vertical_misc_fig}, \cref{fig:mainfig}).

\subsection{Layout-guided Subject-Attribute Association}
\label{subsec:method:attribute_bind}

A challenge in layout-guided scene composition is binding attributes to the right subjects.
To address this, we extract nouns (subjects) and their modifiers (attributes) from the prompt and propose constraints on the attention mechanism.
Specifically, we propose that the spatial cross-attention patterns should be similar for aligned subject-attribute pairs \emph{and} dissimilar for unaligned pairs.

Concretely, consider a set of subject tokens indicated through their indices $\MS = \{s_1, \ldots, s_S\}$ in the prompt and associated with boxes $\MB = \{b_1, \ldots, b_S\}$.
Let us denote the set of attribute token indices for each subject as $\MA = \{a_1, \ldots, a_S\}$ (no attribute is subsumed with $a_i = \varnothing$).
In addition to the subject token and bounding box $(s_i, b_i)$, we now include attribute token indices $(s_i, b_i, a_i)$.
Note, $a_i$ may correspond to multiple attribute tokens depending on the prompt (\eg~white and marble lion in \cref{fig:teaser}).
Recall, $A^c_t \in \real{hw \times n}$ represents the cross-attention map at time step $t$.
For a specific subject token $s_i$, we denote $A^c_t(s_i)[b_i] \in \real{h_i w_i}$ as the cross-attention in the spatial region corresponding to the bounding box $b_i$ of size $h_i \times w_i$.
Similarly $A^c_t(a_i)[b_i]$ corresponds to the attention scores of the box region with respect to the attribute token(s).
We renormalize these attention patches to be a probability.

\paragraph{Subject-attribute similarity.}
The association between paired subjects and attributes is improved by encouraging similar cross-attention maps within the bounding box.
We calculate the similarity loss between subject $s_i$ and attribute $a_i$ using a symmetric KL divergence:
\begin{align}
&\Lsim(i) = D_\text{sym}(A^c_t(s_i)[b_i], A^c_t(a_i)[b_i]), 
\quad \text{where} \\
&D_\text{sym}(P, Q) = \frac{1}{2} D_\text{KL}(P \| Q) + \frac{1}{2} D_\text{KL}(Q \| P) .
\end{align}
The total similarity loss $\Lsim = \text{mean}_i \Lsim(i)$.

\paragraph{Subject $\times$ attribute dissimilarity.}
Minimizing the similarity loss alone is insufficient to bind attributes to objects (\cref{fig:loss_ablation}e).
Similar to the triplet loss~\cite{schroff2015facenet},
we propose a dissimilarity loss between unaligned subject-attribute boxes.
We consider all mismatched subject-attribute combinations $(s_i, b_i, a_j), j\neq i$ for whom the cross-attention maps should be dissimilar.
The dissimilarity loss is formulated as a negative symmetric KL divergence operating on the cross-attention region $b_i$ for tokens $s_i$ and $a_j$:
\begin{equation}
\Ldis(i,j) = - D_{\text{sym}}(A_t^c(s_i)[b_i] , A_t^c(a_j)[b_i]).
\end{equation}
The total loss is $\Ldis = \text{mean}_{i,j,i\neq j} \Ldis(i,j)$.

\paragraph{Why adopt symmetric KL divergence?}
Cross-attention maps have been treated as a probability distribution in prior works, and the distance between subject and attribute cross-attention maps is minimized to improve binding.
\citet{li2023divide} employ JS-Divergence, while \citet{jiang2024pac} use symmetric KL divergence. 
However, we observe that directly minimizing the distance between cross-attention maps is ineffective in the presence of multiple subjects with multiple attributes.
Layout conditioning in our method specifies the exact regions where the subject and attribute cross-attention maps need to be similar.
Thus, we minimize the distance between \textit{normalized cross-attention maps} within the masked regions corresponding to each subject and its attributes. 
However, even after this alignment, attributes may still leak to other subjects in multi-subject scenes (\cref{fig:loss_ablation}e).
Our masked dissimilarity loss prevents this leakage by maximizing the distance between each subject and its non-corresponding attributes. Finally, we empirically find symmetric KL to be much more effective than JS-divergence.

\paragraph{Total attribute loss and visualization.}
The final attribute association loss,
$\Latt = \lambdasim \Lsim + \lambdadis \Ldis$,
where $\lambdasim$ and $\lambdadis$ are hyperparameters, controlling the strength of similarity and dissimilarity.
\cref{fig:method}c shows cross-attention maps for the example prompt ``a black dog and a white cat'' with ($s_1$: dog, $a_1$: black) and ($s_2$: cat, $a_2$: white).
The cross-attention maps for aligned pairs (dog, black) or (cat, white) show high activations within the provided box guidance.
Similarly, the attention maps for unaligned pairs (cat, black) and (dog, white) show significant differences.

\paragraph{Final training objective.}
The overall objective is a combination of multiple terms inducing regularization, alignment to the prior distribution, and encouraging correct subject-attribute association:
\begin{equation}
\label{eq:final_loss}
\ML = \Liou + \lambdamask \Lmask + \lambdakl \Lkl + \lambdaatt \Latt.
\end{equation}
This combined formulation improves background semantic leakage, reduces out-of-distribution images, and provides more accurate attribute binding even with multiple attributes for each subject, generating high fidelity compositional scenes.

%% file: sec/4_experiments.tex
\input{figures/mainfig}

\section{Experiments}
\label{sec:experiments}

\paragraph{Baselines.}
We compare \modelshort{} against seven previous approaches in layout-guided T2I.
They include:
GLIGEN~\cite{li2023gligen},
Attention Refocusing~\cite{phung2024grounded},
BoxDiff~\cite{xie2023boxdiff}, 
LayoutGuidance~\cite{chen2024training}, 
ReCo~\cite{yang2023reco},
R\&B~\cite{xiao2024rnb}, and
Bounded Attention (BA)~\cite{dahary2024yourself} (equivalent of $\Liou$ only loss).
We present results with two backbones:
SD-1.5 and SDXL~\cite{podell2024sdxl}, for fair comparisons.

\paragraph{Benchmarks.}
We evaluate layout-guided T2I methods on two benchmarks:
DrawBench~\cite{saharia2022imagen} and 
HRS~\cite{bakr2023hrs}.
DrawBench is well-established, with challenging prompts to evaluate spatial reasoning and counting capabilities of T2I models.
It provides 39 prompts for \textit{counting} (19) and \textit{spatial} (20) relationships.
We also evaluate \modelshort's ability to bind attributes correctly using 25 prompts featuring 9 colors from the \textit{color} task.
Notably, prior methods have skipped this category
On HRS benchmark, we follow the protocol established by R\&B~\cite{xiao2024rnb}, and report results on
spatial, color, and size,
to demonstrate the effectiveness of our approach.
We use bounding boxes provided by~\citet{phung2024grounded}, that are generated automatically using GPT-4.

\paragraph{Evaluation metrics.}
We follow standard evaluation protocols~\cite{phung2024grounded, dahary2024yourself, xiao2024rnb}.
For counting, we use an off-the-shelf object detector and compare its output to the ground-truth prompt and calculate precision, recall, and F1 score.
Accuracy is adopted for the spatial, color, and size prompts.

\paragraph{Implementation details.}
All experiments are performed on A6000 GPUs.
On DrawBench, results are averaged across 4 seeds (0, 42, 2718, 31415).
For HRS, we adopt seed 0.
All qualitative comparisons are made across the same seed.
We empirically choose $\lambdamask {=} 0.01$, $\lambdakl {=} 5$, and $\lambdasim {=} \lambdadis {=} \lambdaatt {=} 1$ as they give consistently good results.
$\Lmask$ and $\Lkl$ are applied for the first 5 denoising steps, while $\Latt$ is applied for the first 18 of 50 denoising steps.
At each denoising step when the loss term is applied we perform $k{=}5$ gradient descent iterations.
For the step size $\alpha$ in Eq.~(1), we linearly decrease it from 30 to 8 across the denoising steps and the attention map layers used for optimization are the same as BA~\cite{dahary2024yourself}.

\vspace{-2mm}
\subsection{Results}
\label{subsec:exp:sota}

\paragraph{Qualitative comparison.}
\cref{fig:vertical_misc_fig} demonstrates \modelshort{}'s ability to create compositional scenes involving multiple subjects and multiple attribute types.
Our approach performs well beyond the typical color attribute and showcases subjects with various material properties.
For example, in columns 1-3, our method accurately applies color and material properties to each subject,
while BA creates extra subjects and confuses attributes across them.
Columns 4-7 further highlight our model's ability to create images with correct
pizza shapes,
location and color for multiple subjects (spheres),
and rich art-like scenes (concept art) that adhere to the layout guidance.
The spheres image that requires creation of 7 subjects is particularly challenging for the baseline (BA) resulting in erroneous number of objects and blended colors.

Next, we compare generated images on DrawBench in \cref{fig:mainfig}.
\modelshort{} produces outputs with adherence to the prompt and layout as compared to other works.
The first example (black apple, green backpack) is notable as most methods generate a red apple while BA shows some leakage in the color across subjects.
The third example with 3 dogs and 2 cats is difficult and most methods get it wrong.
AttnRefocus has unnatural artifacts,
BoxDiff has wrong count and location,
LayoutGuidance generates questionable dogs,
R\&B generates only 2 subjects, and
BA generates people in the background.
While \modelshort{} (SD) also generates a person, with SDXL, our image is the only correct generation.
Additional qualitative results can be seen in \cref{fig:misc_outputs1} and \cref{fig:misc_outputs2}.
The former shows good outputs from \modelshort{} for multiple random seeds as compared to BA, while the latter presents several examples of complex prompts and layouts.
We also show generated images by varying the subject attributes across two dimensions in \cref{fig:color_variation}.

\input{figures/color_variation}

\paragraph{Quantitative comparison.}
\cref{tab:sota} shows that \modelshort{} outperforms all prior methods across the various tasks on DrawBench and two of three tasks on HRS.
We ensure fair comparisons and promote reproducibility by running all methods using the same random seeds.
We also present results while using comparable backbones, SD-1.5 and SDXL.
For \textul{spatial} prompts, \modelshort{} improves over the strongest baseline (GLIGEN) by +6.0\% on DrawBench and R\&B by +3.7\% on HRS.
A similar large improvement is seen in \textul{color} prompts with +17\% improvement on DrawBench and +3.8\% on HRS.
These gains highlight the effectiveness of the subject-attribute association losses in generating more accurate associations.
On the DrawBench \textul{counting} task, our method achieves the highest F1 score (0.88), matching the previous best baseline (R\&B) while outperforming others.
While \modelshort{} performs well on spatial, color, and counting metrics, we observe lower performance on the HRS size task, partly owing to inaccurate and confusing aspect ratios of guiding boxes.

Notably, \modelshort{} consistently outperforms Bounded Attention, our closest baseline, that uses the same SDXL backbone by a significant margin.
With SD-1.5 on DrawBench, \modelshort{} (\vs~BA) achieves
0.78 (+10\%) on spatial accuracy,
0.42 (+8\%) on color accuracy, and
a comparable F1 score of 0.85 (+1\%) on counting.
Similarly, on HRS with SD-1.5, \modelshort{} reaches 31.8\% (+0.9\%) on spatial accuracy and 40.4\% (+7.4\%) on color accuracy.

\input{tables/perceptual_fidelity}

\paragraph{FID scores.}
We compare perceptual quality of generated images on Drawbench for Vanilla SDXL, BA, and \modelshort{}.
\cref{tab:fid_scores} shows that all three models achieve comparable scores on perceptual fidelity.
In fact, \modelshort{} outperforms BA slightly and is comparable to SDXL.

\input{tables/sota-small}

\subsection{User Studies}
\label{subsec:userstudy}
We conduct two user studies to further validate and compare images generated by BA and \modelshort{}.

\paragraph{Average human ranking study.}
In the first study, we select all four challenging prompts visualized in \cref{fig:misc_outputs1} and generate outputs using \modelshort{} and BA for 10 random seeds (0-9).
Five independent (non-author) raters score each of the 80 images (4 prompts, 10 seeds, 2 methods) on a Likert scale of 1-5, yielding a total of 400 ratings.
The mean scores in \cref{tab:userstudy}a show that \modelshort{} consistently outperforms BA across all prompts.
Further, the standard deviations provide interesting insights.
For prompts 1, 3, and 4, BA shows low mean and low standard deviation, indicating poor outputs across most seeds.
For prompt 2, \modelshort{} shows high mean with low standard deviation, indicating consistently strong outputs across the seeds. 

\paragraph{Error analysis.}
In the second user study, we analyze the kind of errors exhibited in generated images.
Beyond the three main error types addressed in this work (see \cref{fig:teaser}),
we group all \textit{other} observed issues into the fourth category ``Other Errors''.
For the same 80 images, we ask five raters to identify whether an error type is visible in each image (400 ratings).
As seen in \cref{tab:userstudy}b, \modelshort{} exhibits errors in fewer images as compared to BA.

\input{tables/user_study}

\subsection{Ablation Study}
\label{subsec:exp:ablation}

\input{figures/loss_ablation}

We perform ablation experiments for different modules of our method and show their contributions.
Specifically, we assess the three new loss terms (see \cref{eq:final_loss}):
(i)~masked latent regularization,
(ii)~in-distribution alignment, and
(iii)~subject-attribution association.

\cref{tab:loss_ablation} presents results of including/removing different loss terms on DrawBench.
As expected, applying masked regularization ($\Lmask$) results in a performance drop (row 2).
While this term is effective at preventing background semantic leakage, it causes the latents to drift away from the prior distribution, a common issue in latent optimization methods.
In contrast, incorporating the KL alignment term leads to small, but consistent improvements across all DrawBench tasks (row 3), highlighting its role in stabilizing latent representations.
Interestingly, when both loss terms are applied together, the method achieves highest performance by simultaneously mitigating background leakage and stabilizing the latent optimization process (row 4).
Finally, including the attribute loss ($\Latt$) results in a significant improvement to color accuracy (row 5).

\input{tables/loss_ablations}

We show the qualitative impact of masked regularization, KL alignment loss, and components of our attribute loss, in \cref{fig:loss_ablation}.
Only using the masked latent regularization term effectively prevents background semantic leakage, but results in reduced visual fidelity, \eg~a black-and-white background (row 1) or a slightly cartoonish duck (row 2).
When combined with the KL alignment term, image quality improves while background semantic leakage continues to be suppressed.
However, attribute leakage is observed, \eg~the blended pink and blue chicken (row 1) and out-of-shape duck (row 2).
Among the subject-attribute association losses, only including one of the similarity or dissimilarity loss (columns e, f) fails to resolve attribute leakage.
In fact, we also see some subject leakage with the chicken having a dog's head (col f, row 1).
In contrast, our final formulation with all loss terms (masked regularization, KL alignment, and both similarity and dissimilarity subject-attribute losses), results in correct subject identities with accurate attributes bound to each entity (col g).
These qualitative observations support the quantitative results presented in \cref{tab:loss_ablation}.

\input{figures/limitations}

\subsection{Limitations}
\label{subsec:limitations}
While \modelshort{} is effective in compositional scene generation using layout guidance, it has certain limitations.
First, its performance is inherently constrained by the generative capabilities of SDXL, which may result in suboptimal images where SDXL faces challenges.
Second, while our method performs reasonably well with partially overlapping bounding boxes, it may produce images that do not adhere to the layout in cases involving fully overlapping bounding-boxes (\cref{fig:limitations_comparison}).
Atypical aspect ratios due to automatically generated bounding box layouts (\eg~on HRS) are challenging, but unlikely with real user interactions.
In column 1, our method fails to generate a small person in front of a large dog;
while in column 2, the banana appears larger than the specified bounding box.
However, in columns 3 and 4, our method performs reasonably well with partially overlapping boxes.
Future advancements in dealing with overlapping boxes may improve robustness in such cases.

%% file: figures/mainfig.tex
\begin{figure*}[t]
\centering
\small
DrawBench Color: A \textbf{\color{black} black apple} and a \textbf{\color{green} green backpack}. \\
\includegraphics[width=0.12\linewidth]{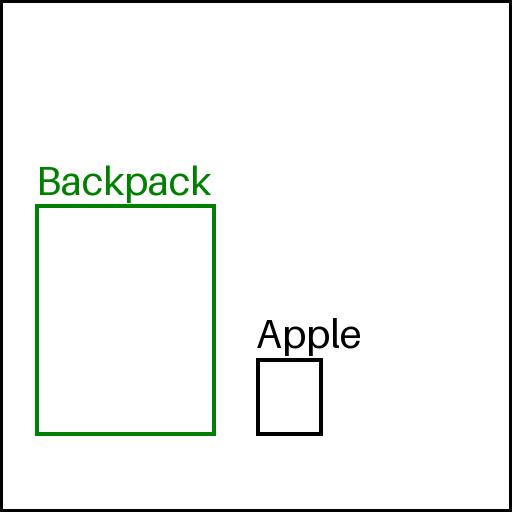}
\includegraphics[width=0.12\linewidth]{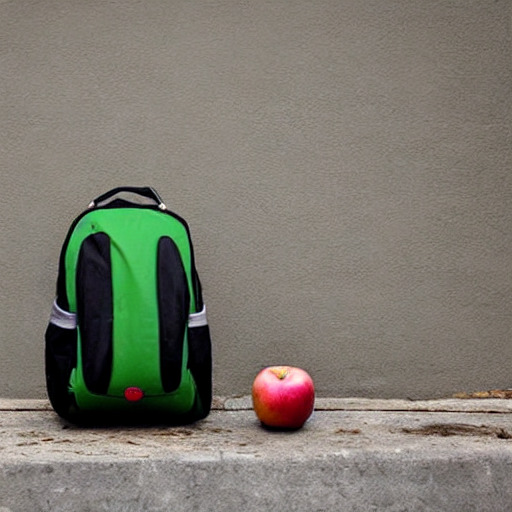}
\includegraphics[width=0.12\linewidth]{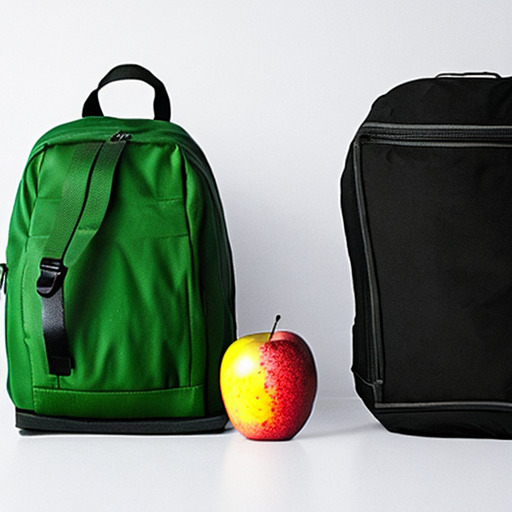}
\includegraphics[width=0.12\linewidth]{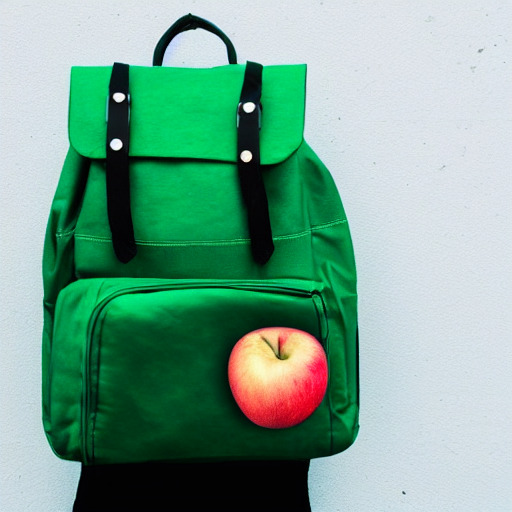}
\includegraphics[width=0.12\linewidth]{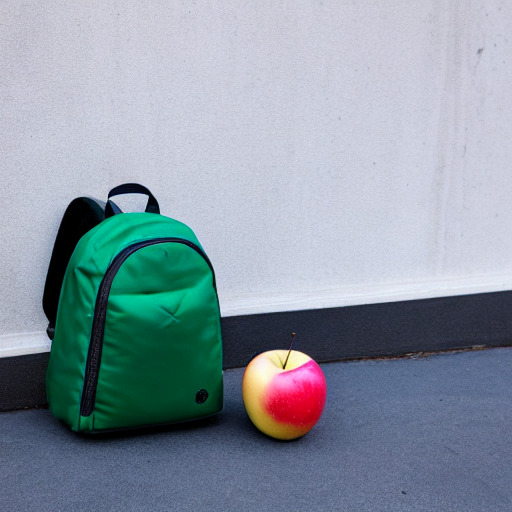}
\includegraphics[width=0.12\linewidth]{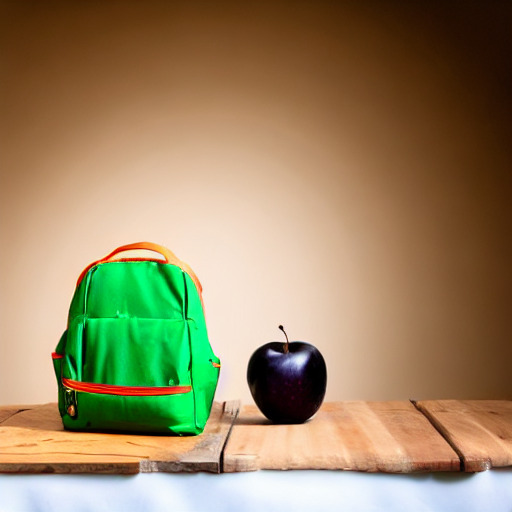}
\includegraphics[width=0.12\linewidth]{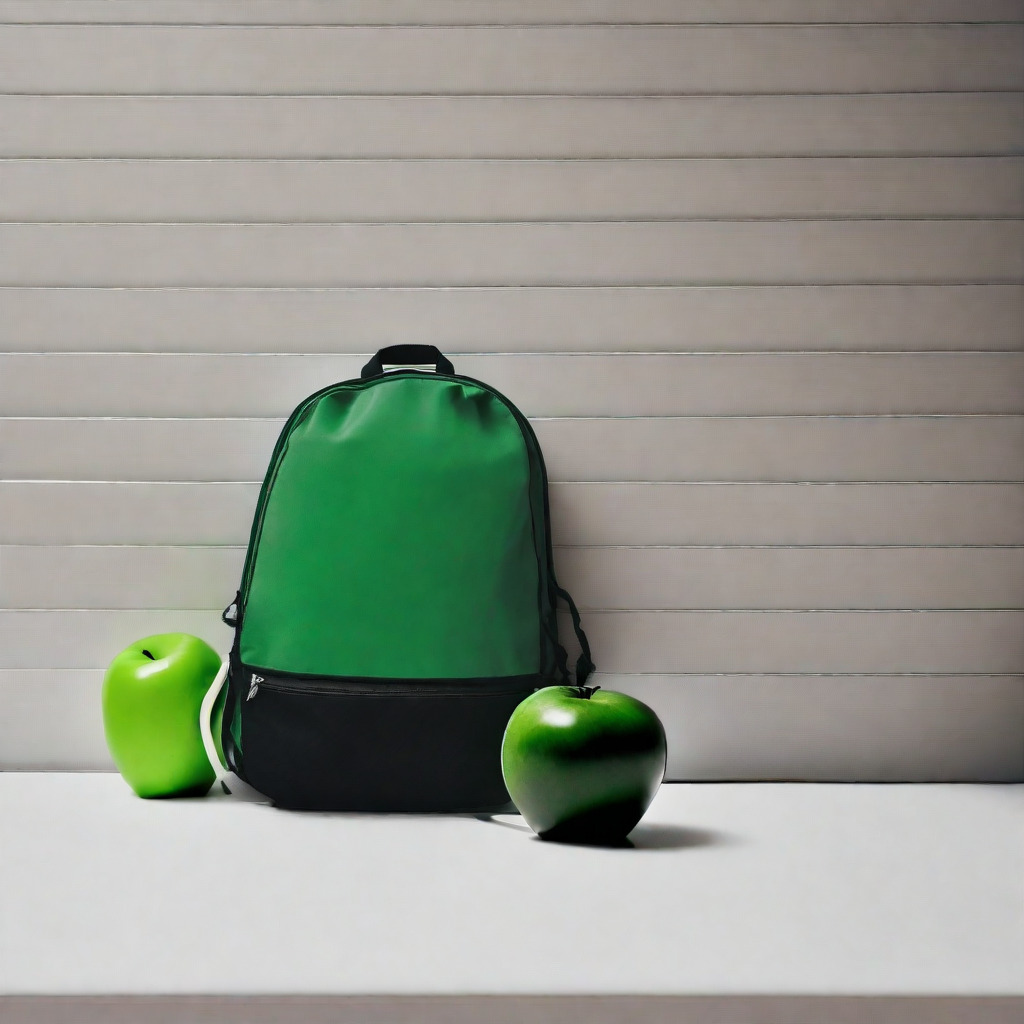}
\includegraphics[width=0.12\linewidth]{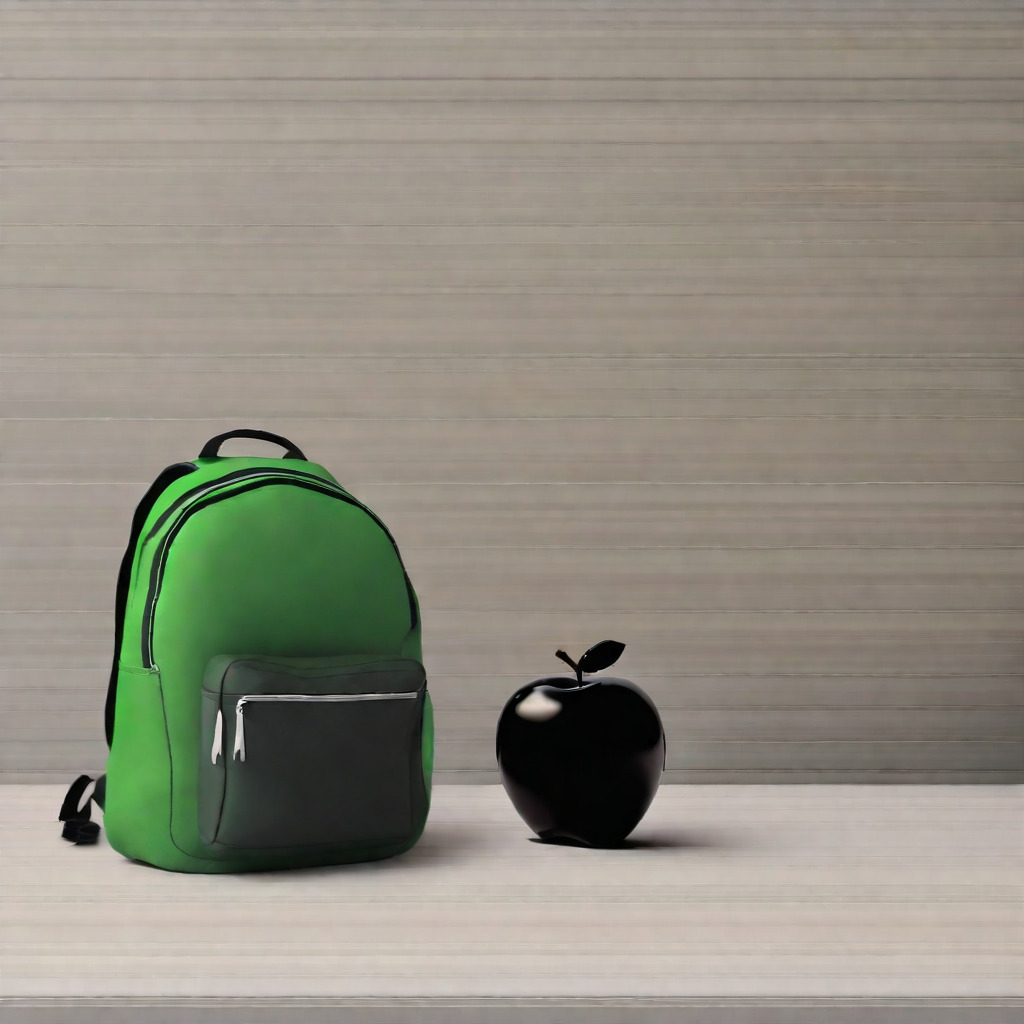} \\

DrawBench Color: A \textbf{\color{brown} brown bird} and a \textbf{\color{blue} blue bear}. \\
\includegraphics[width=0.12\linewidth]{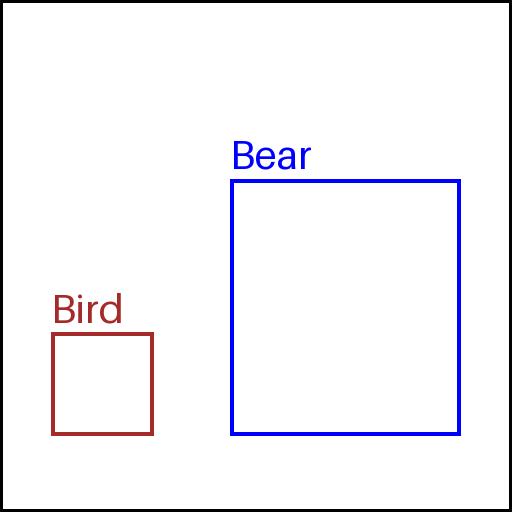}
\includegraphics[width=0.12\linewidth]{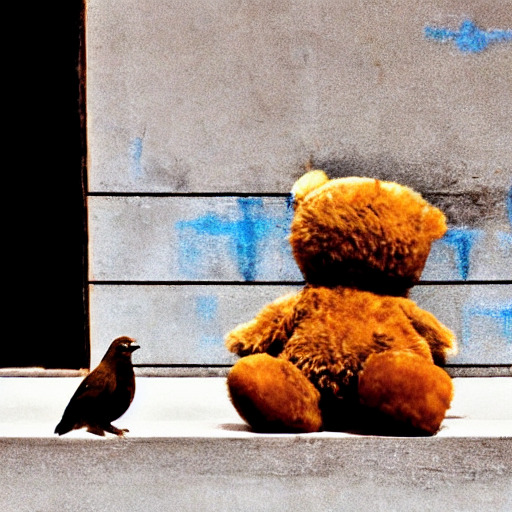}
\includegraphics[width=0.12\linewidth]{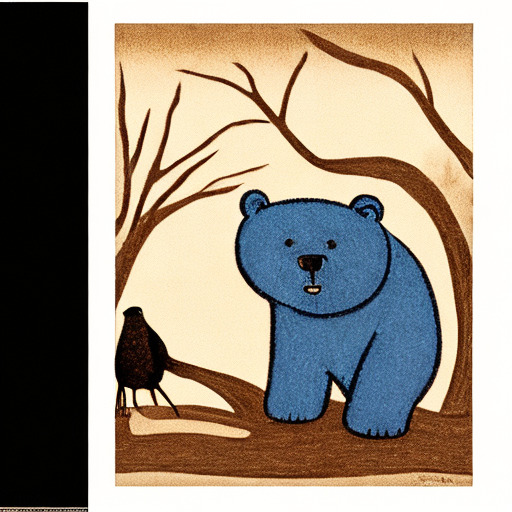}
\includegraphics[width=0.12\linewidth]{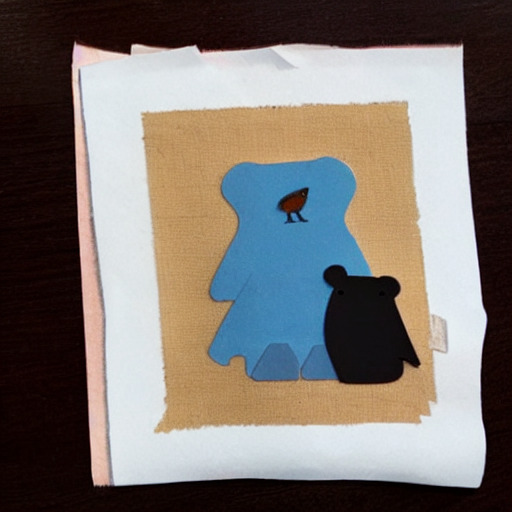}
\includegraphics[width=0.12\linewidth]{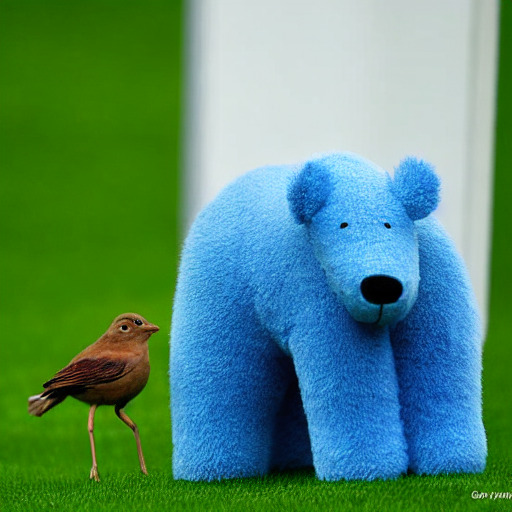}
\includegraphics[width=0.12\linewidth]{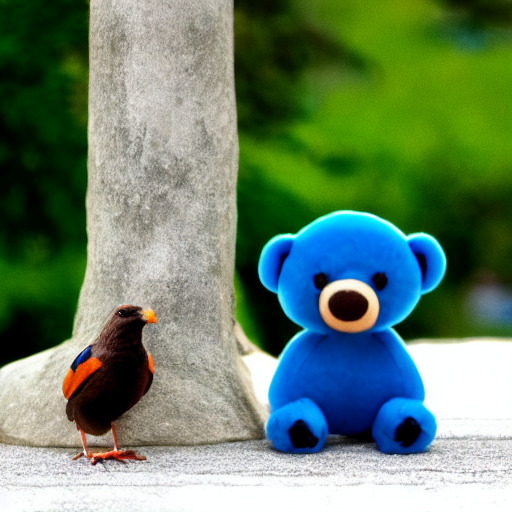}
\includegraphics[width=0.12\linewidth]{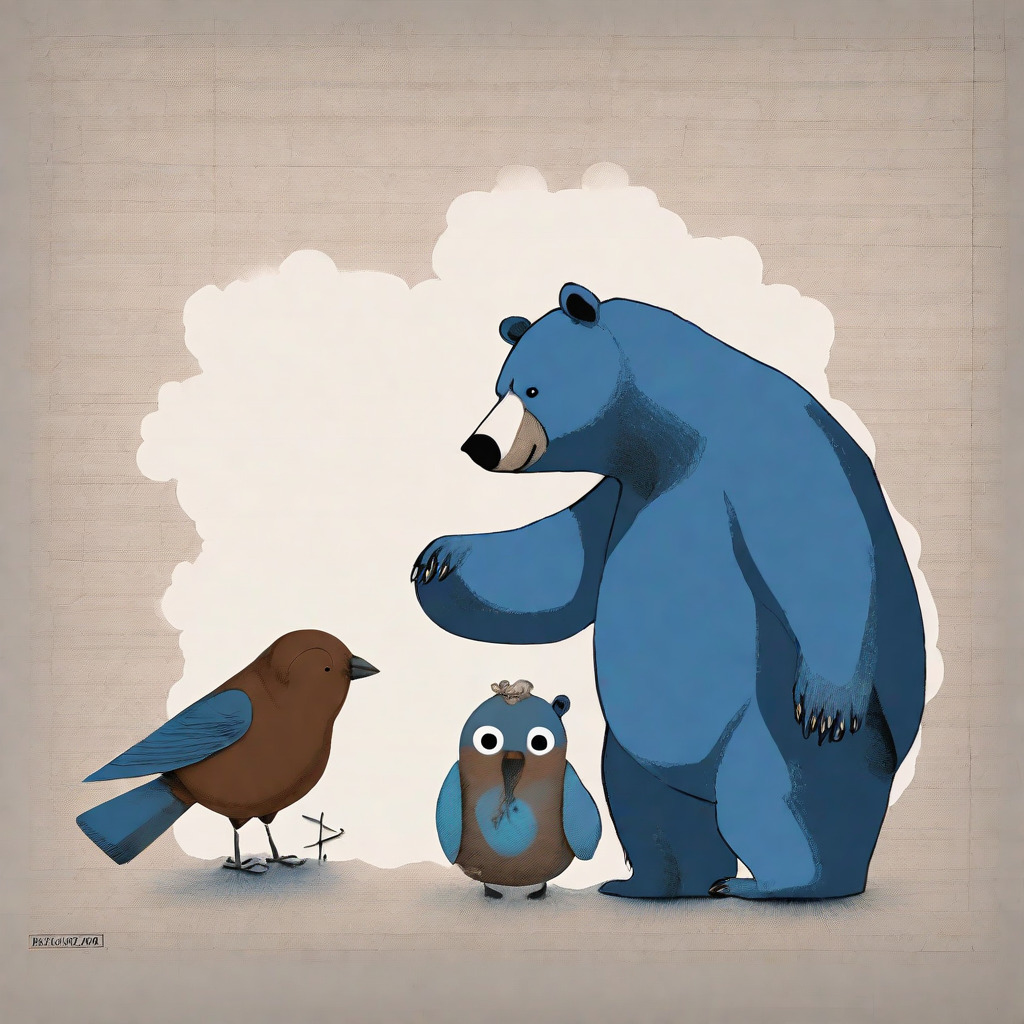}
\includegraphics[width=0.12\linewidth]{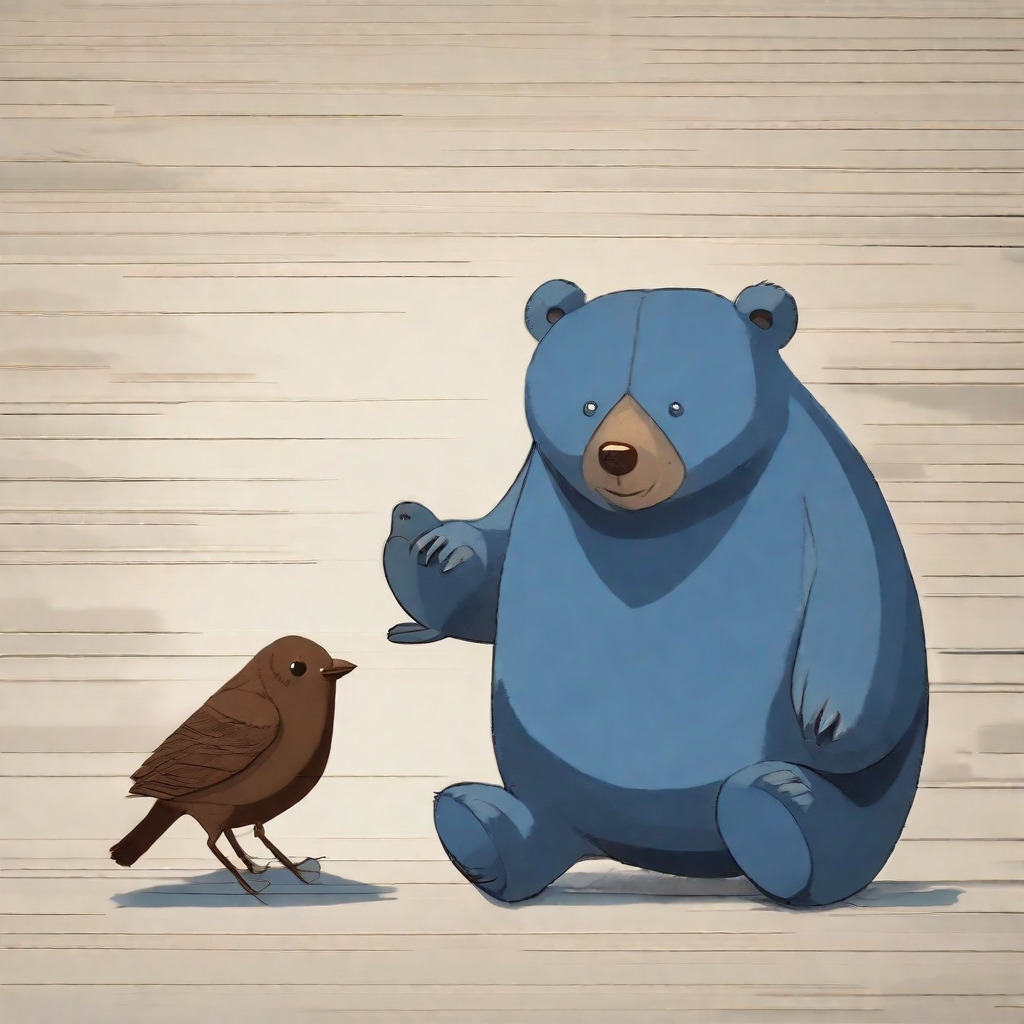} \\

DrawBench Counting: \textbf{\color{blue} Three cats} and \textbf{\color{red} two dogs} sitting on the grass. \\
\includegraphics[width=0.12\linewidth]{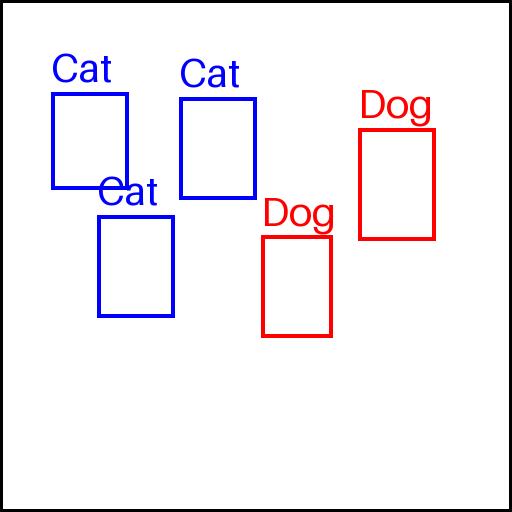}
\includegraphics[width=0.12\linewidth]{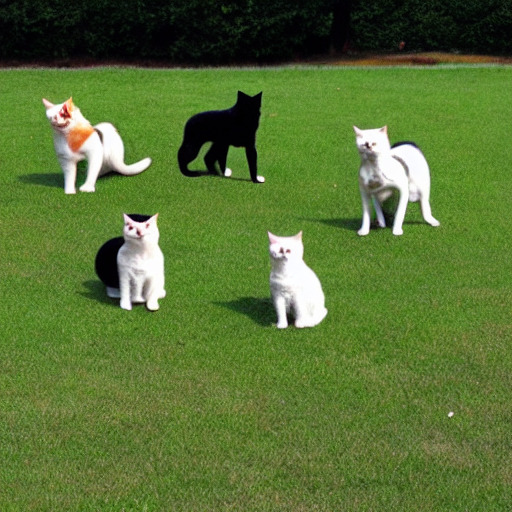}
\includegraphics[width=0.12\linewidth]{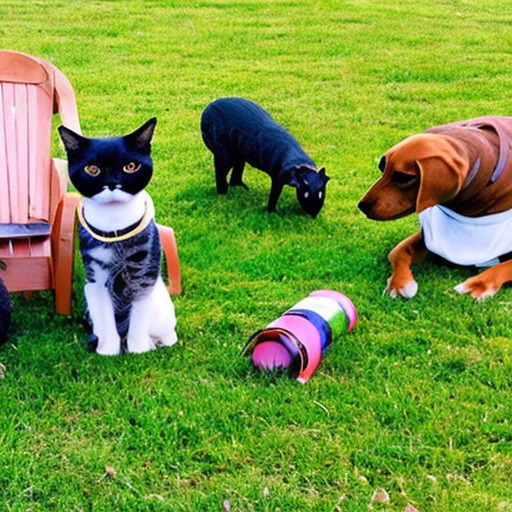}
\includegraphics[width=0.12\linewidth]{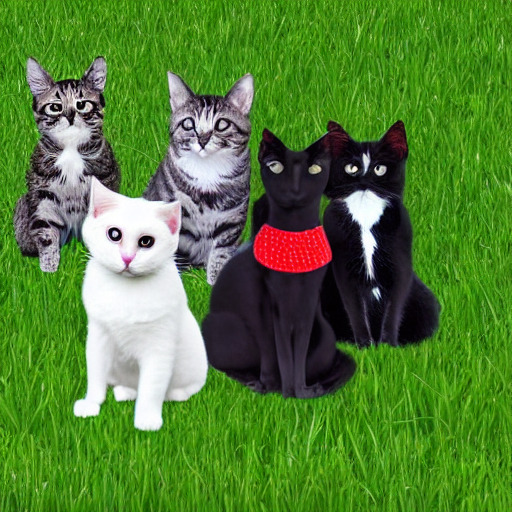}
\includegraphics[width=0.12\linewidth]{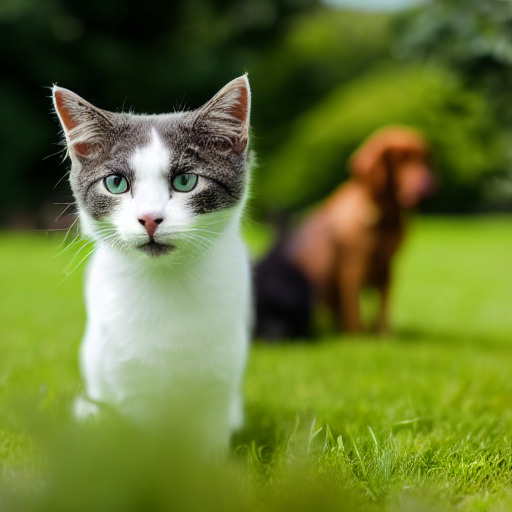}
\includegraphics[width=0.12\linewidth]{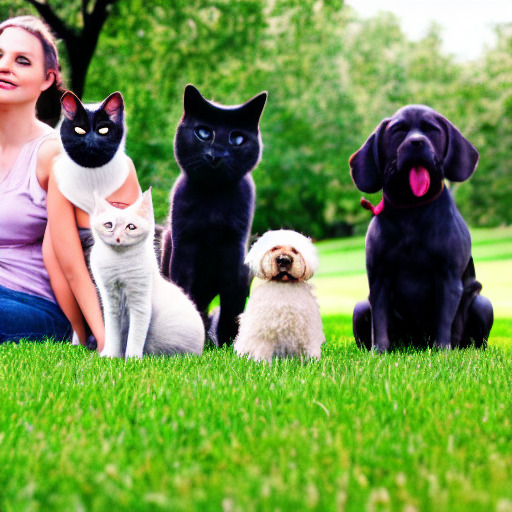}
\includegraphics[width=0.12\linewidth]{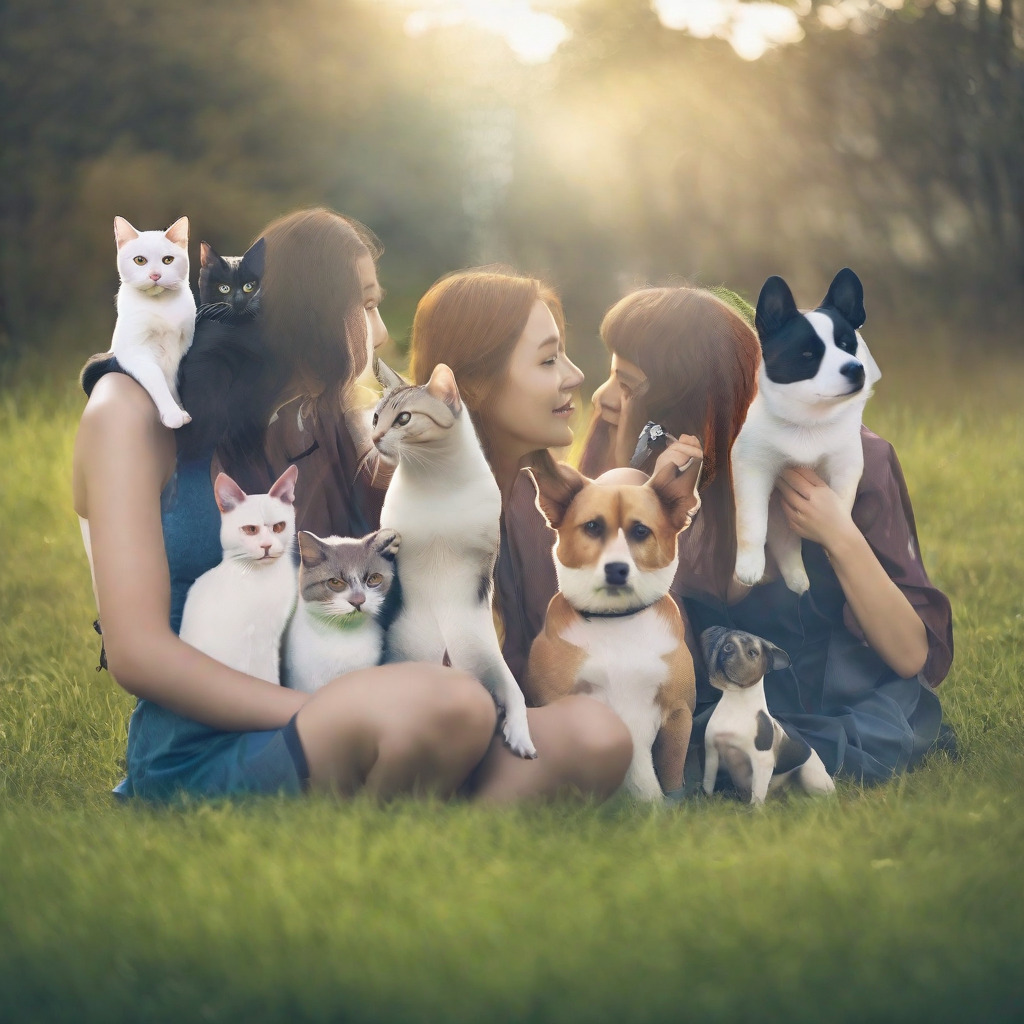}
\includegraphics[width=0.12\linewidth]{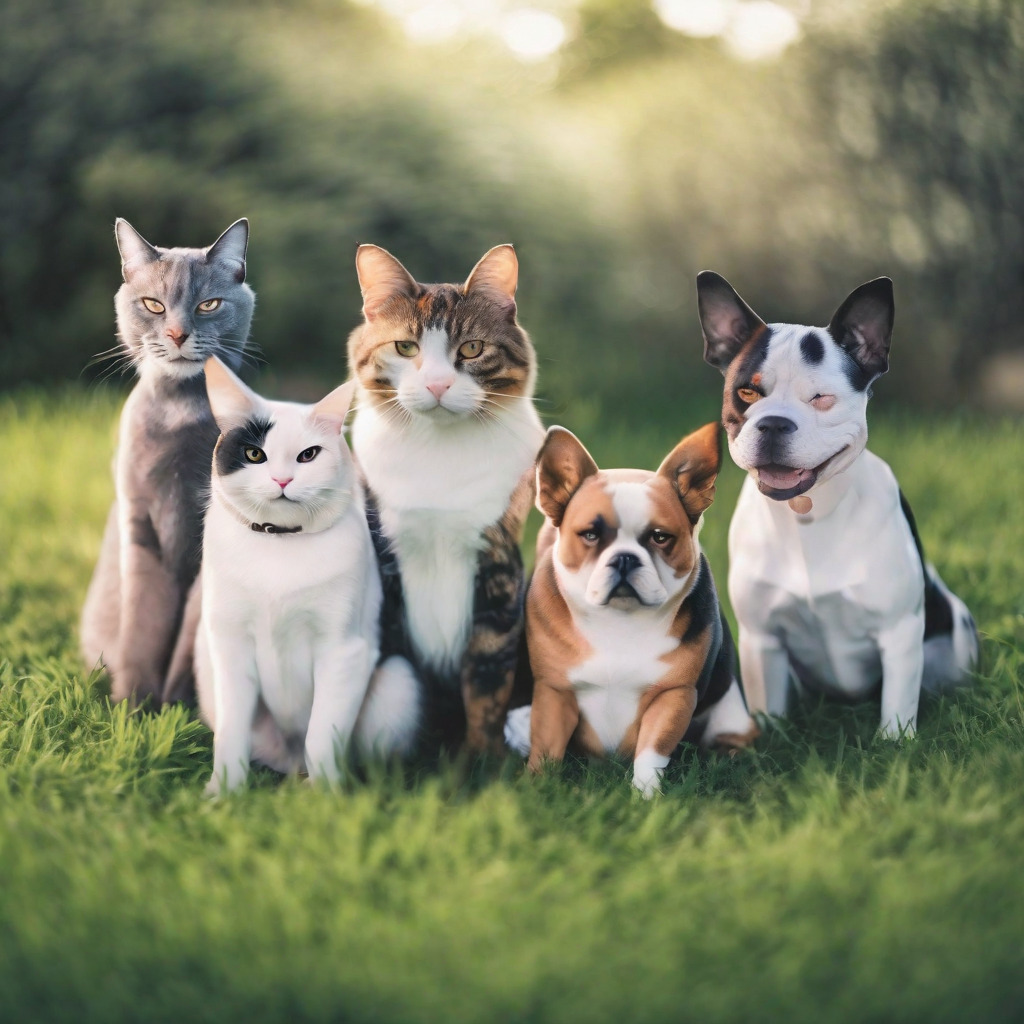} \\

DrawBench Spatial: A \textbf{\color{red} train} on \textbf{top} of an \textbf{\color{blue} surf board}. \\
\includegraphics[width=0.12\linewidth]{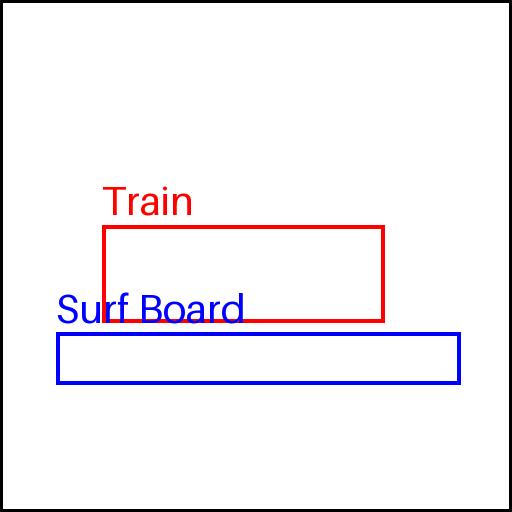}
\includegraphics[width=0.12\linewidth]{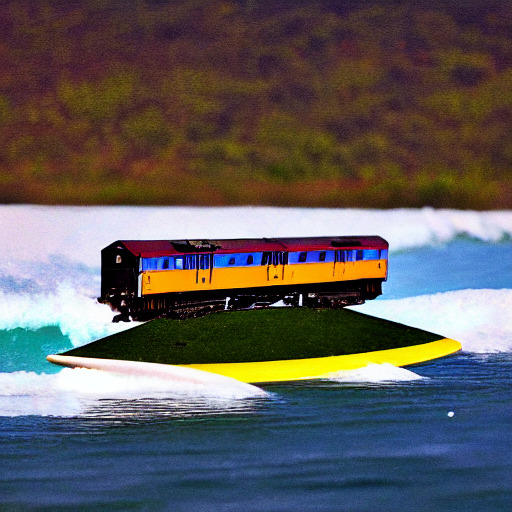}
\includegraphics[width=0.12\linewidth]{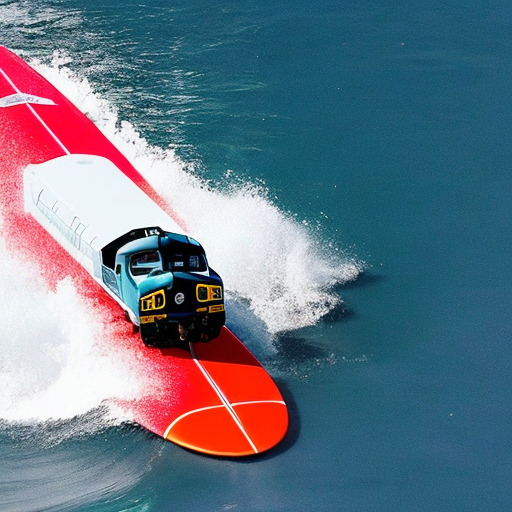}
\includegraphics[width=0.12\linewidth]{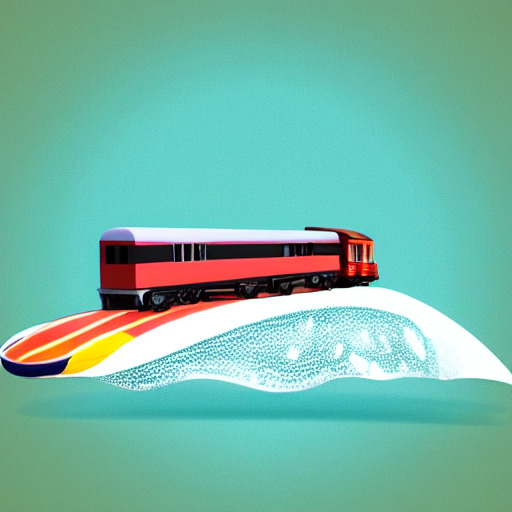}
\includegraphics[width=0.12\linewidth]{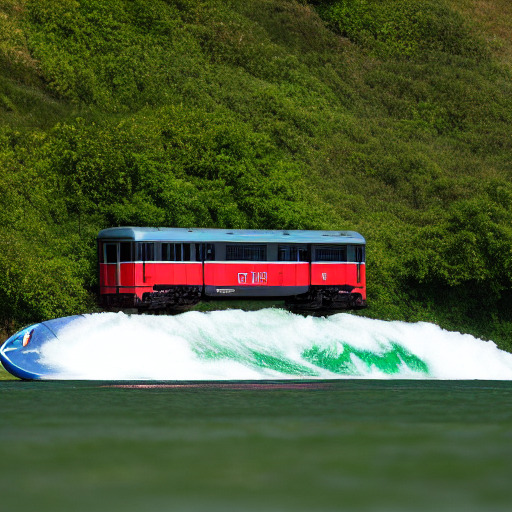}
\includegraphics[width=0.12\linewidth]{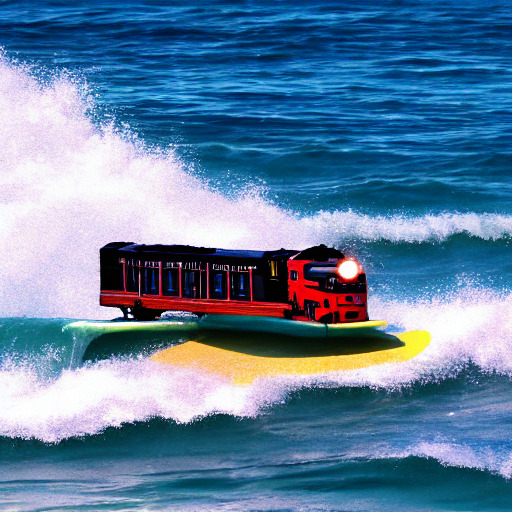}
\includegraphics[width=0.12\linewidth]{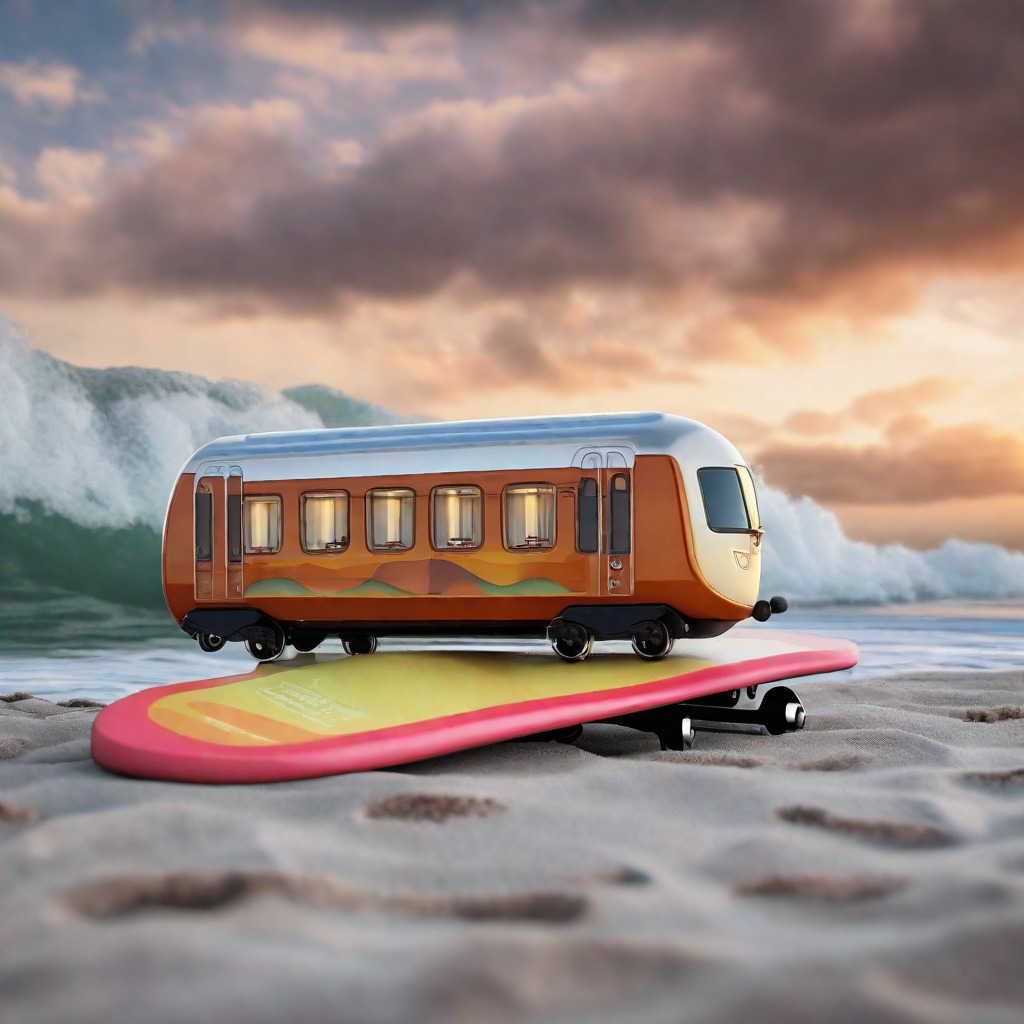}
\includegraphics[width=0.12\linewidth]{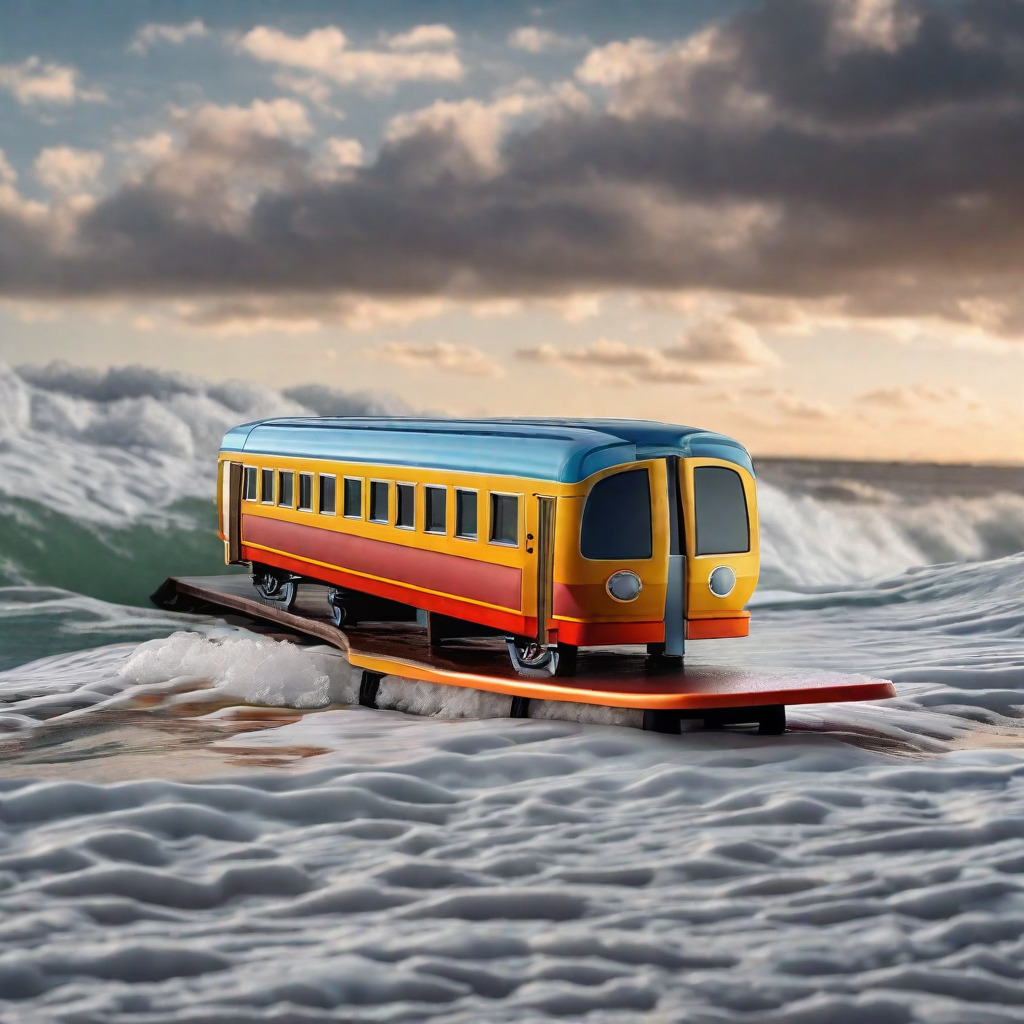} \\

DrawBench Spatial: A \textbf{\color{blue} sheep} to the \textbf{right} of a \textbf{\color{red} wine glass}. \\
\includegraphics[width=0.12\linewidth]{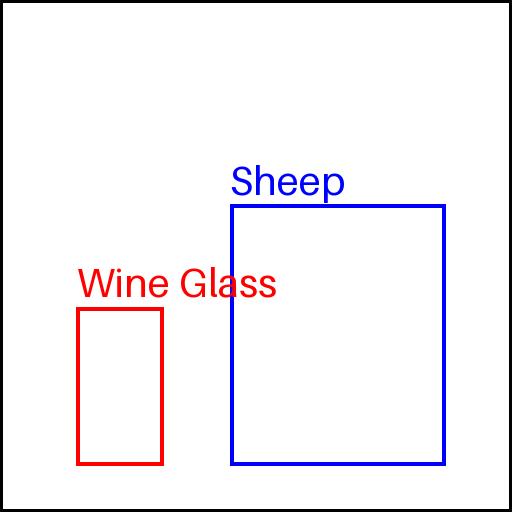}
\includegraphics[width=0.12\linewidth]{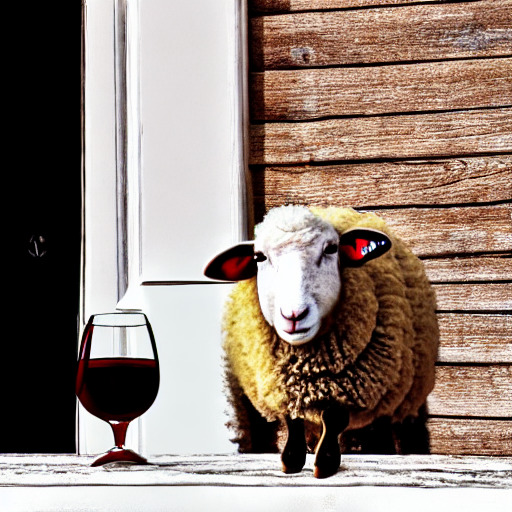}
\includegraphics[width=0.12\linewidth]{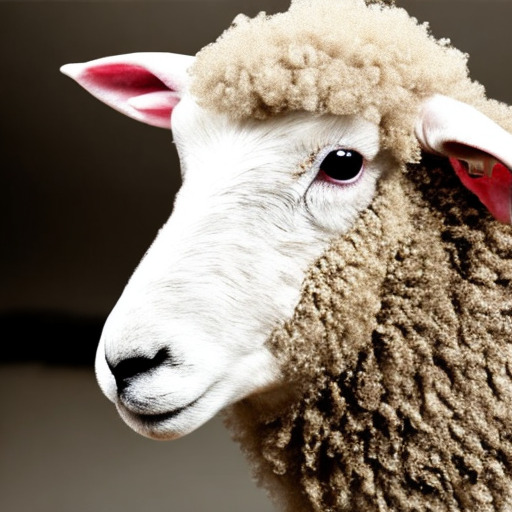}
\includegraphics[width=0.12\linewidth]{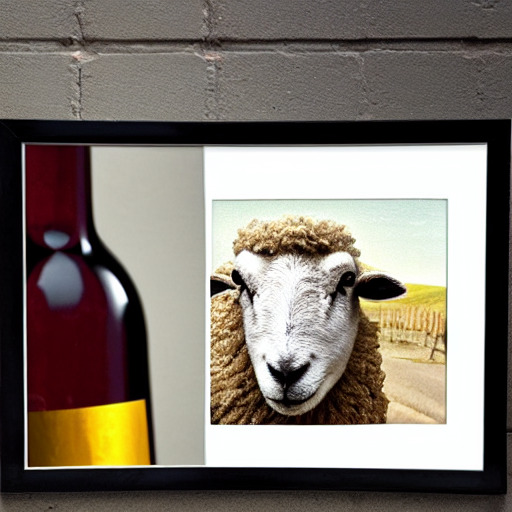}
\includegraphics[width=0.12\linewidth]{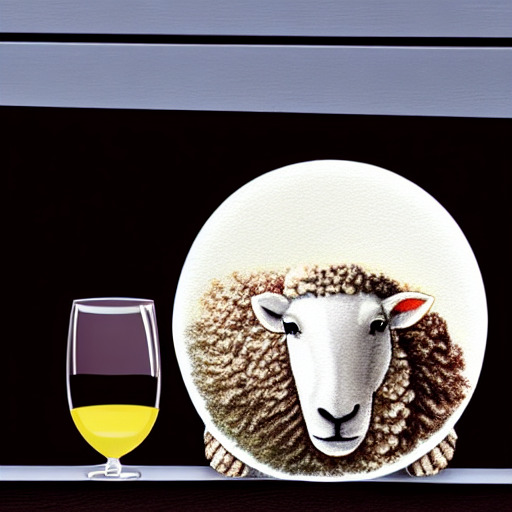}
\includegraphics[width=0.12\linewidth]{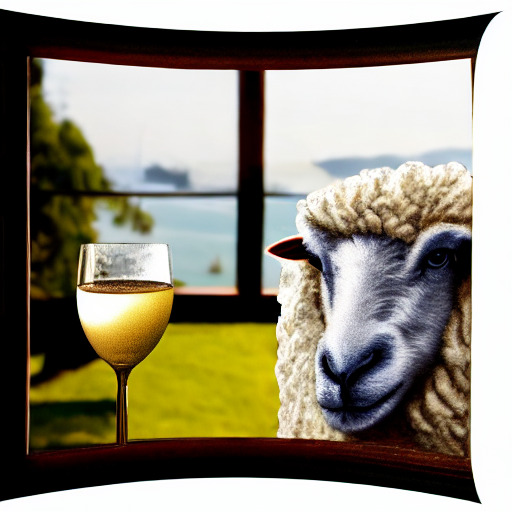}
\includegraphics[width=0.12\linewidth]{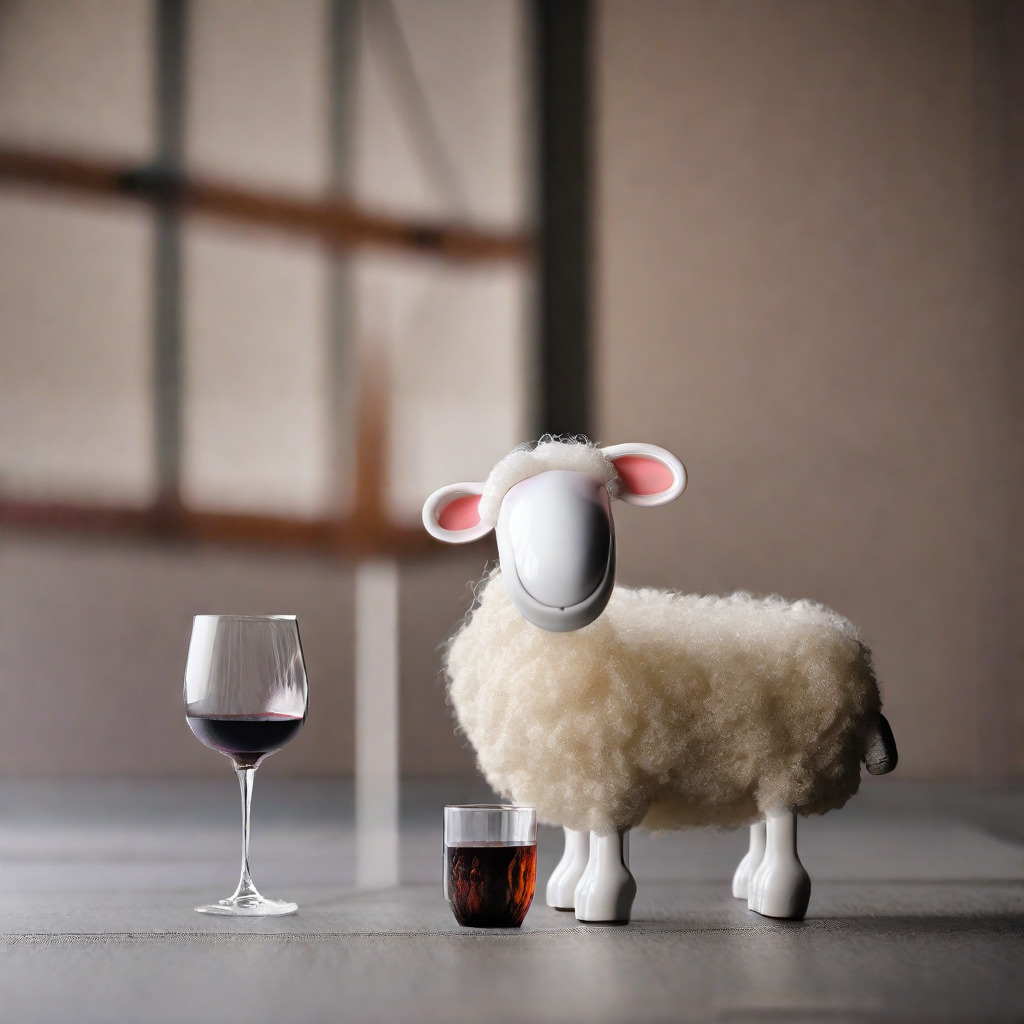}
\includegraphics[width=0.12\linewidth]{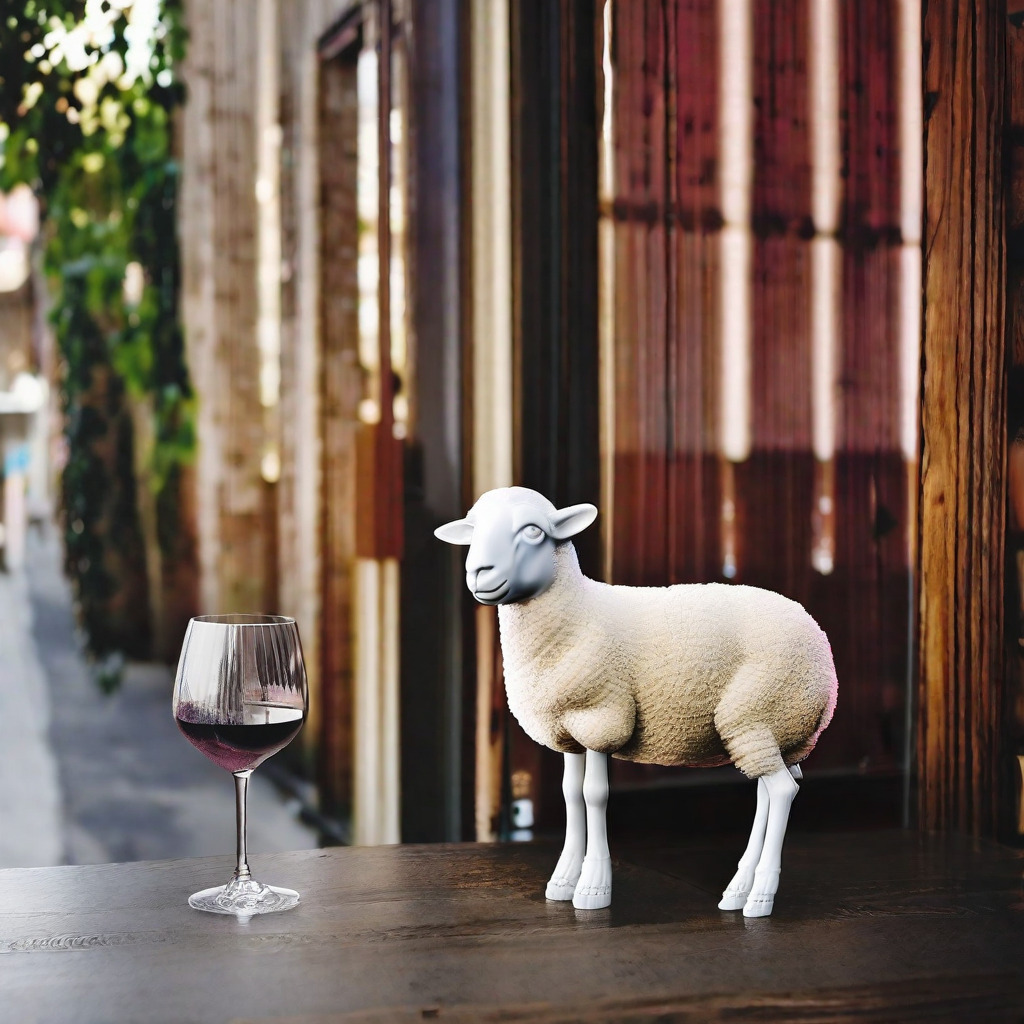} \\

\begin{tabularx}{\linewidth}{*{8}{>{\centering\arraybackslash}X}}
\centering
Layout &
AttnRefocus. &
BoxDiff &
LayoutGuidance &
R\&B &
\textbf{\modelshort{}} (SD) &
BA (SDXL) &
\textbf{\modelshort{}} (SDXL) \\
\end{tabularx}

\vspace{-3mm}
\caption{We compare generated images on DrawBench.
Each row uses the same random seed.
The text prompt is shown above and layout in column 1.
We compare \modelshort{} (SD) against:
Attention Refocusing~\cite{phung2024grounded},
BoxDiff~\cite{xie2023boxdiff},
Layout Guidance~\cite{chen2024training}, and
R\&B~\cite{xiao2024rnb}.
Bounded Attention (BA)~\cite{dahary2024yourself} with SDXL is compared against \modelshort{} (SDXL).
\modelshort{} shows strong adherence to the prompt (subjects, attributes, and layout), generates high quality images without background semantic leakage, and correctly localizes the subjects.
}
\vspace{-3mm}
\label{fig:mainfig}
\end{figure*}

%% file: figures/color_variation.tex
\begin{figure}[t]
\centering
\tabcolsep=0.05cm
\small
\begin{tabular}{lcccc}

\includegraphics[width=0.22\linewidth]{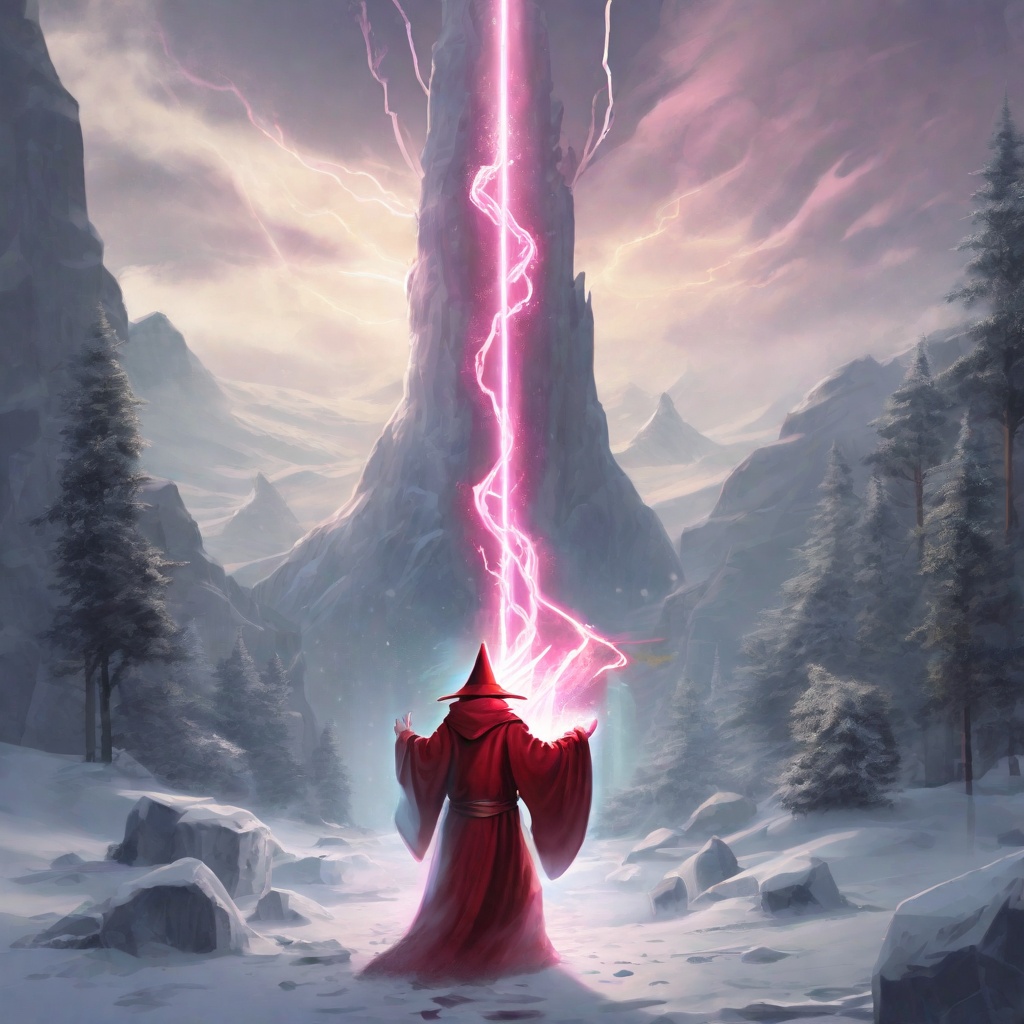} &
\includegraphics[width=0.22\linewidth]{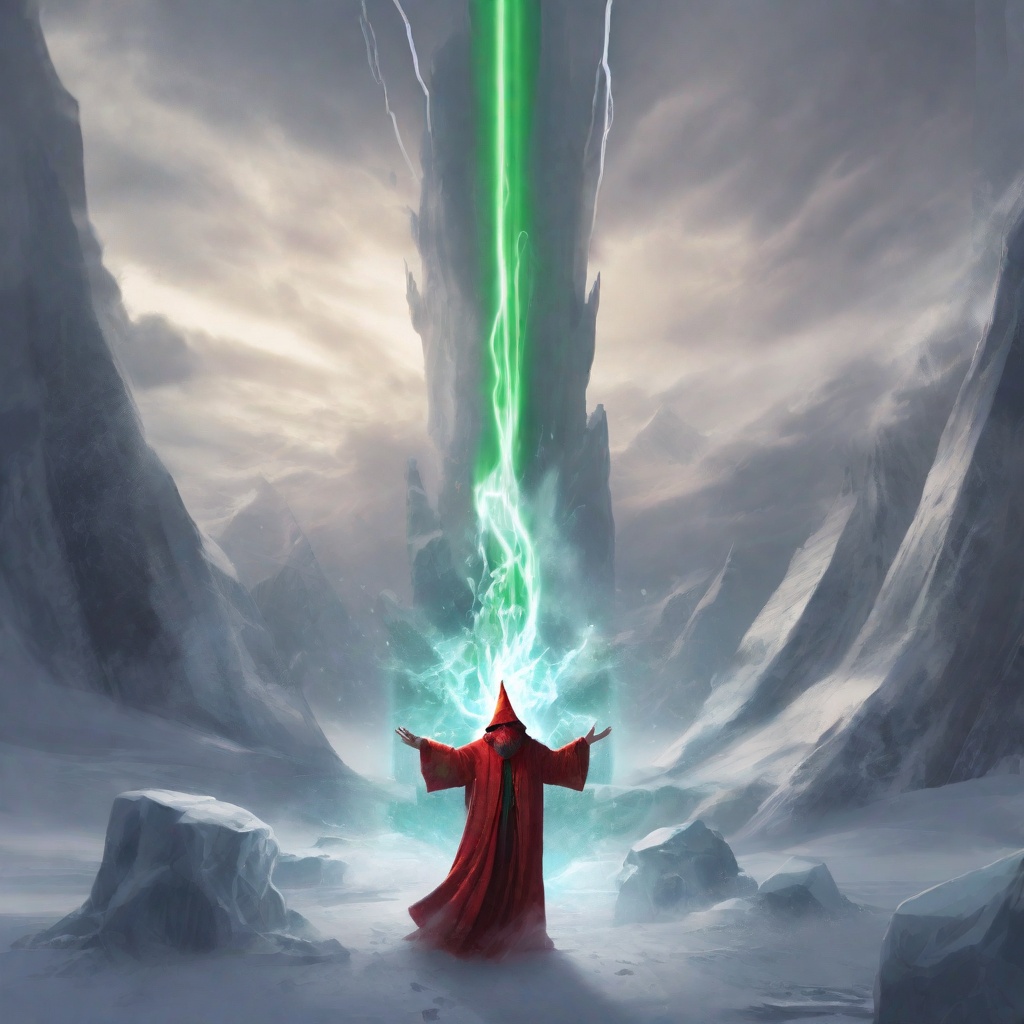} &
\includegraphics[width=0.22\linewidth]{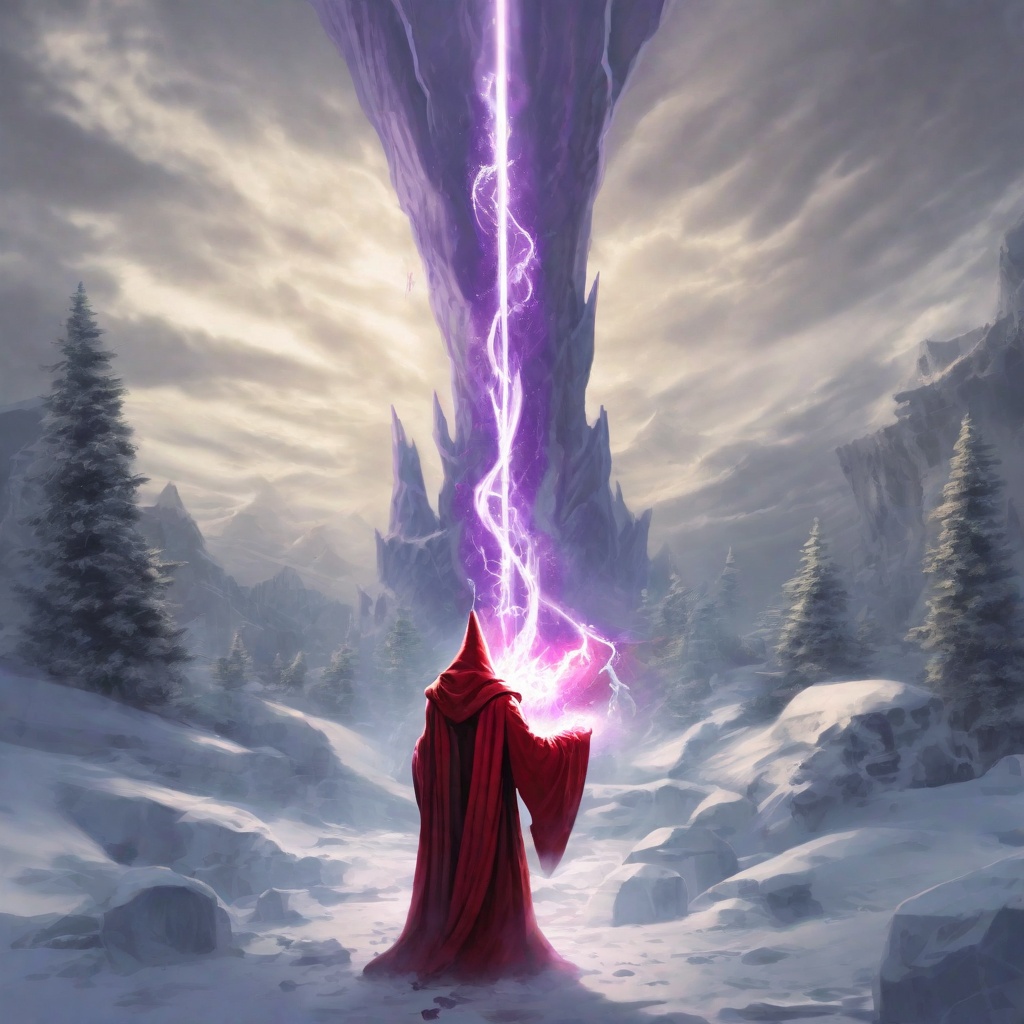} &
\includegraphics[width=0.22\linewidth]{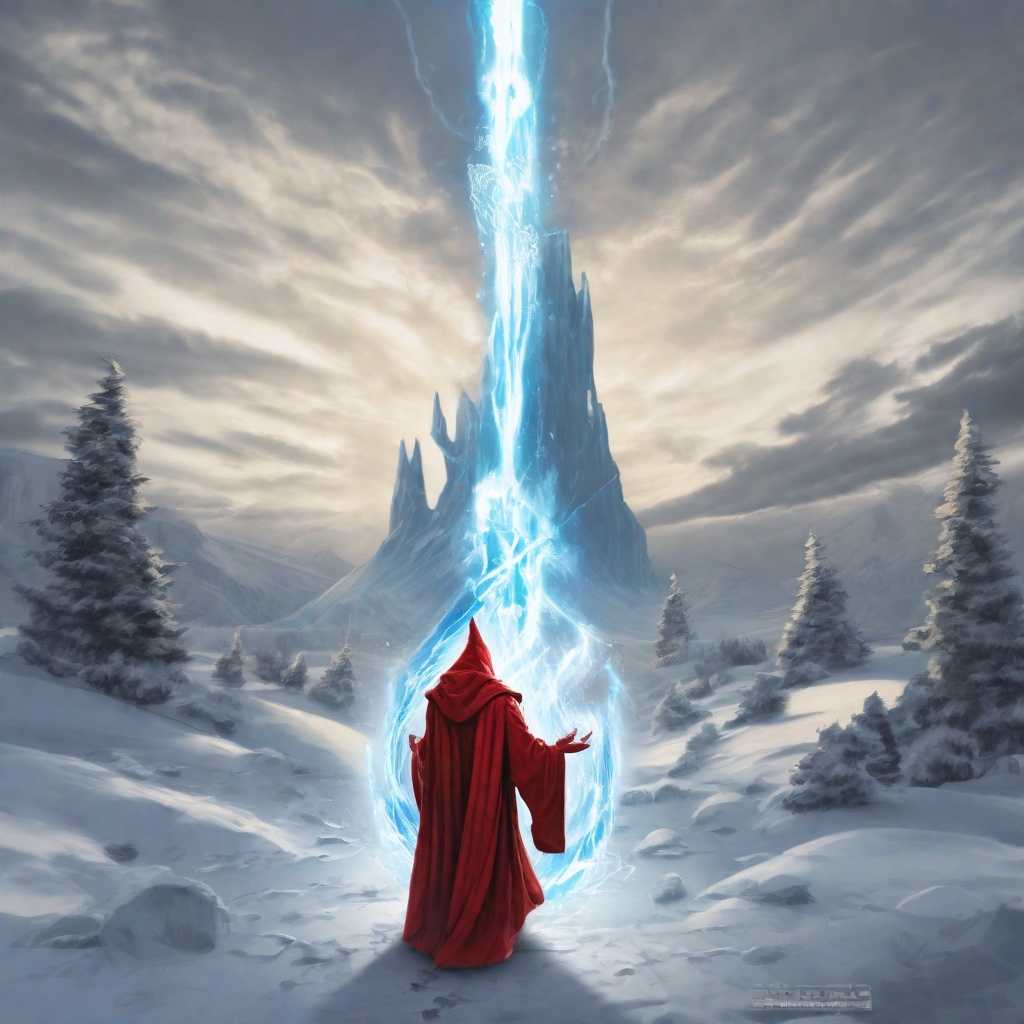} \\

\includegraphics[width=0.22\linewidth]{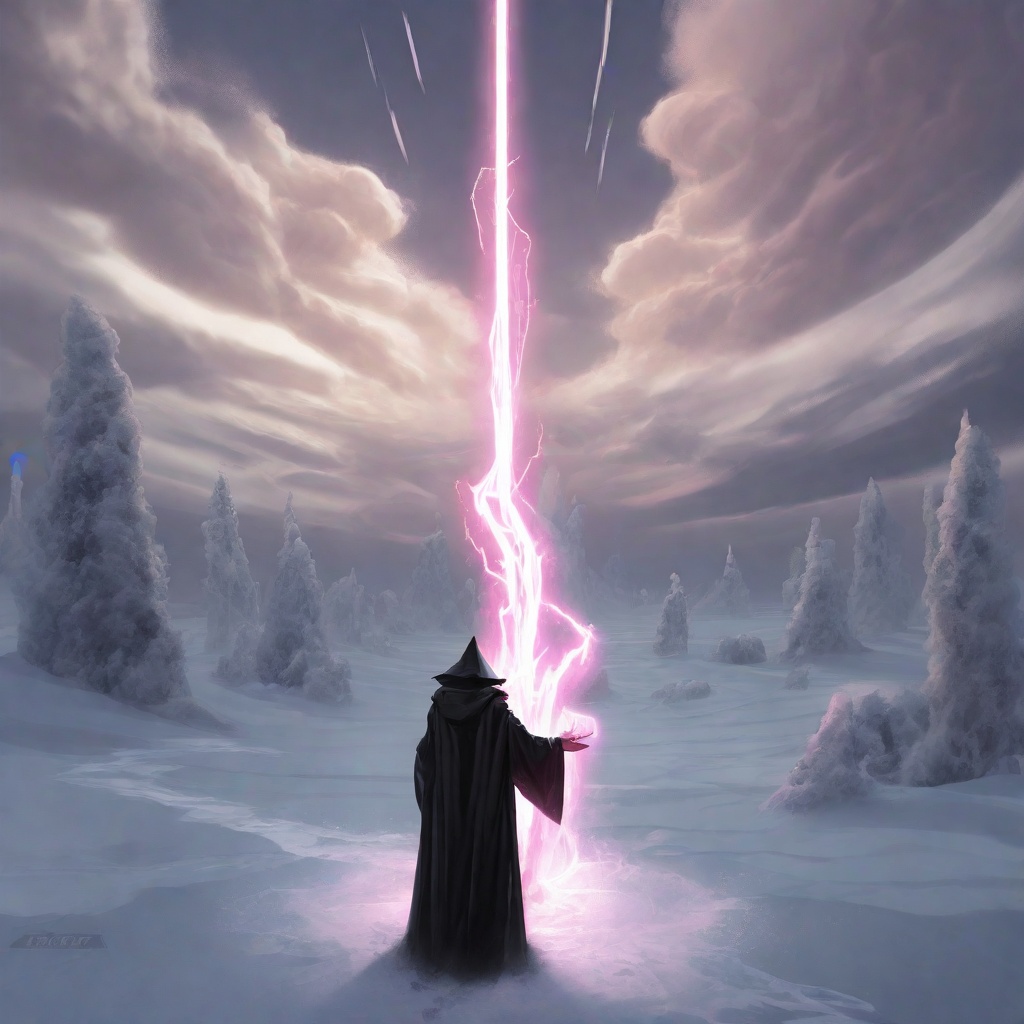} &
\includegraphics[width=0.22\linewidth]{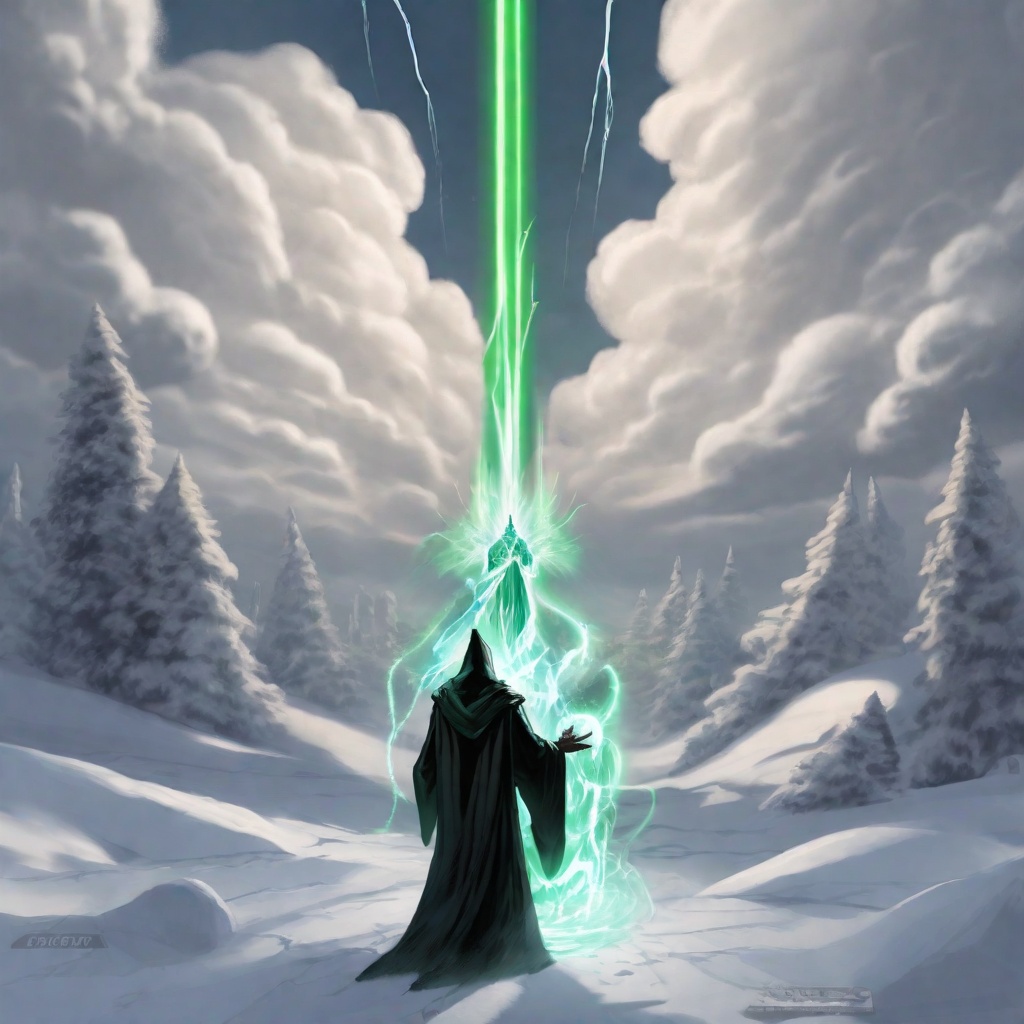} &
\includegraphics[width=0.22\linewidth]{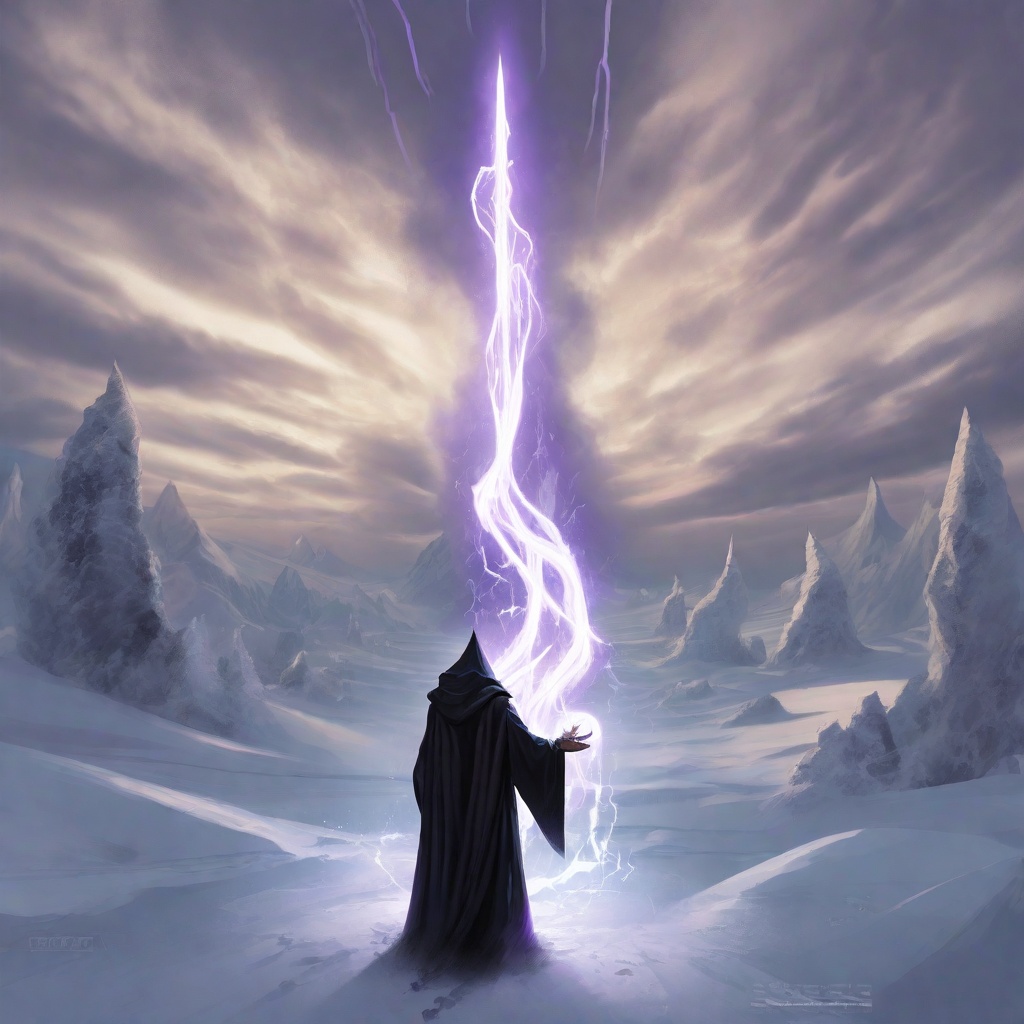} &
\includegraphics[width=0.22\linewidth]{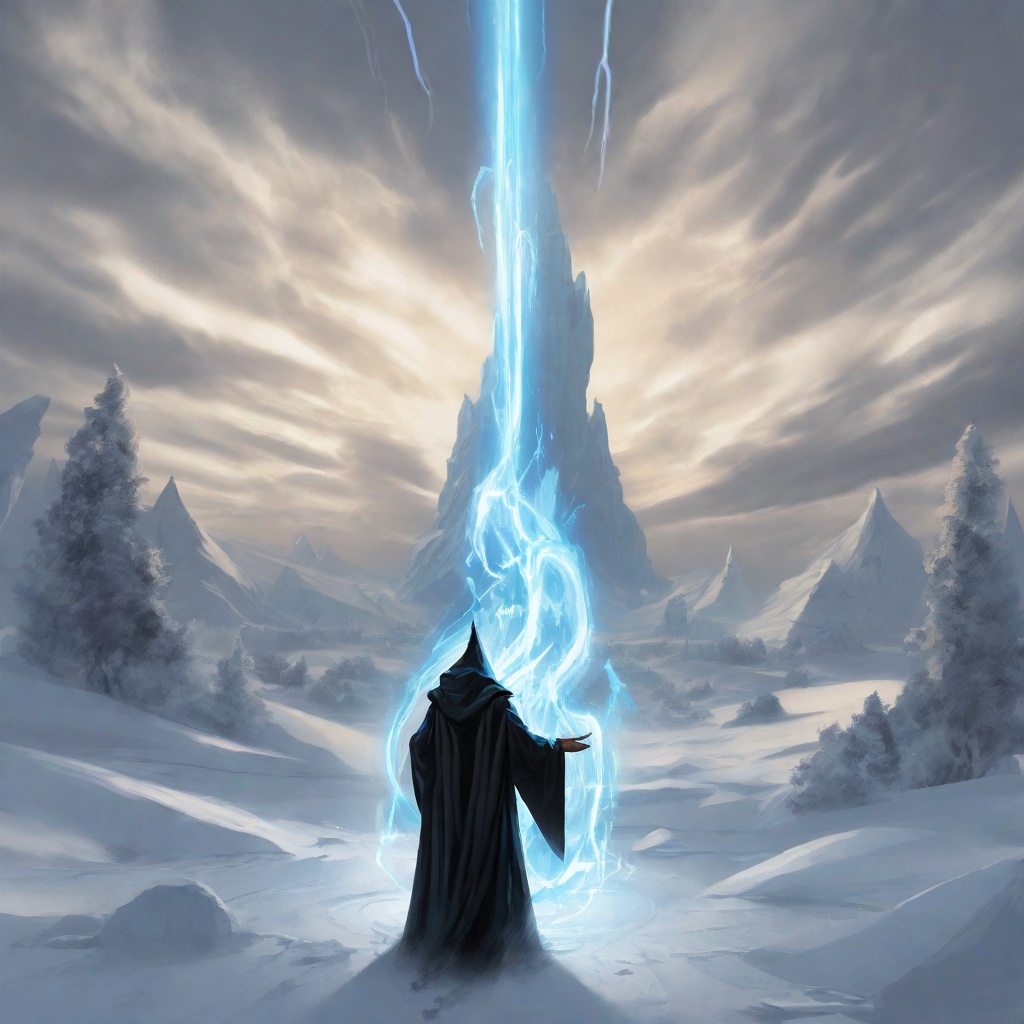} \\

\end{tabular}
\vspace{-3mm}
\caption{
\textbf{Color variation} across the wizard and the thunderbolt for the prompt: \textit{A concept art of an icy landscape with a \{red | black\} robe wizard summoning a \{pink | green | purple | blue\} colored magic thunderbolt from air}.
All images are generated with the same random seed.
}
\label{fig:color_variation}
\vspace{-4mm}
\end{figure}

%% file: tables/perceptual_fidelity.tex
\setlength{\columnsep}{10pt}
\setlength{\intextsep}{3pt}
\begin{wraptable}{r}{4.0cm}
\vspace{-1mm}
\centering
\small
\tabcolsep=0.10cm
\caption{FID scores on DrawBench for Vanilla SDXL, BA, and MALeR.}
\label{tab:fid_scores}
\vspace{-4mm}

\begin{tabular}{lccc}
\toprule
 & SDXL & BA & \cellcolor{yellow!20}MALeR \\

FID $(\downarrow)$ & 161.03 & 164.59 & \cellcolor{yellow!20}163.02 \\
\bottomrule

\end{tabular}
\end{wraptable}

%% file: tables/sota-small.tex
\begin{table}[t]
\tabcolsep=0.12cm
\small
\centering
\caption{
We report results on DrawBench (Spatial, Color, Counting) and HRS Benchmark (Spatial, Color, Size).
Baseline acronyms are:
AttnRef: Attention Refocusing,
LGuidance: Layout Guidance, and
BoundAttn: Bounded Attention (equivalent of only $\Liou$ loss).
ReCo with *, fine-tunes SD.
Except HRS Size (some layout challenges), \modelshort{} achieves best performance.
}
\label{tab:sota}
\vspace{-3mm}
\begin{tabular}{lc ccccc p{0cm} ccc}
\toprule
\multirow{3}{*}{Method} & \multirow{3}{*}{Base} & \multicolumn{5}{c}{DrawBench} & & \multicolumn{3}{c}{HRS Benchmark} \\
\cmidrule(lr){3-7} \cmidrule(lr){9-11}
& &
Spat. & Col. & \multicolumn{3}{c}{Counting} & &
Spat. & Col. & Size \\
&           & Acc. & Acc. & P    & R    & F1   & & Acc.  & Acc.  & Acc.  \\
\midrule
GLIGEN %
& SD1.4     & 0.75 & 0.12 & 0.77 & 0.78 & 0.77 & & 27.7 & 15.8 & 66.5  \\
ReCo %
& SD1.4*    & 0.70 & 0.19 & 0.75 & 0.96 & 0.84 & & 25.9 & 20.0 & 75.5  \\
AttnRef. %
& SD1.4     & 0.74 & 0.31 & 0.80 & 0.79 & 0.79 & & 32.0 & 31.3 & 72.5  \\
LGuidance %
& SD1.5     & 0.65 & 0.31 & 0.83 & 0.75 & 0.79 & & 15.9 & 17.4 & 60.5  \\
R\&B %
& SD1.5     & 0.68 & 0.35 & 0.94 & 0.82 & \textbf{0.88} & & 34.0 & 34.3 & \textbf{77.8}  \\
BoxDiff %
& SD2.1     & 0.66 & 0.26 & 0.92 & 0.77 & 0.84 & & 21.0 &  1.9 & 74.9  \\

BoundAttn %
& SD1.5      & 0.68 & 0.34 & 0.86 & 0.82 & 0.84 & & 30.9 & 33.0 & 57.5  \\
\rowcolor{yellow!20}
\modelshort{} %
& SD1.5      & 0.78 & 0.42 & 0.89 & 0.81 & 0.85 & & 31.8 & 40.4 & 56.3  \\
\midrule
BoundAttn %
& SDXL      & 0.69 & 0.44 & 0.74 & 0.95 & 0.83 & & 33.9 & 37.5 & 64.7  \\

\rowcolor{yellow!20}
\modelshort{} %
& SDXL      & \textbf{0.81} & \textbf{0.61} & 0.81 & 0.96 & \textbf{0.88} & & \textbf{37.7} & \textbf{41.3} & 59.9  \\
\bottomrule
\end{tabular}
\vspace{-3mm}
\end{table}

%% file: tables/user_study.tex
\begin{table}[t]
\centering
\small
\tabcolsep=0.14cm
\caption{
User study results: We conduct user studies to compare BA and \modelshort{}.
(a)~Average Human Ranking (AHR) on four challenging prompts with mean Likert scores (1–5); images shown in \cref{fig:misc_outputs1}.
(b)~Fraction of images showing error-types: Background Semantic Leakage (BSL), Out of Distribution (OOD), Attribute Leakage (AL), and Other Errors (OE).
The user studies confirm that \modelshort{}'s outputs are better than BA.}
\label{tab:userstudy}
\vspace{-2mm}

(a)~Average Human Ranking

\begin{tabular}{l cccc c}

\toprule
Method & Prompt 1 & Prompt 2 & Prompt 3 & Prompt 4 & Avg. \\
\midrule
BA & 2.12{\scriptsize$\pm$0.688} & 2.50{\scriptsize$\pm$1.047} & 1.84{\scriptsize$\pm$0.506} & 2.14{\scriptsize$\pm$0.584} & 2.15 \\
\rowcolor{yellow!20}
MALeR & \textbf{3.44}{\scriptsize$\pm$0.837} & \textbf{3.92}{\scriptsize$\pm$0.736} & \textbf{2.26}{\scriptsize$\pm$0.769} & \textbf{2.90}{\scriptsize$\pm$0.796} & \textbf{3.13} \\
\bottomrule
\end{tabular}

\vspace{2mm}

(b)~Error-Type Analysis

\begin{tabular}{l cccc}

\toprule
Method & BSL & OOD & AL & OE \\
\midrule
BA & 0.445 & 0.305 & 0.590 & 0.325 \\
\rowcolor{yellow!20}
MALeR & 0.240 & 0.230 & 0.400 & 0.185 \\
Relative-Improvement $\Delta$ & \textbf{+46\%} & \textbf{+25\%} & \textbf{+32\%} & \textbf{+43\%} \\
\bottomrule
\end{tabular}

\vspace{-4mm}
\end{table}

%% file: figures/loss_ablation.tex
\begin{figure*}[t]
\centering
\small
A realistic photo of a \textbf{\color{pink} pink chicken} and a \textbf{\color{blue} blue dog} \\
\includegraphics[width=0.138\linewidth]{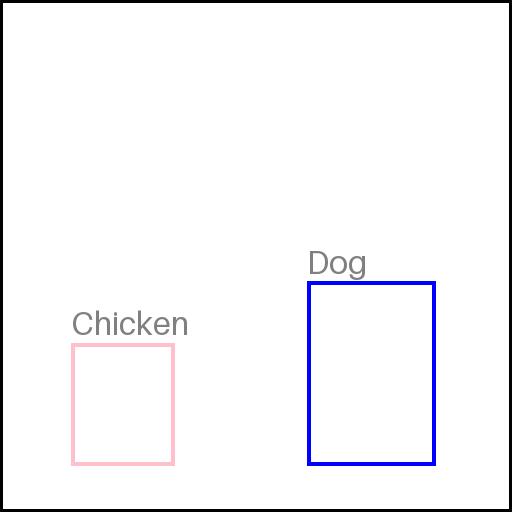} \hfill
\includegraphics[width=0.138\linewidth]{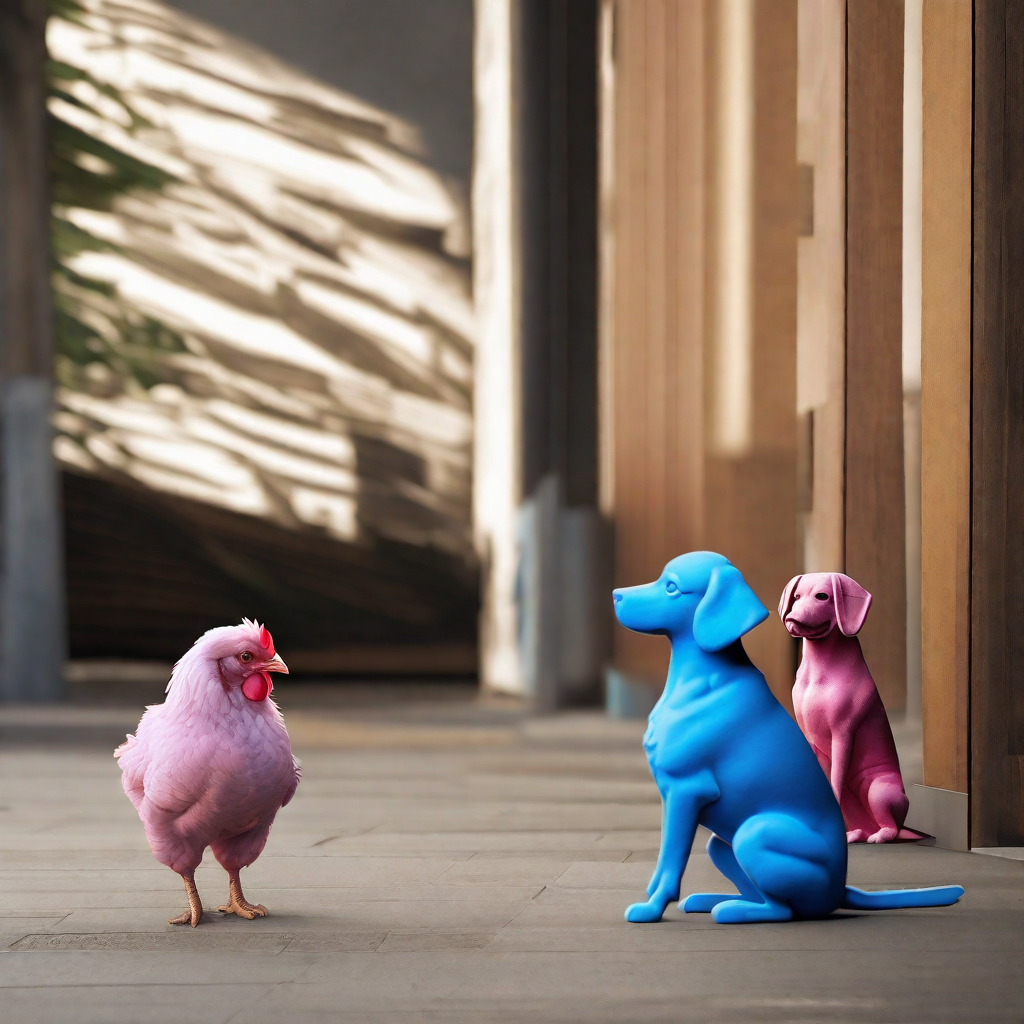} \hfill
\includegraphics[width=0.138\linewidth]{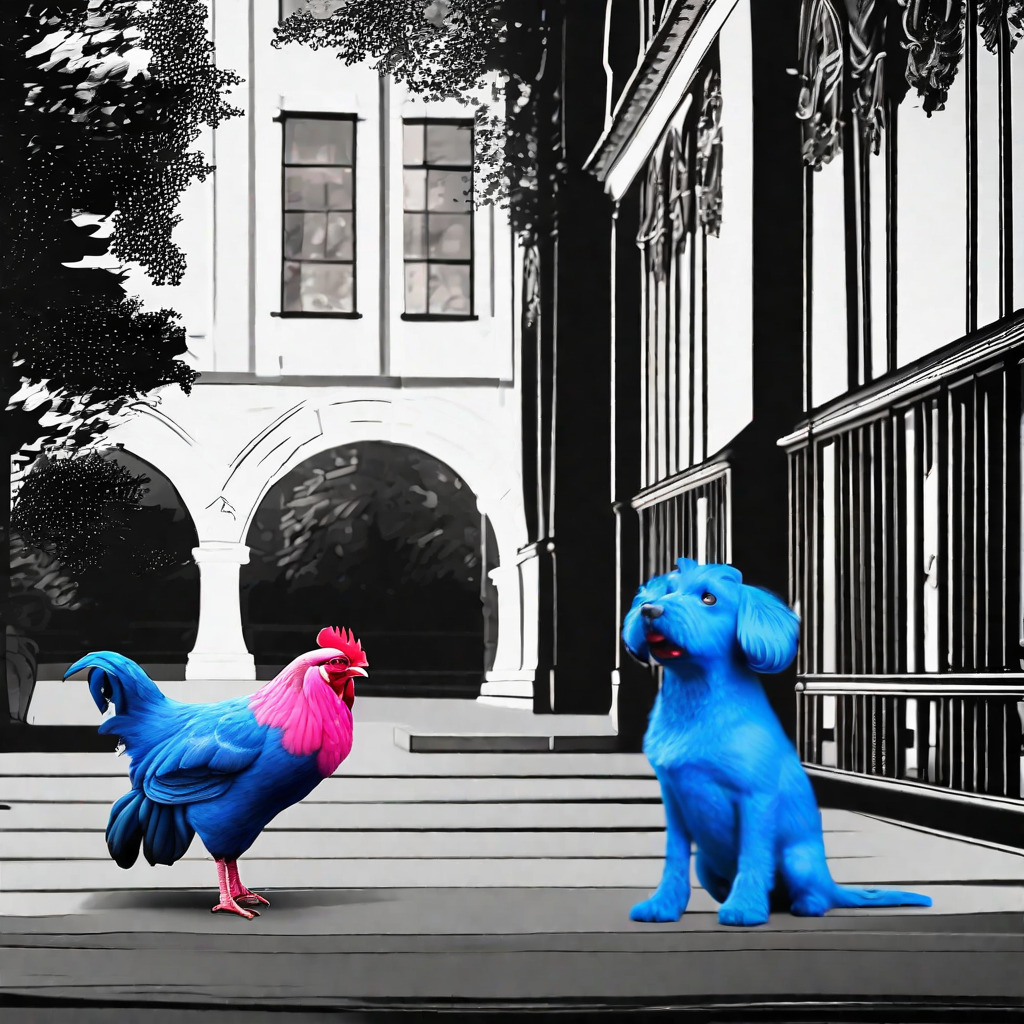} \hfill
\includegraphics[width=0.138\linewidth]{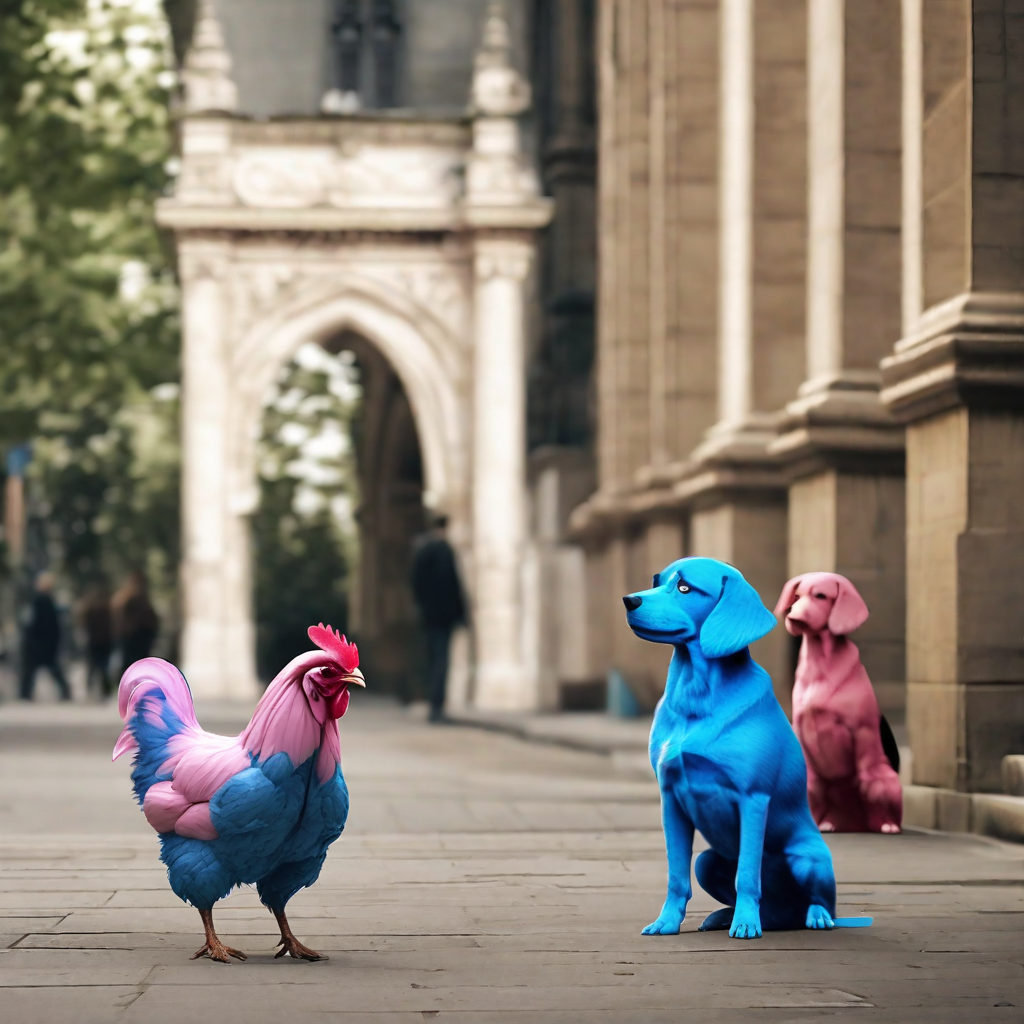} \hfill
\includegraphics[width=0.138\linewidth]{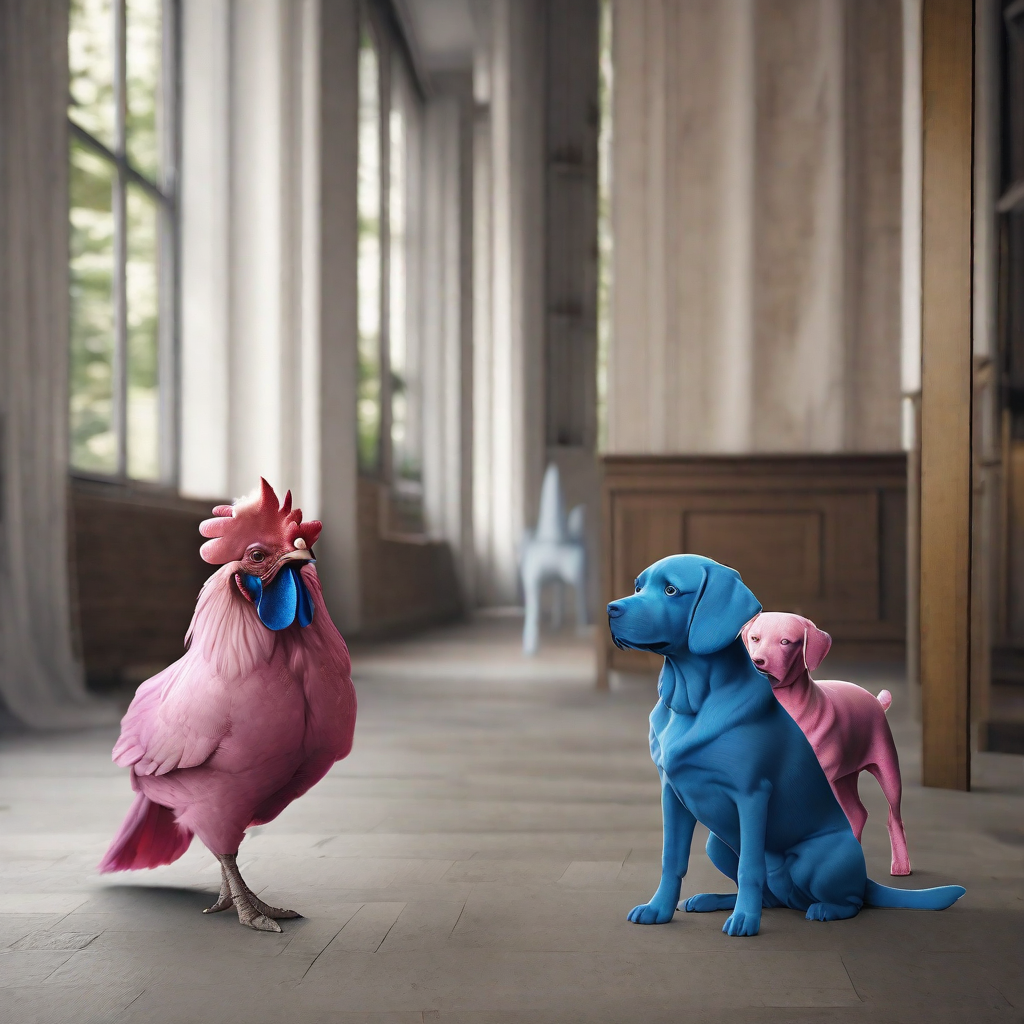} \hfill
\includegraphics[width=0.138\linewidth]{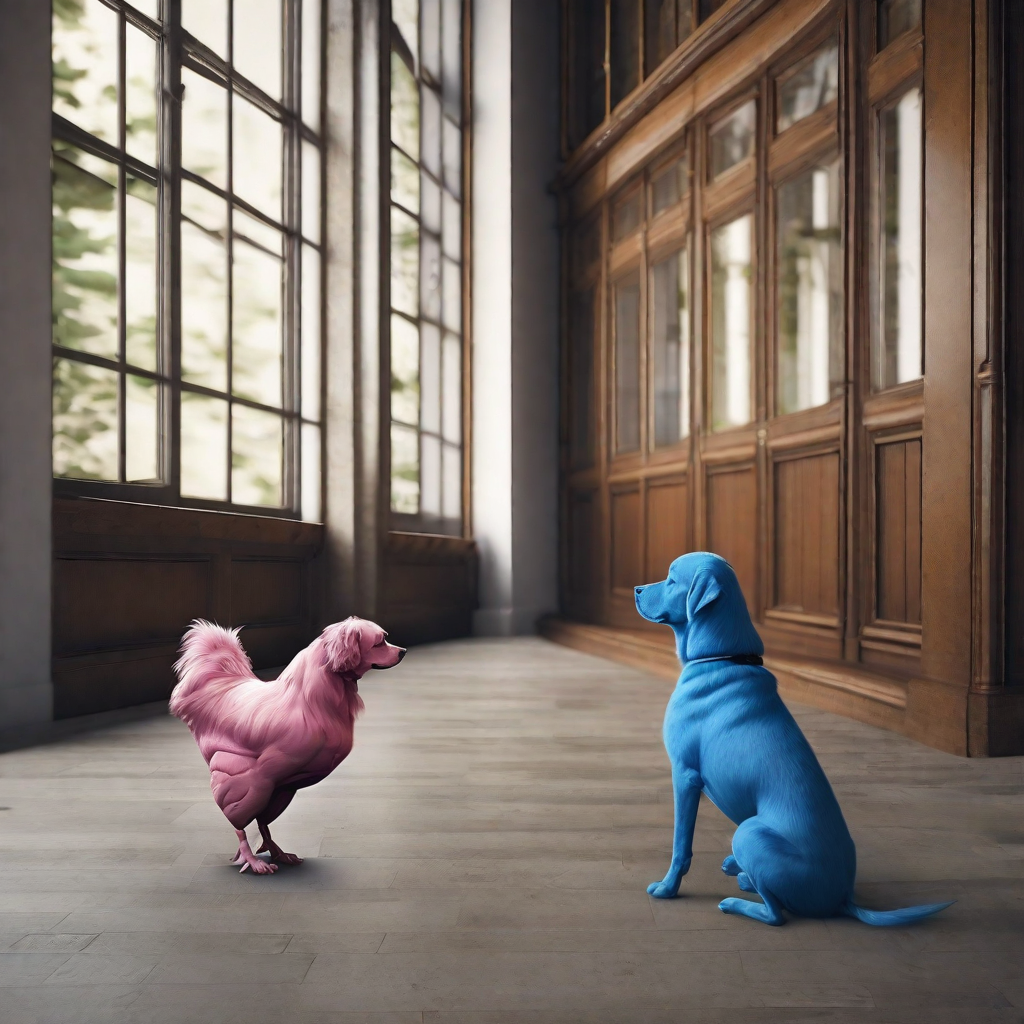} \hfill
\includegraphics[width=0.138\linewidth]{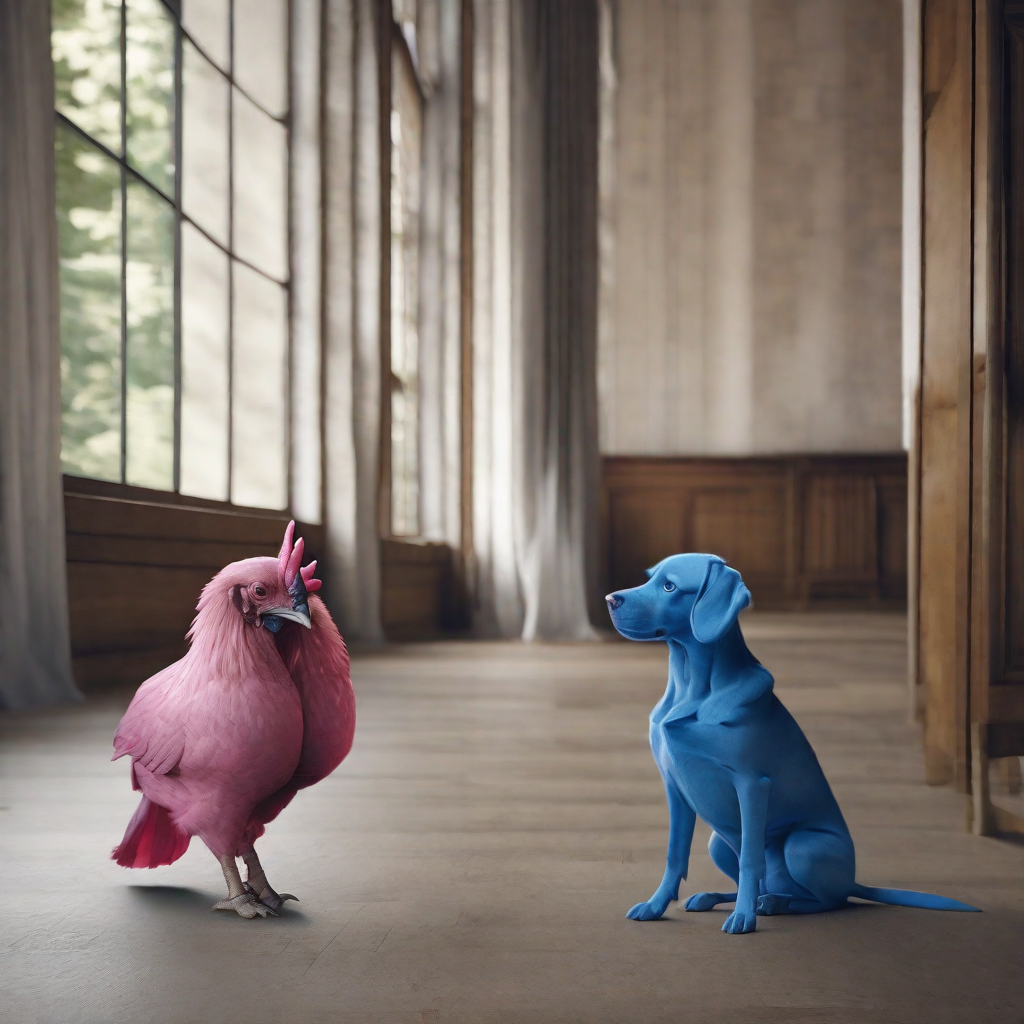} \\

A realistic photo of a \textbf{\color{Goldenrod} yellow flamingo} and a \textbf{\color{pink} pink duck} and a \textbf{\color{purple} purple rabbit} \\
\includegraphics[width=0.138\linewidth]{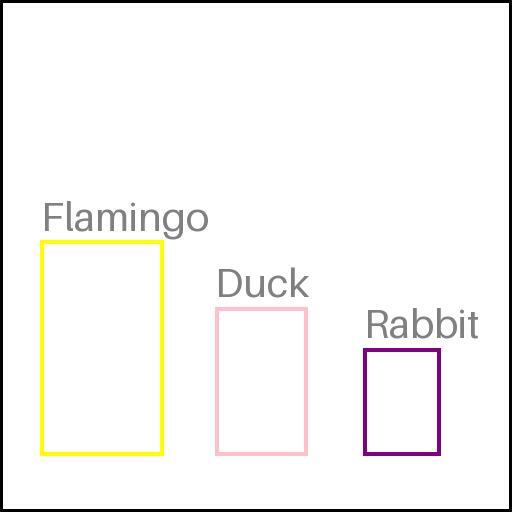} \hfill
\includegraphics[width=0.138\linewidth]{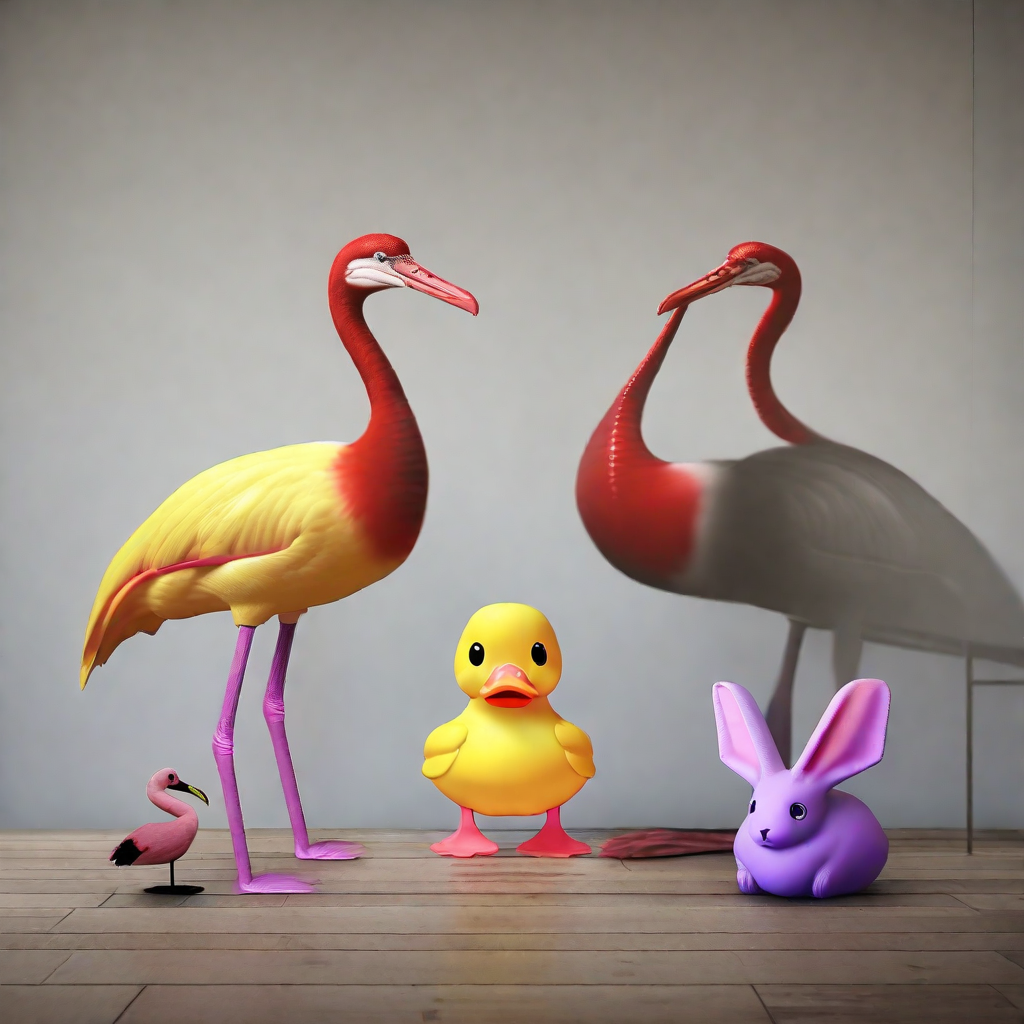} \hfill
\includegraphics[width=0.138\linewidth]{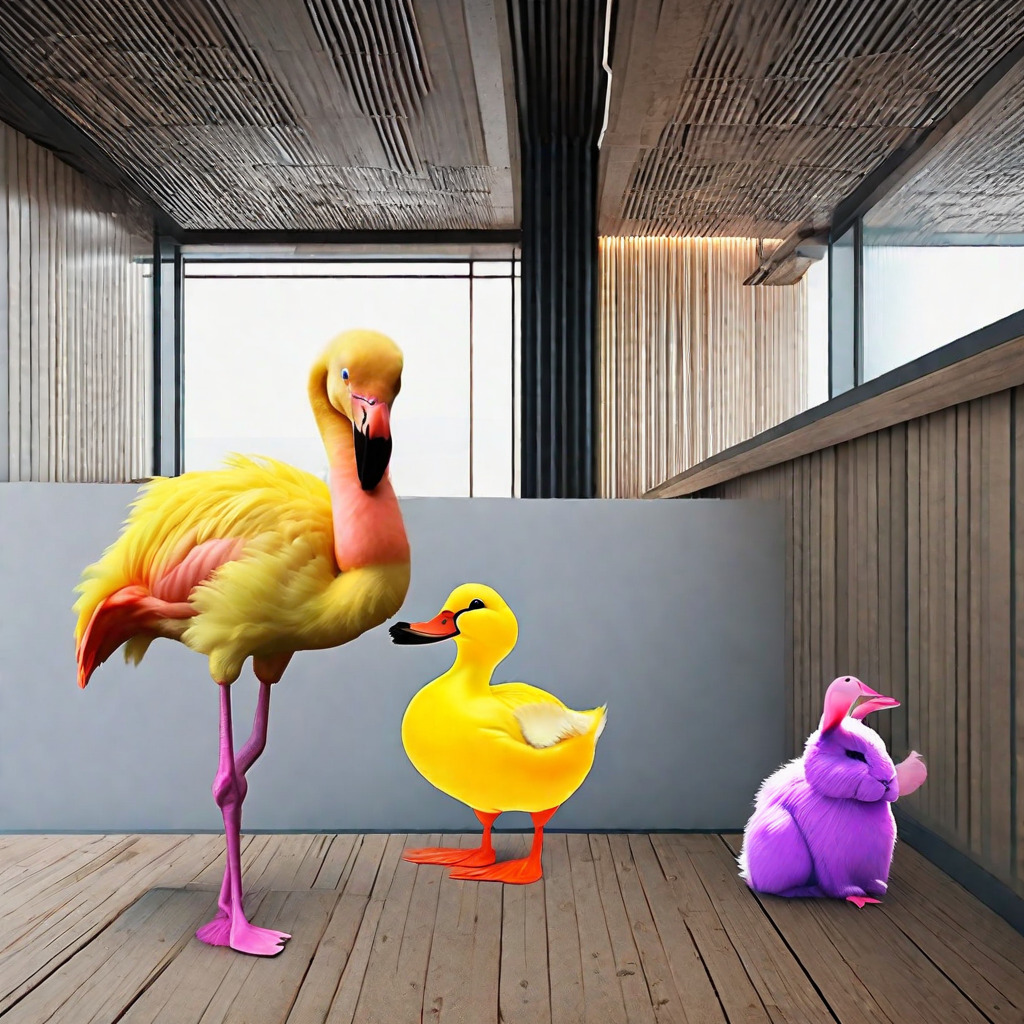} \hfill
\includegraphics[width=0.138\linewidth]{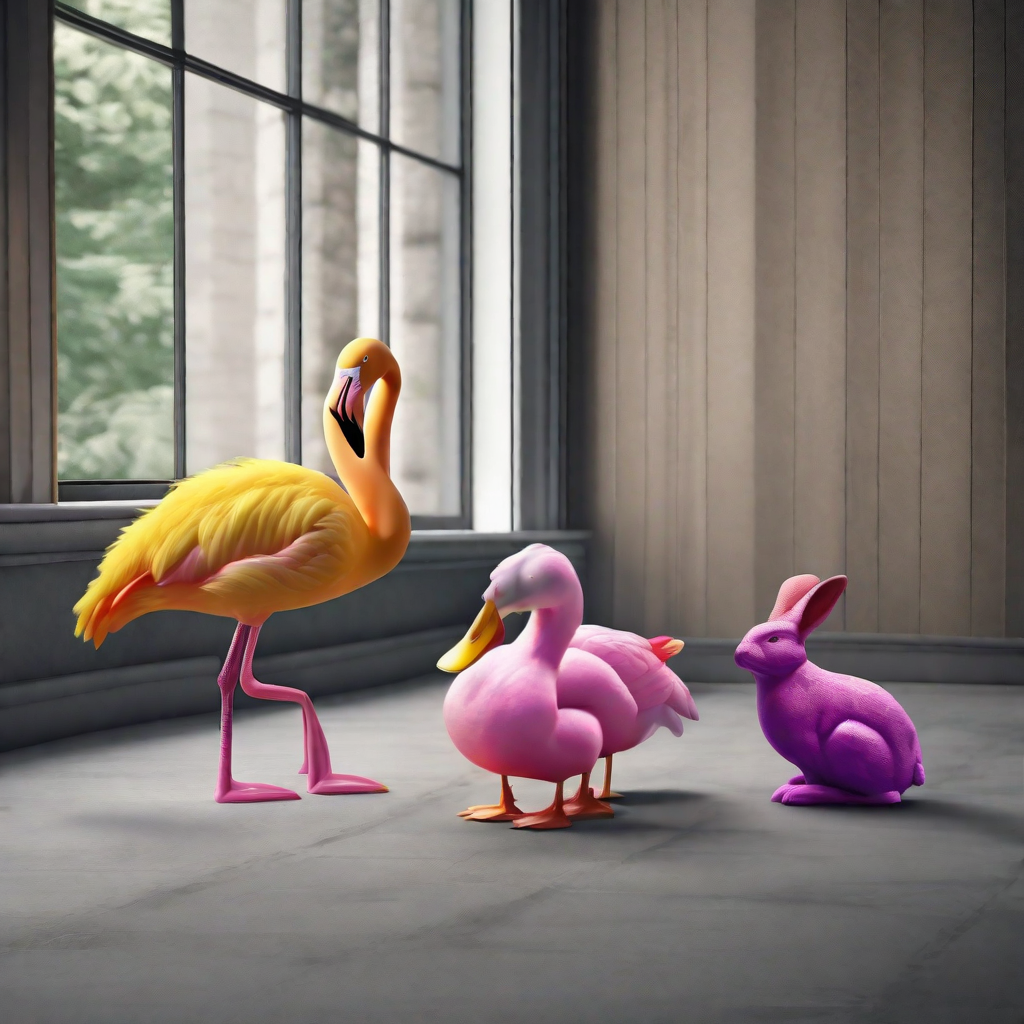} \hfill
\includegraphics[width=0.138\linewidth]{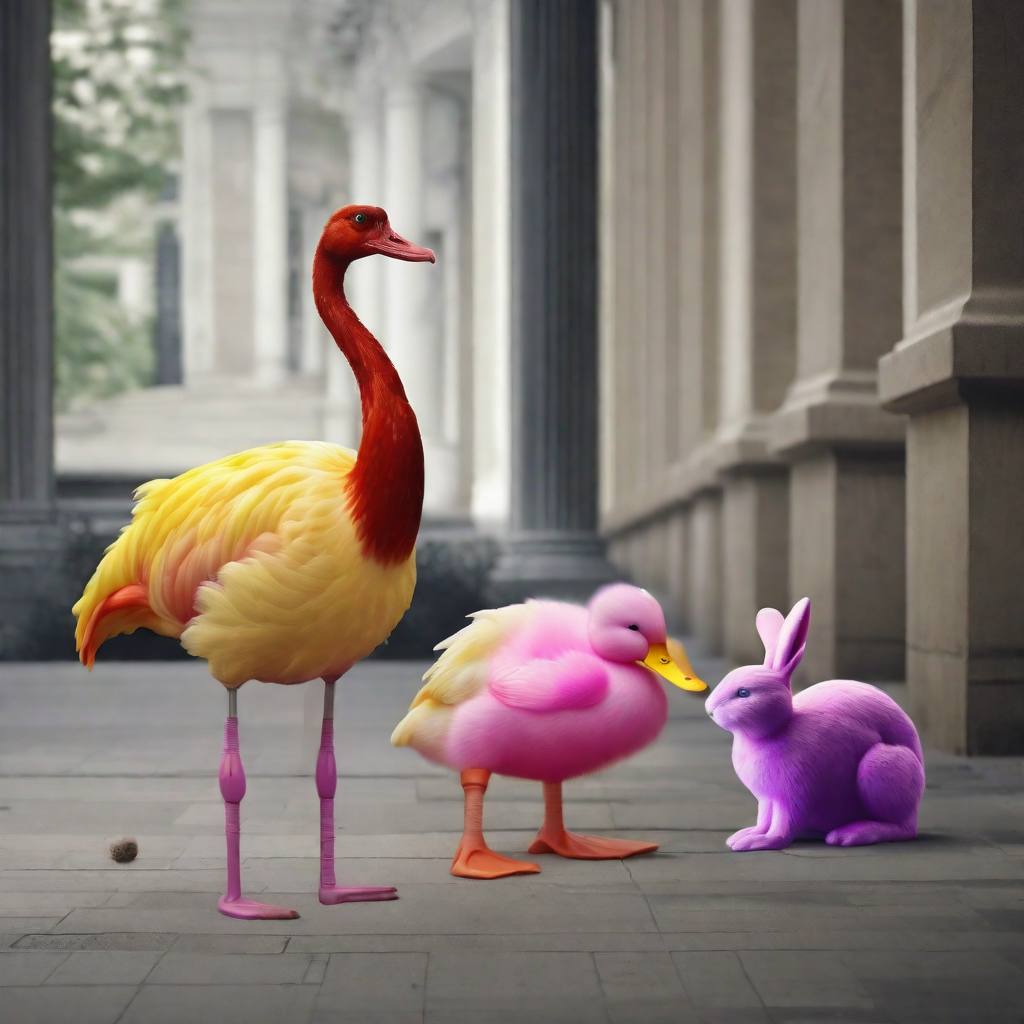} \hfill
\includegraphics[width=0.138\linewidth]{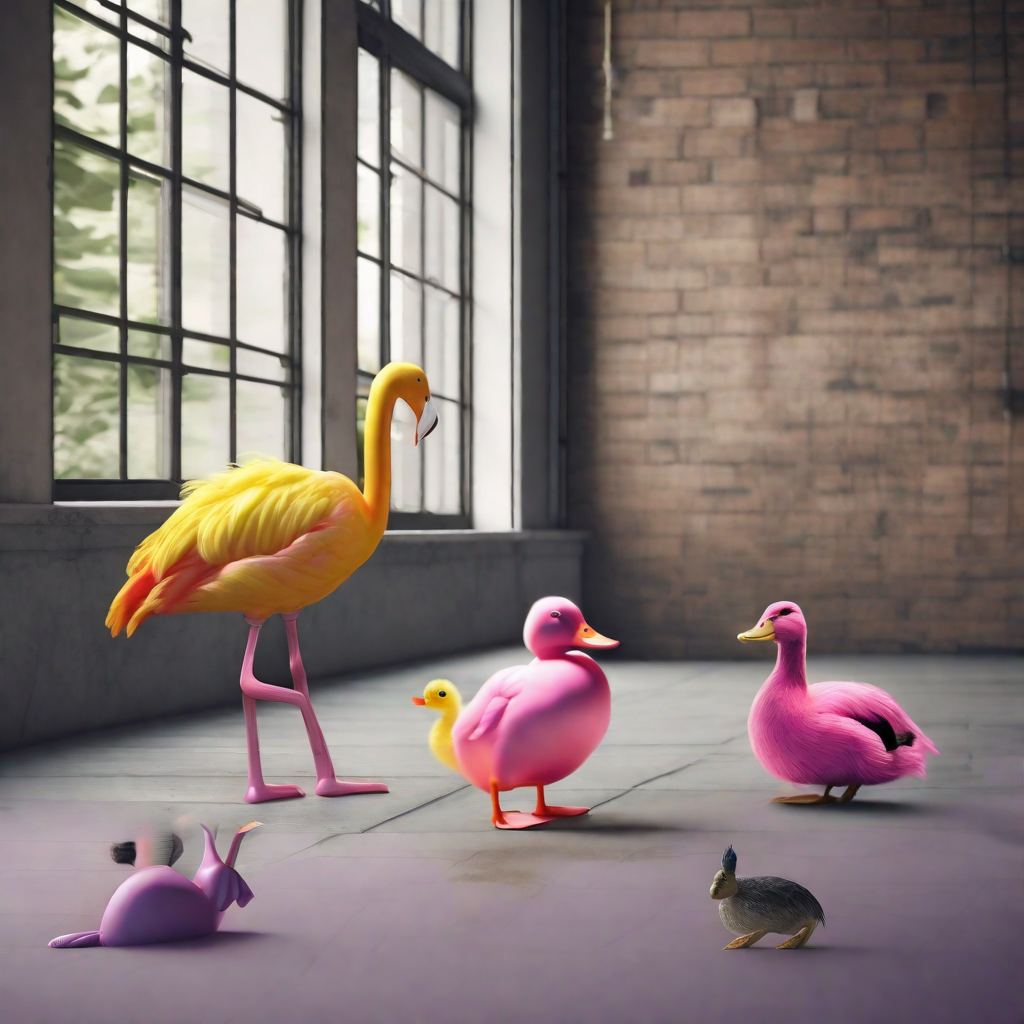} \hfill
\includegraphics[width=0.138\linewidth]{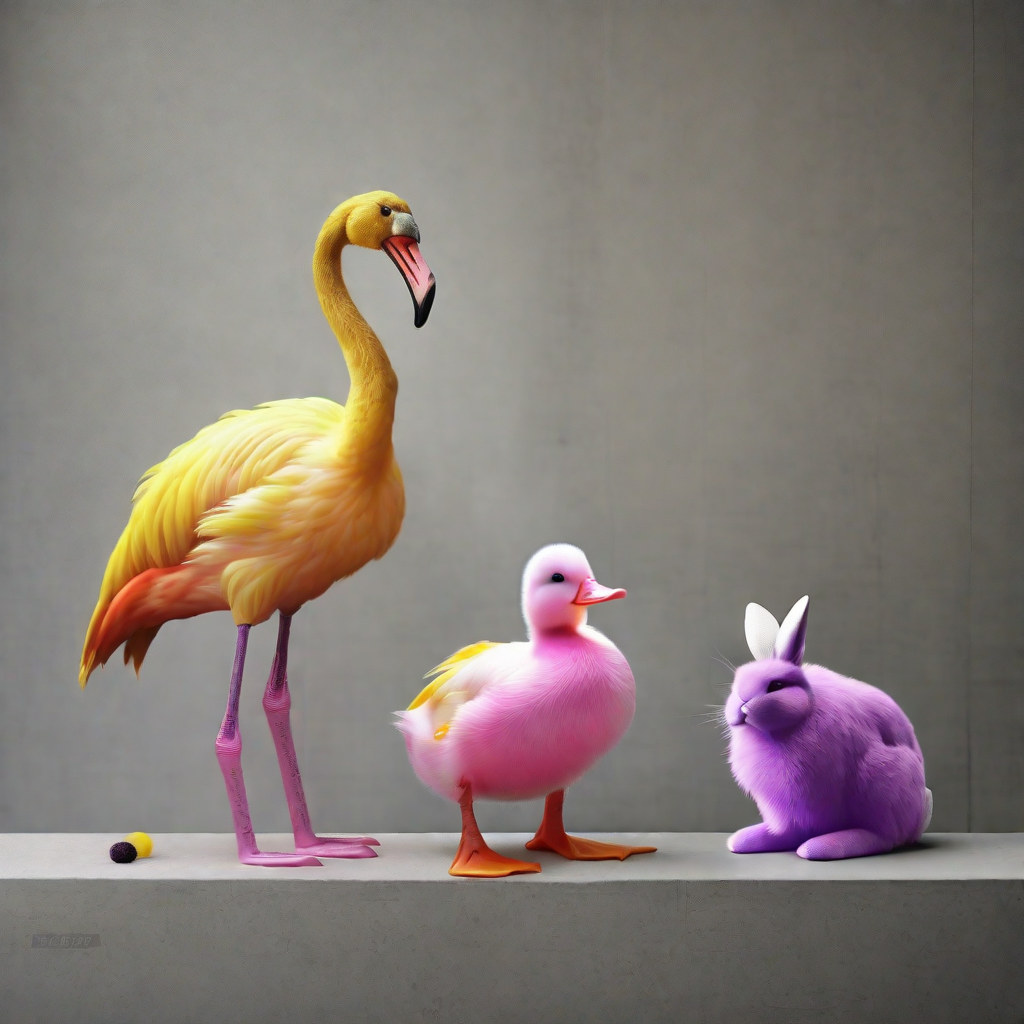} \\

\begin{tabularx}{\linewidth}{*{7}{>{\centering\arraybackslash}X}}
\centering
(a) Layout &
(b) $ \Liou $ &
(c) $+ \Lmask$ &
(d) $+ \Lmask + \Lkl$ &
(e) $+ \text{ only } \Lsim$ &
(f) $+ \text{ only } \Ldis$ &
(g) $+ \Latt$ \\
\end{tabularx}
\vspace{-4mm}
\caption{
Qualitative ablation on the impact of our various loss terms in \cref{eq:final_loss}.
From L2R, columns are:
(a)~layout prompt,
(b)~output of $\Liou$ only,
(c)~effect of masked latent regularization ($\Liou + \Lmask$), and
(d)~together with KL regularization ($\Liou + \Lmask + \Lkl$).
Next, we present the impact of adding attribute losses to (d):
(e)~similarity loss ($+\Lsim$),
(f)~dissimilarity loss ($+\Ldis$), and
(g)~full attribute loss ($+\Latt$) consisting of all loss terms, and corresponding to our final approach, \modelshort{}.
All images are generated using seed 0.}
\label{fig:loss_ablation}
\vspace{-3mm}
\end{figure*}

%% file: tables/loss_ablations.tex
\begin{table}[t]
\centering
\small
\tabcolsep=0.14cm
\caption{
Ablation of individual loss components on DrawBench. 
}
\label{tab:loss_ablation}
\vspace{-3mm}

\begin{tabular}{c cccc c c ccc}
\toprule
& \multicolumn{4}{c}{Losses} & Spatial & Color & \multicolumn{3}{c}{Counting} \\

& $\Liou$ & $\Lmask$ & $\Lkl$ & $\Latt$ & Acc. & Acc. & P & R & F1 \\
\midrule
1 & \cmark & - & - & - &
0.69 & 0.44 & 0.74 & 0.95 & 0.83 \\
2 & \cmark & \cmark & - & - &
0.63 & 0.39 & 0.78 & 0.85 & 0.81 \\
3 & \cmark & - & \cmark & - &
0.74 & 0.44 & 0.75 & 0.96 & 0.84 \\
4 & \cmark & \cmark & \cmark & - &
\textbf{0.81} & 0.47 & 0.81 & 0.96 & \textbf{0.88} \\
\rowcolor{yellow!20}
5 & \cmark & \cmark & \cmark & \cmark &
\textbf{0.81} & \textbf{0.61} & 0.81 & 0.96 & \textbf{0.88} \\
\bottomrule
\end{tabular}
\vspace{-4mm}
\end{table}

%% file: figures/limitations.tex
\begin{figure}[t]
\centering
\tabcolsep=0.05cm
\small
\begin{tabular}{lcccc}
\raisebox{0.37cm}[\height][\depth]{\rotatebox{90}{Layout}} &
\includegraphics[width=0.185\linewidth]{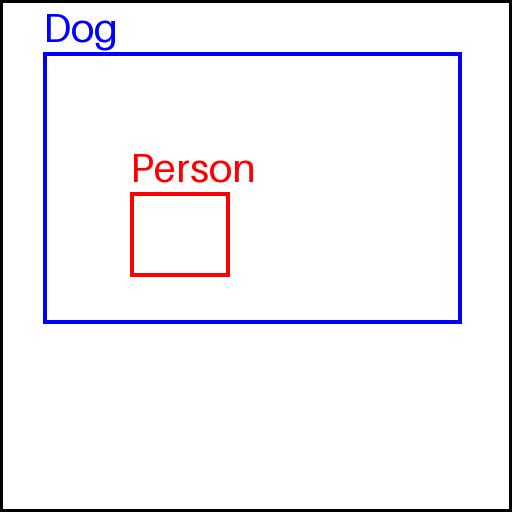} &
\includegraphics[width=0.185\linewidth]{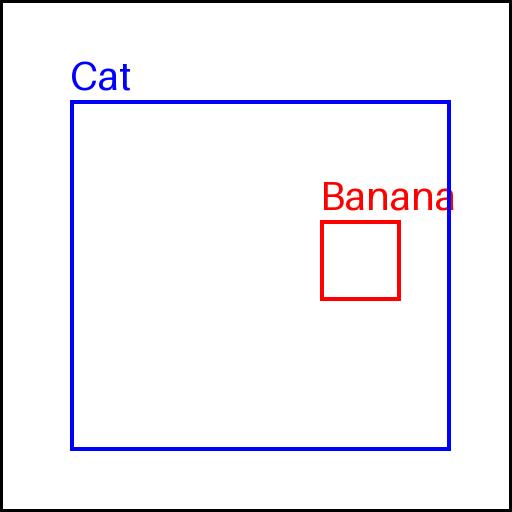} &
\includegraphics[width=0.185\linewidth]{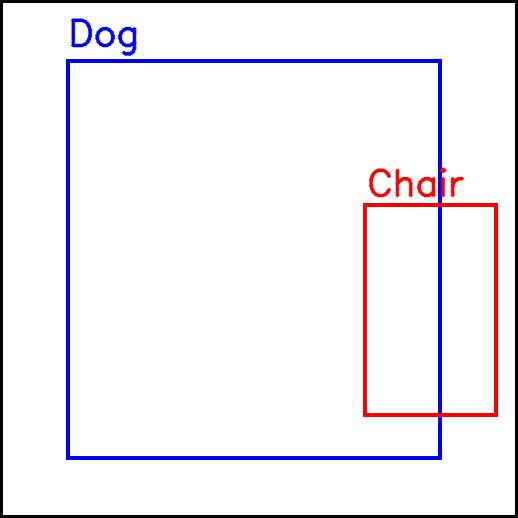} &
\includegraphics[width=0.185\linewidth]{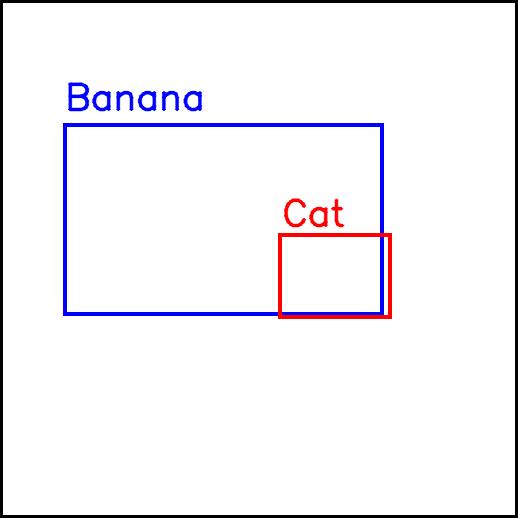} \\
\raisebox{0.6cm}[\height][\depth]{\rotatebox{90}{BA}} &
\includegraphics[width=0.185\linewidth]{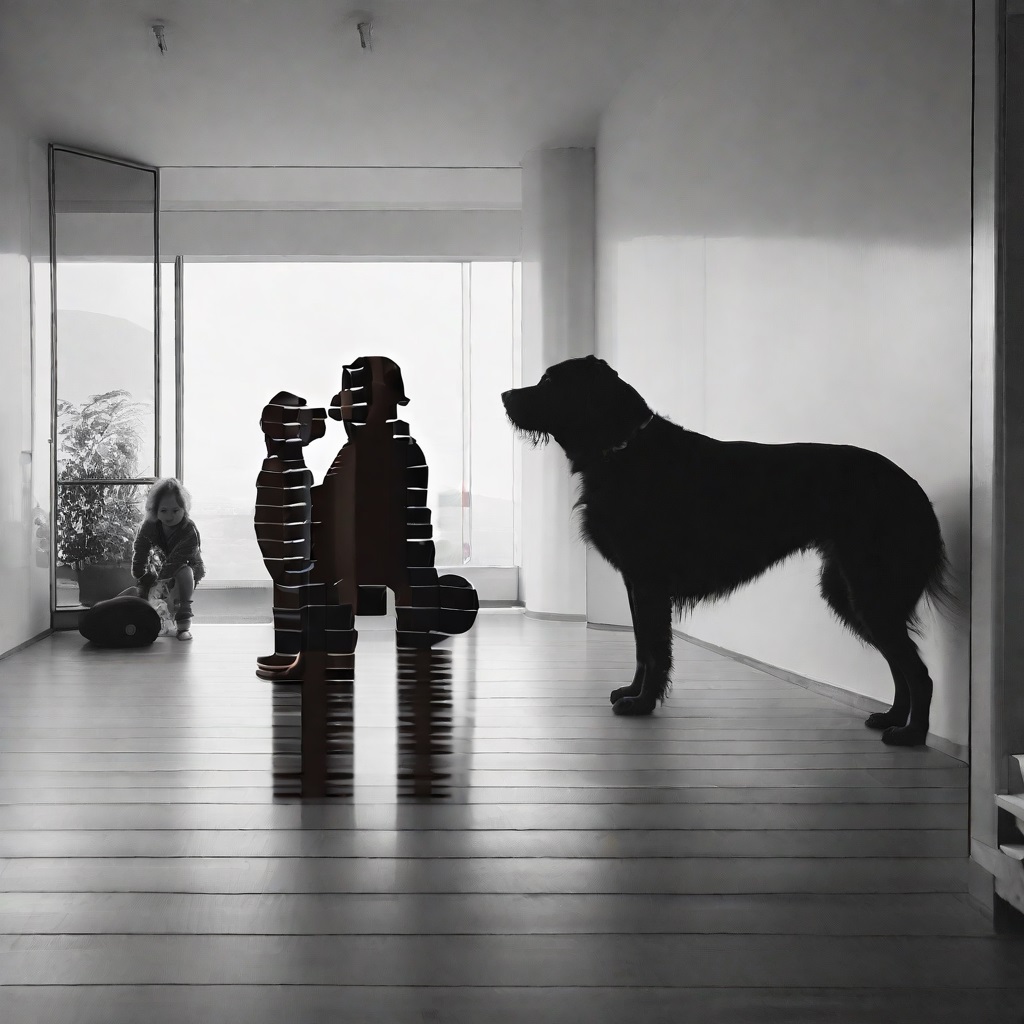} &
\includegraphics[width=0.185\linewidth]{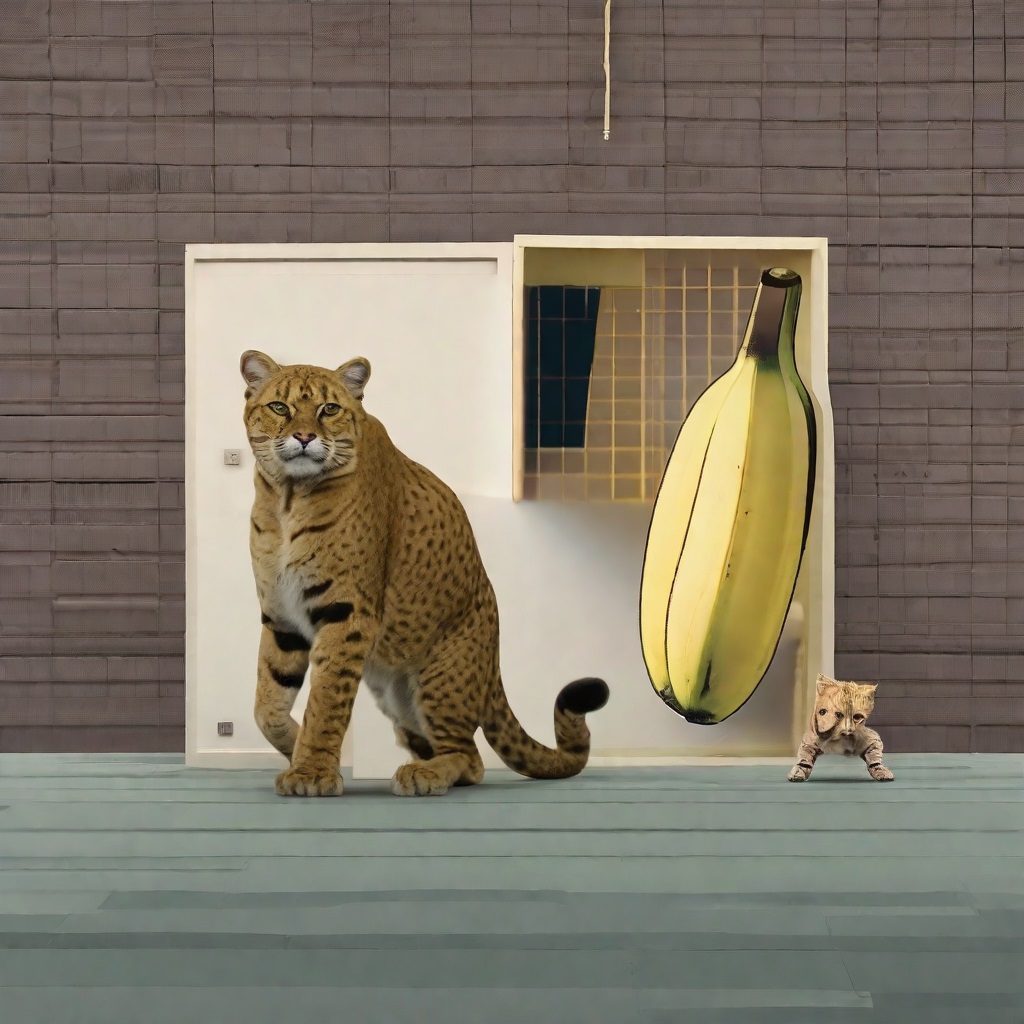} &
\includegraphics[width=0.185\linewidth]{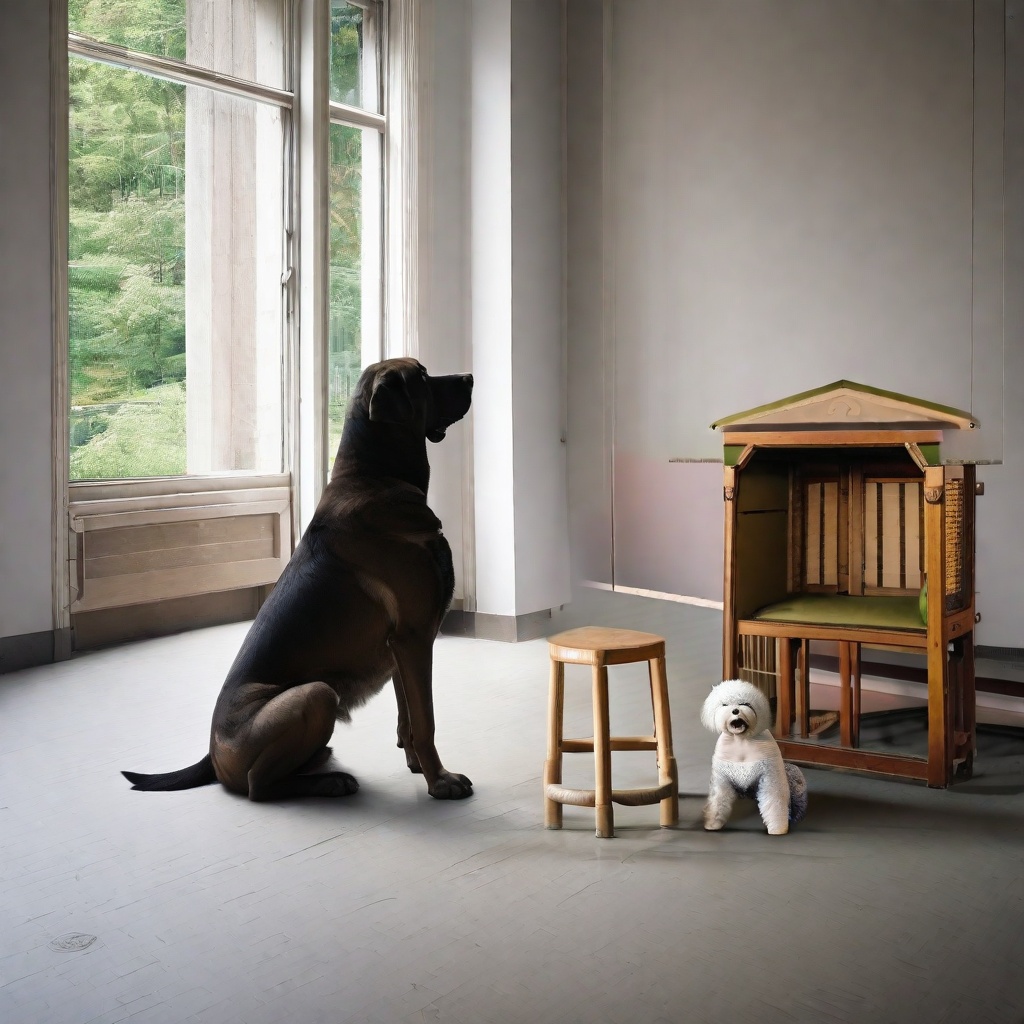} &
\includegraphics[width=0.185\linewidth]{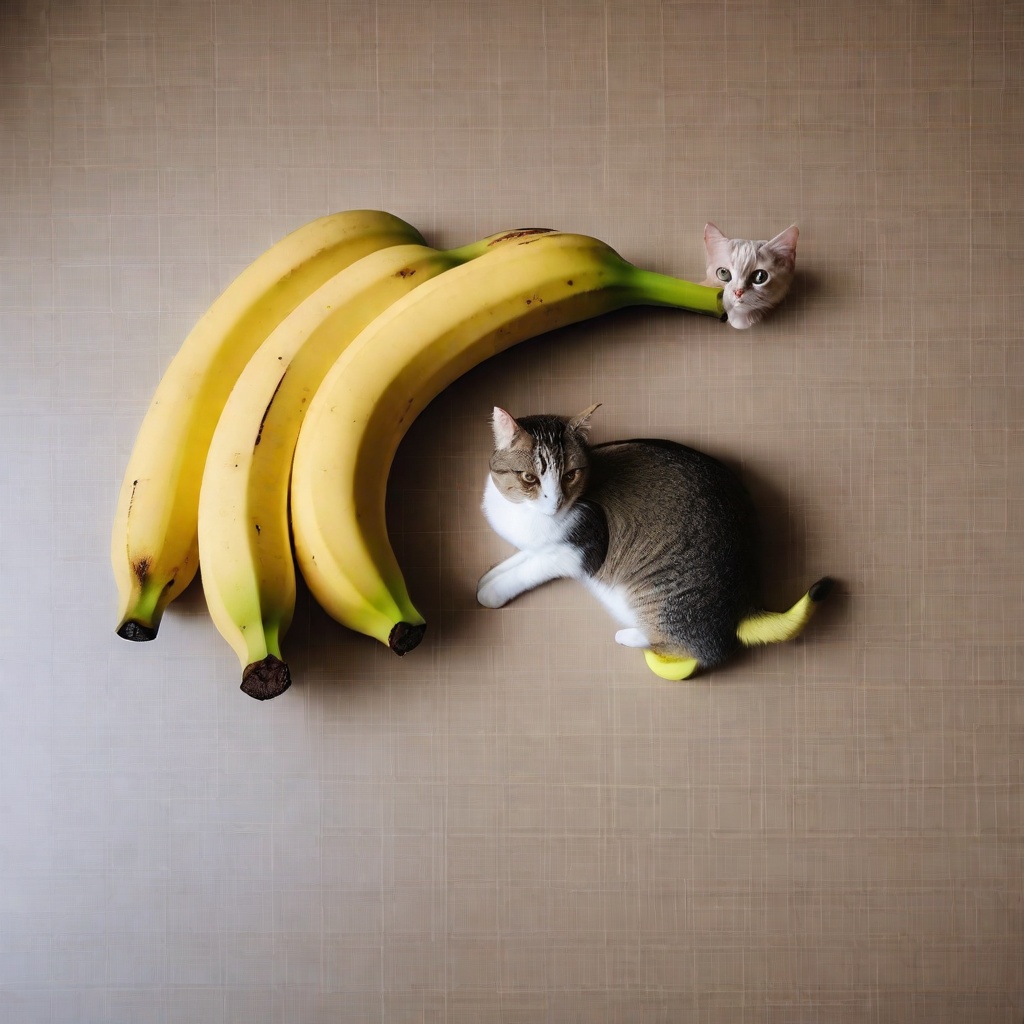} \\
\raisebox{0.37cm}[\height][\depth]{\rotatebox{90}{\modelshort{}}} &
\includegraphics[width=0.185\linewidth]{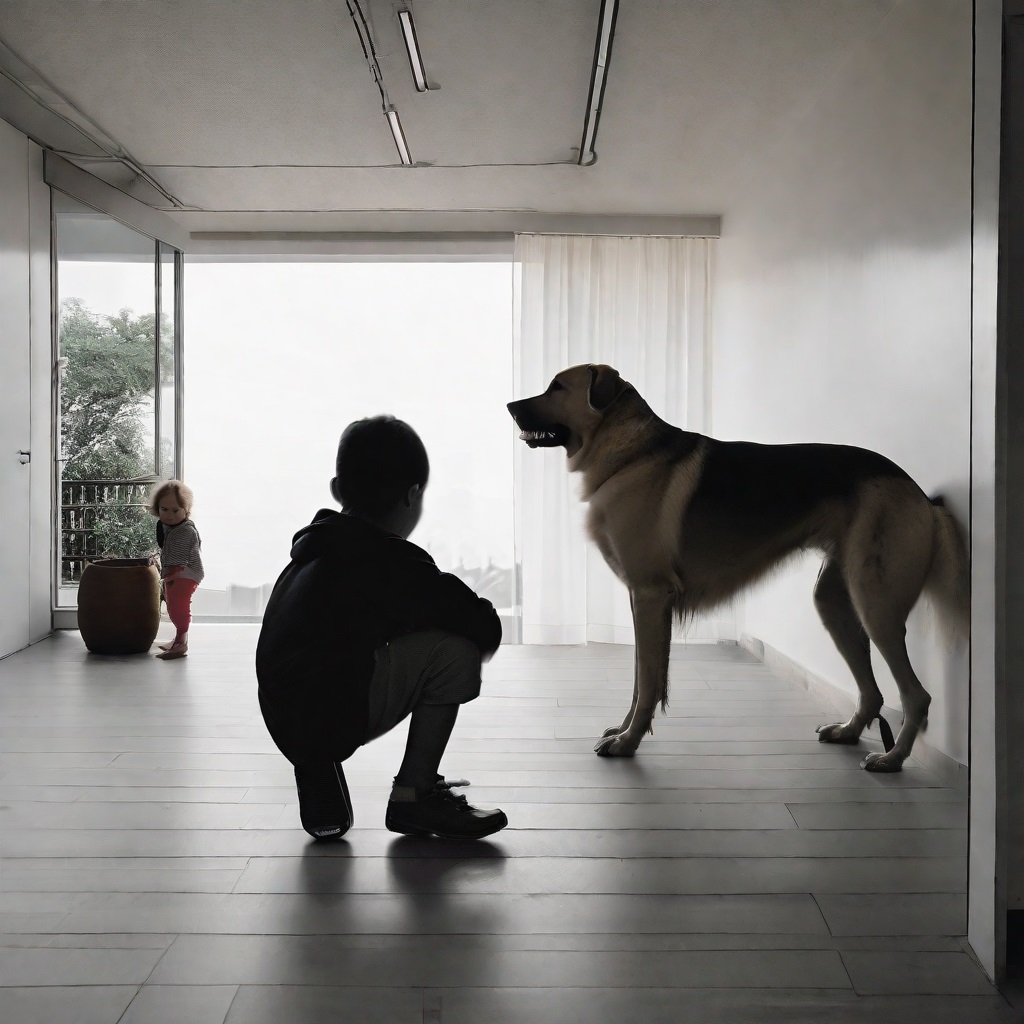} &
\includegraphics[width=0.185\linewidth]{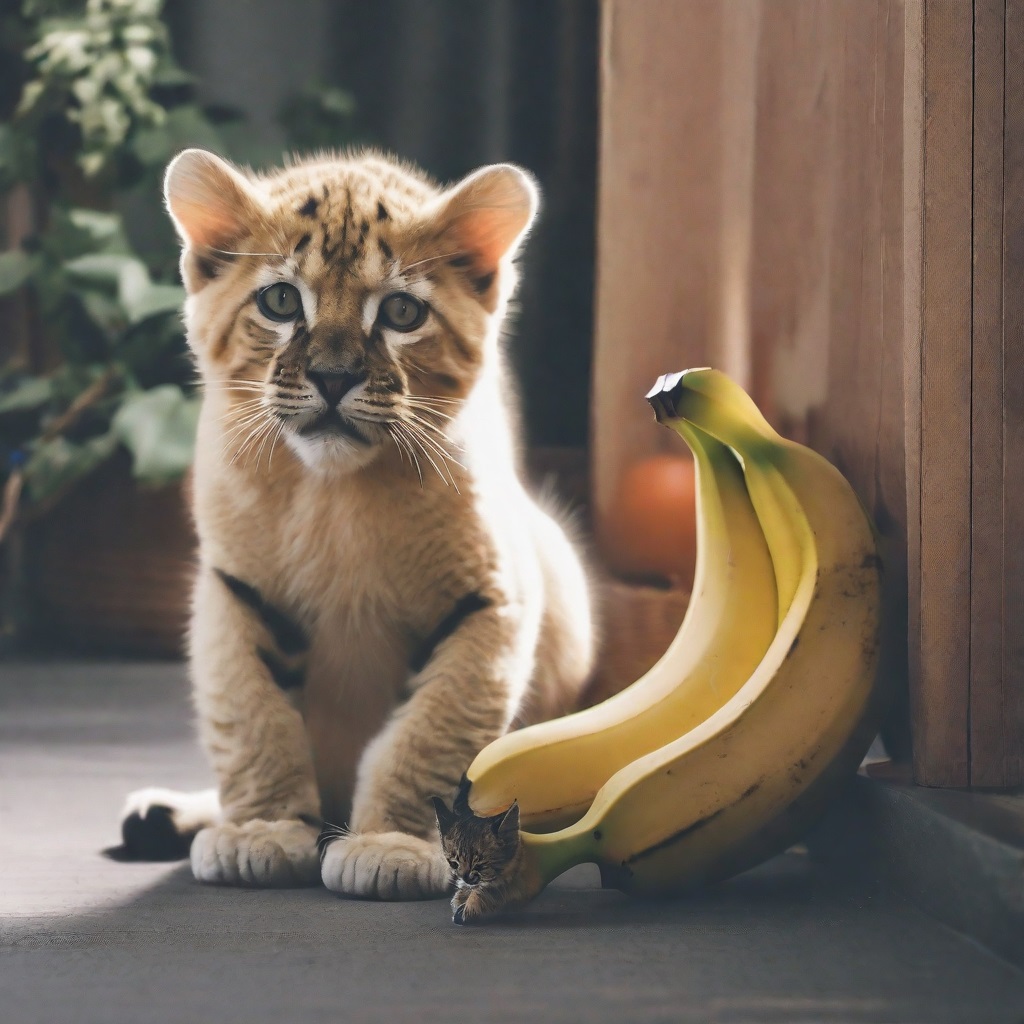} &
\includegraphics[width=0.185\linewidth]{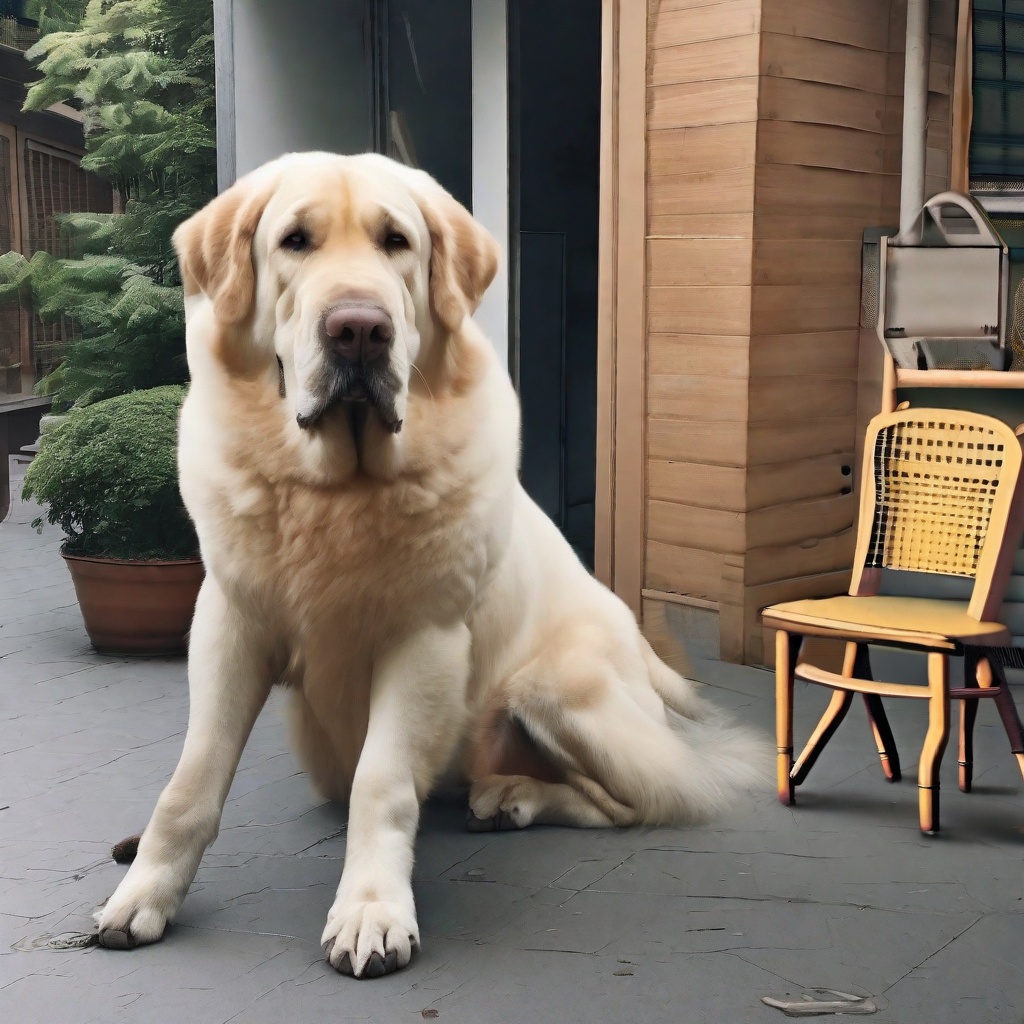} &
\includegraphics[width=0.185\linewidth]{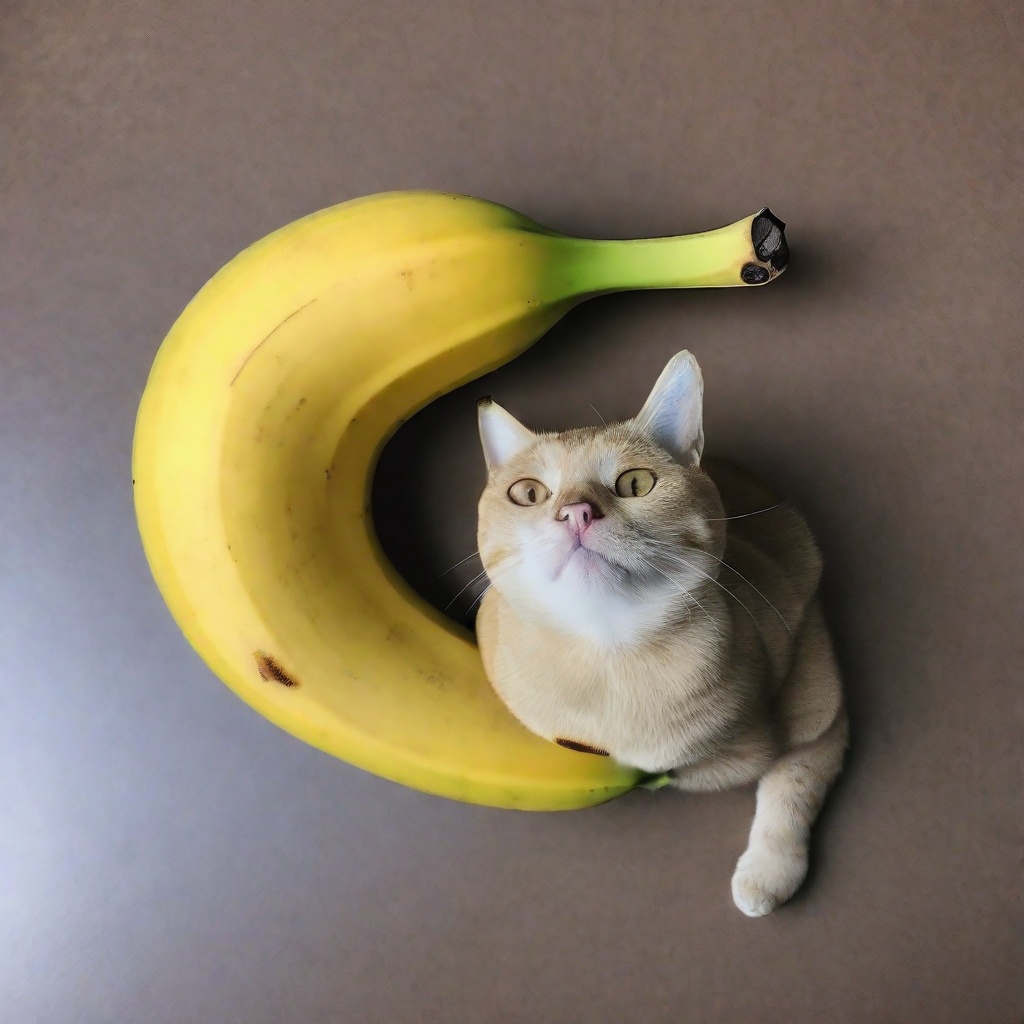} \\
\end{tabular}
\vspace{-3mm}
\caption{
\textbf{Limitations.}
\modelshort{} struggles to generate images when subjects are completely overlapping or have aytpical aspect ratios.
However, generations for partial overlaps are acceptable.
L2R prompts from HRS:
(a) a small person and a big dog;
(b) a small banana and a big cat;
(c) a big dog and a small chair; and
(d) a large banana and a small cat.}
\label{fig:limitations_comparison}
\vspace{-3mm}
\end{figure}

%% file: sec/5_conclusion.tex
\section{Conclusion}
\label{sec:conclusion}

We presented \modelshort{}, a training-free, layout-guided text-to-image approach that enables users to generate compositional scenes involving multiple subjects and multiple attributes.
We pointed out three primary challenges in existing approaches:
(i)~background semantic leakage,
(ii)~out-of-distribution generations, and
(iii)~inaccurate subject-attribute binding in compositional scenes.
\modelshort{} addressed these challenges through
(i)~masked latent regularization,
(ii)~in-distribution latent alignment, and
(iii)~a subject-attribute association loss.
Quantitative comparison showed improved performance over existing approaches while image visualizations showed the ability of \modelshort{} to generate accurate, controllable, and compositional images, with enhanced stability and consistent attribute binding.
We confirmed that \modelshort{} is perceived to generate better images that are more robust to random seeds through user studies.

%% file: sec/6_ack.tex
\begin{acks}

This project was supported by funding from an Adobe Research gift.
MT also thanks SERB SRG/2023/002544 grant for compute support.
We thank all volunteers who participated in our user study.

\end{acks}

%% file: sec/7_figurepages.tex
\input{figures/supp_visuals_page1.tex}
\input{figures/supp_visuals_page2.tex}

%% file: figures/supp_visuals_page1.tex
\begin{figure*}[t]
\tabcolsep=0.05cm
\small
\begin{tabular}{l c cccc}

\begin{tabular}{p{0.2cm}}
\raisebox{0.15cm}[\height][\depth]{\rotatebox{90}{\modelshort{}}} \\
\raisebox{-0.85cm}[\height][\depth]{\rotatebox{90}{Bounded Attention}}
\end{tabular} &

\begin{tabular}{p{5cm}}
\multicolumn{1}{c}{\includegraphics[width=0.133\linewidth]{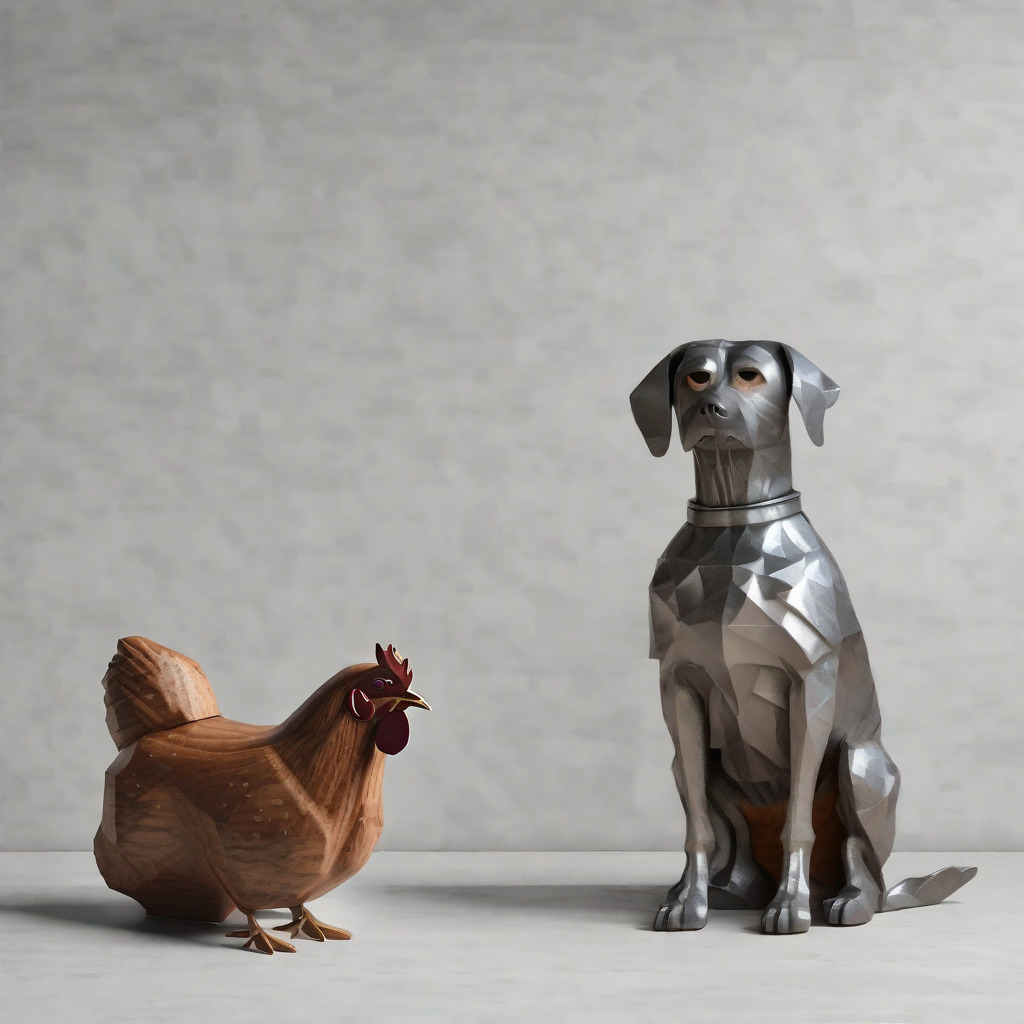}} \\
\multicolumn{1}{c}{\includegraphics[width=0.133\linewidth]{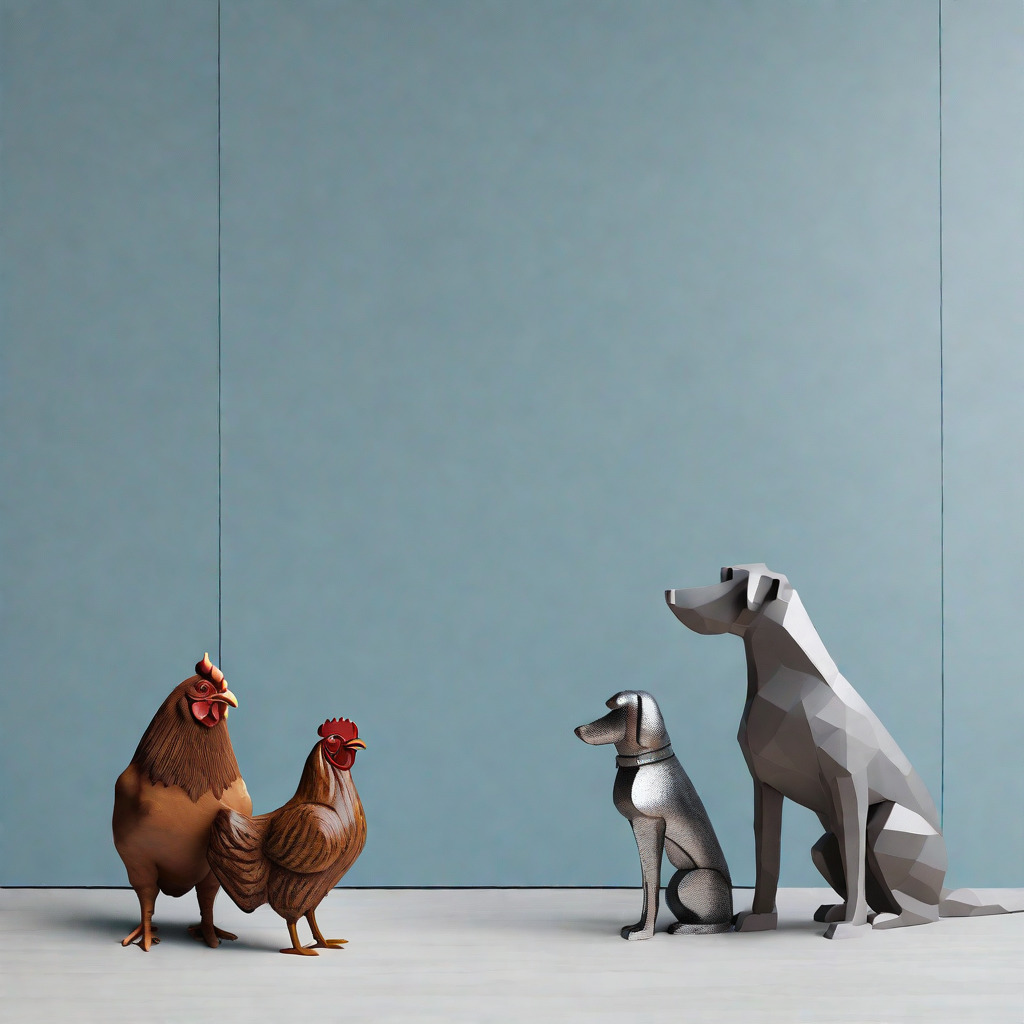}}
\end{tabular} &

\begin{tabular}{p{5cm}}
\multicolumn{1}{c}{\includegraphics[width=0.133\linewidth]{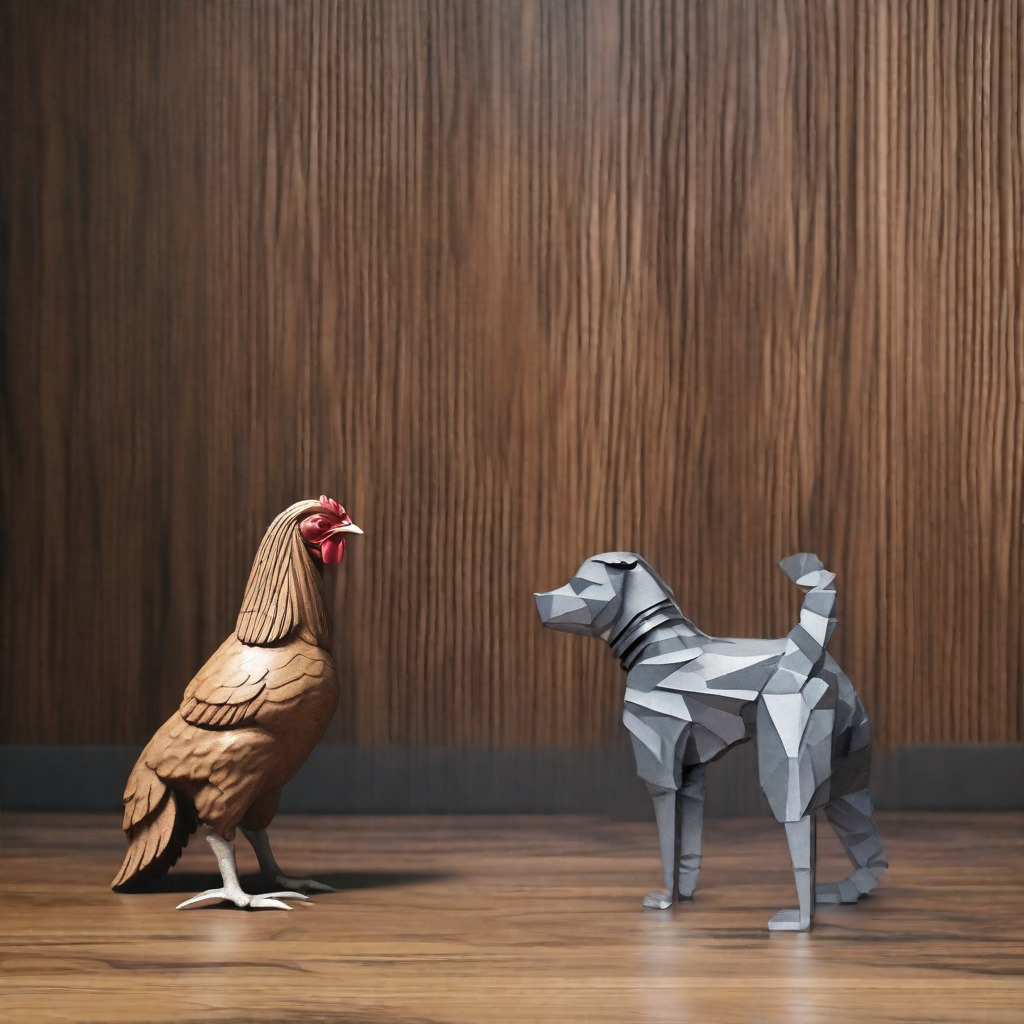}} \\
\multicolumn{1}{c}{\includegraphics[width=0.133\linewidth]{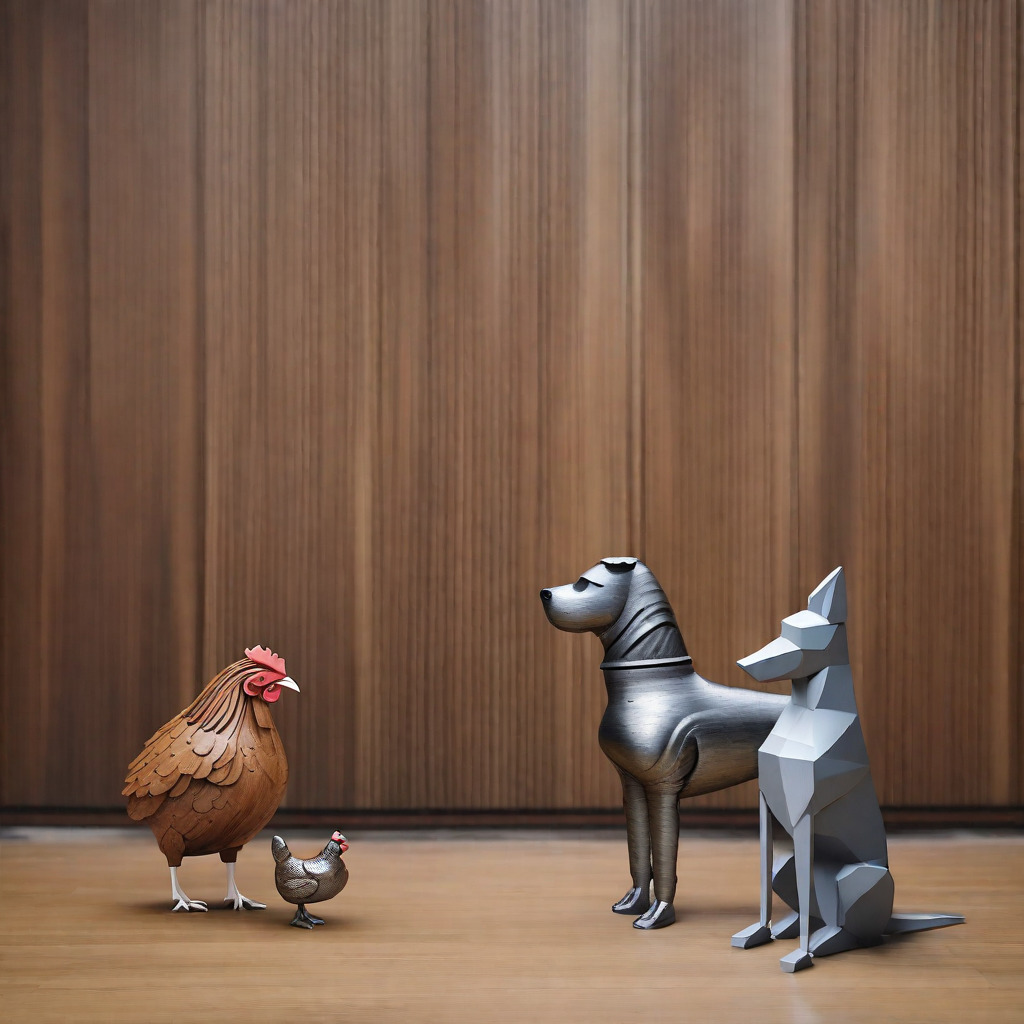}}
\end{tabular} &

\begin{tabular}{p{5cm}}
\multicolumn{1}{c}{\includegraphics[width=0.133\linewidth]{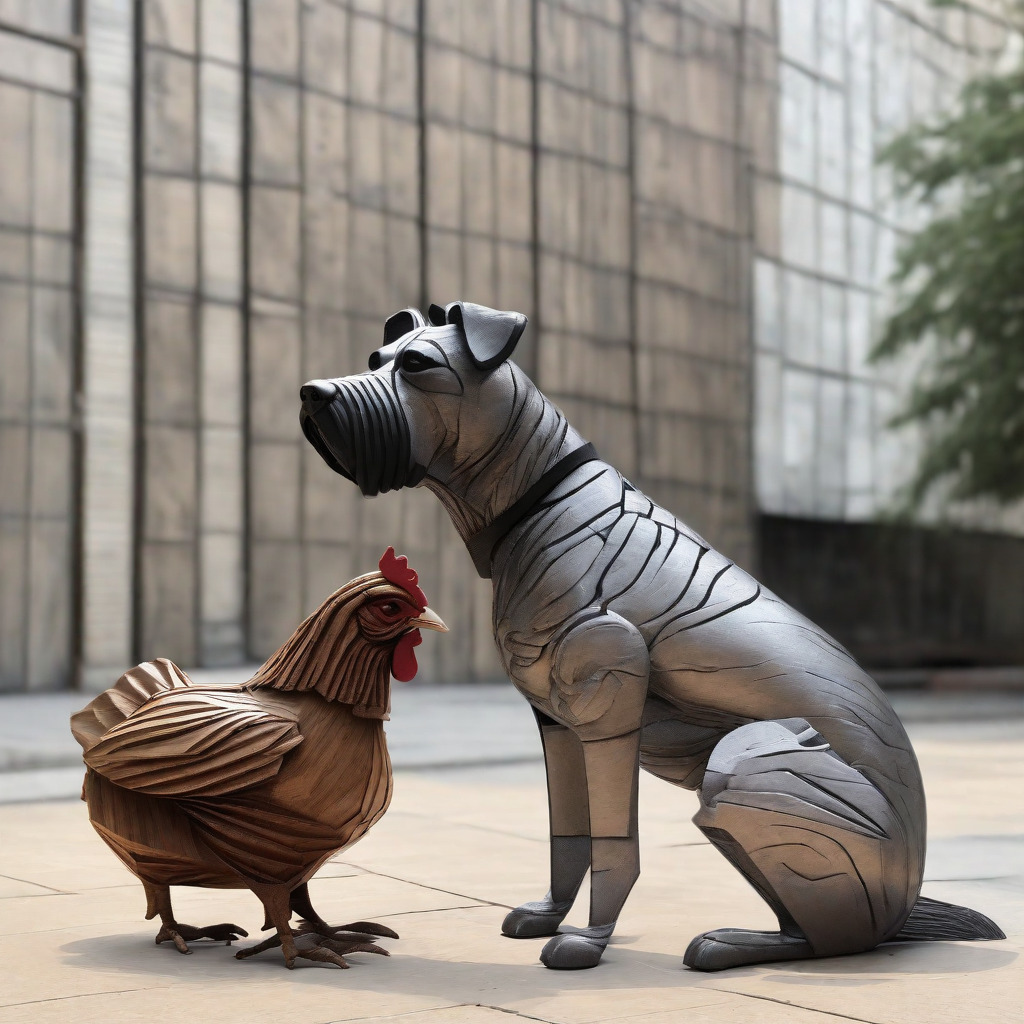}} \\
\multicolumn{1}{c}{\includegraphics[width=0.133\linewidth]{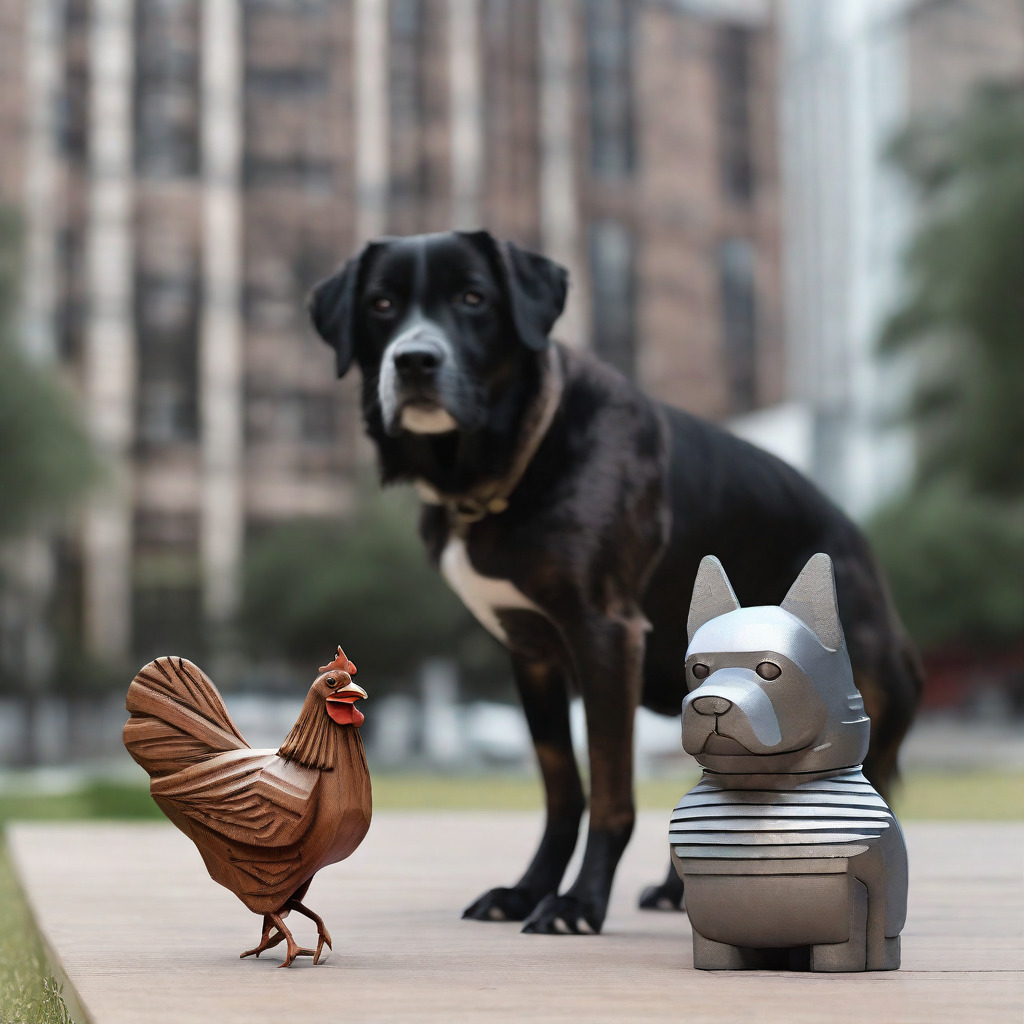}}
\end{tabular} &

\begin{tabular}{p{5cm}}
\multicolumn{1}{c}{\includegraphics[width=0.133\linewidth]{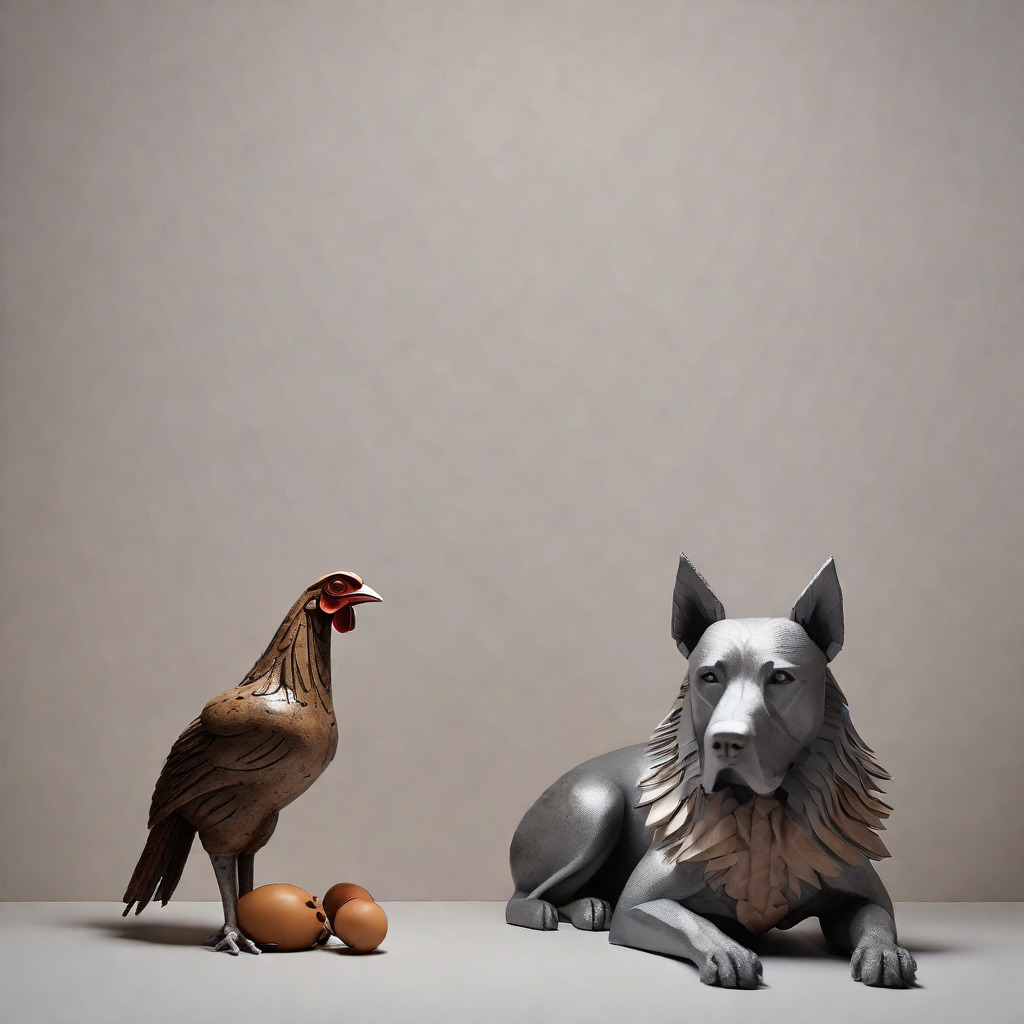}} \\
\multicolumn{1}{c}{\includegraphics[width=0.133\linewidth]{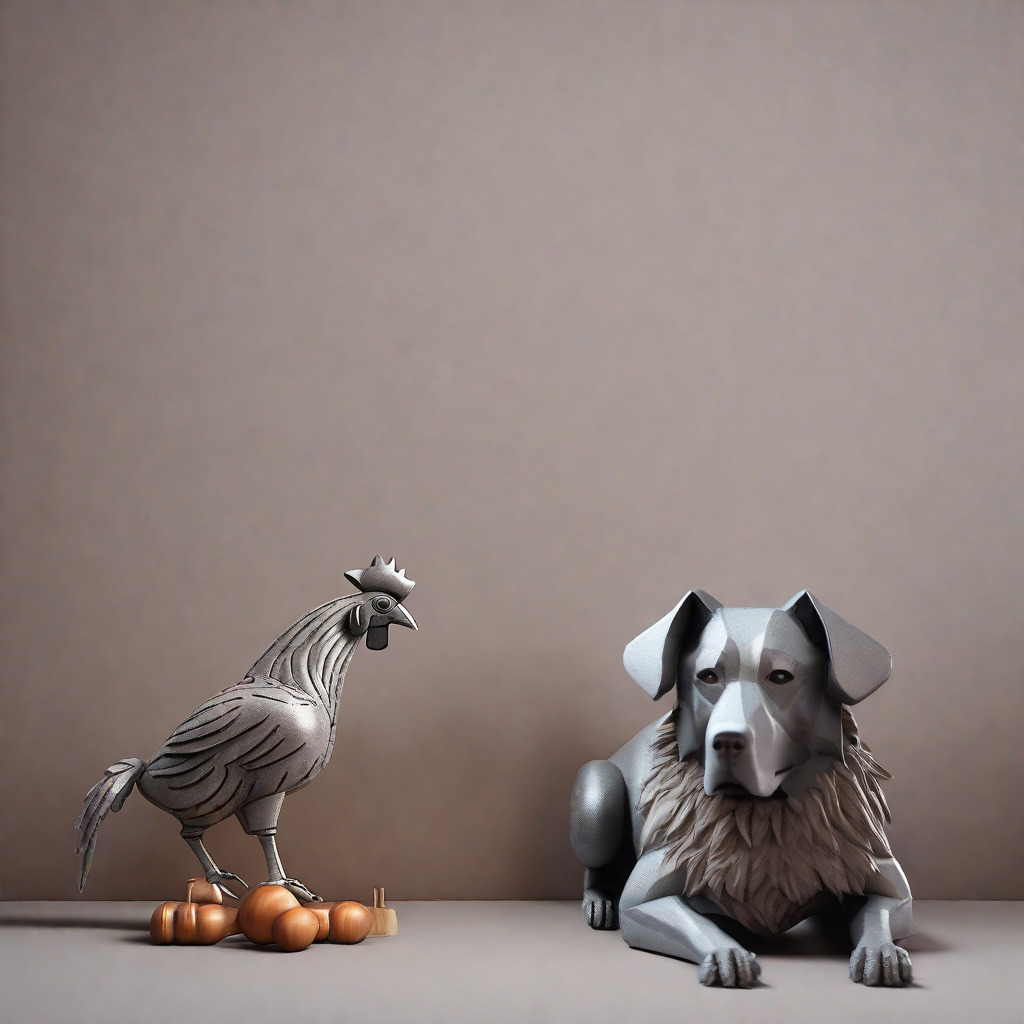}}
\end{tabular} &

\begin{tabular}{p{5cm}}
\multicolumn{1}{p{3.5cm}}{\centering 1. A realistic photo of a \textbf{\color{brown} brown wooden chicken} and a \textbf{\color{gray} gray metallic dog}}. \\
\multicolumn{1}{c}{\includegraphics[width=0.133\linewidth]{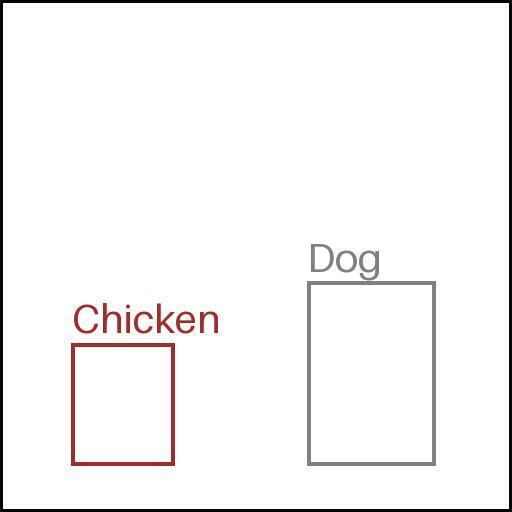}}
\end{tabular}

\\
\midrule

\begin{tabular}{p{0.2cm}}
\raisebox{0.15cm}[\height][\depth]{\rotatebox{90}{\modelshort{}}} \\
\raisebox{-0.85cm}[\height][\depth]{\rotatebox{90}{Bounded Attention}}
\end{tabular} &

\begin{tabular}{p{5cm}}
\multicolumn{1}{c}{\includegraphics[width=0.133\linewidth]{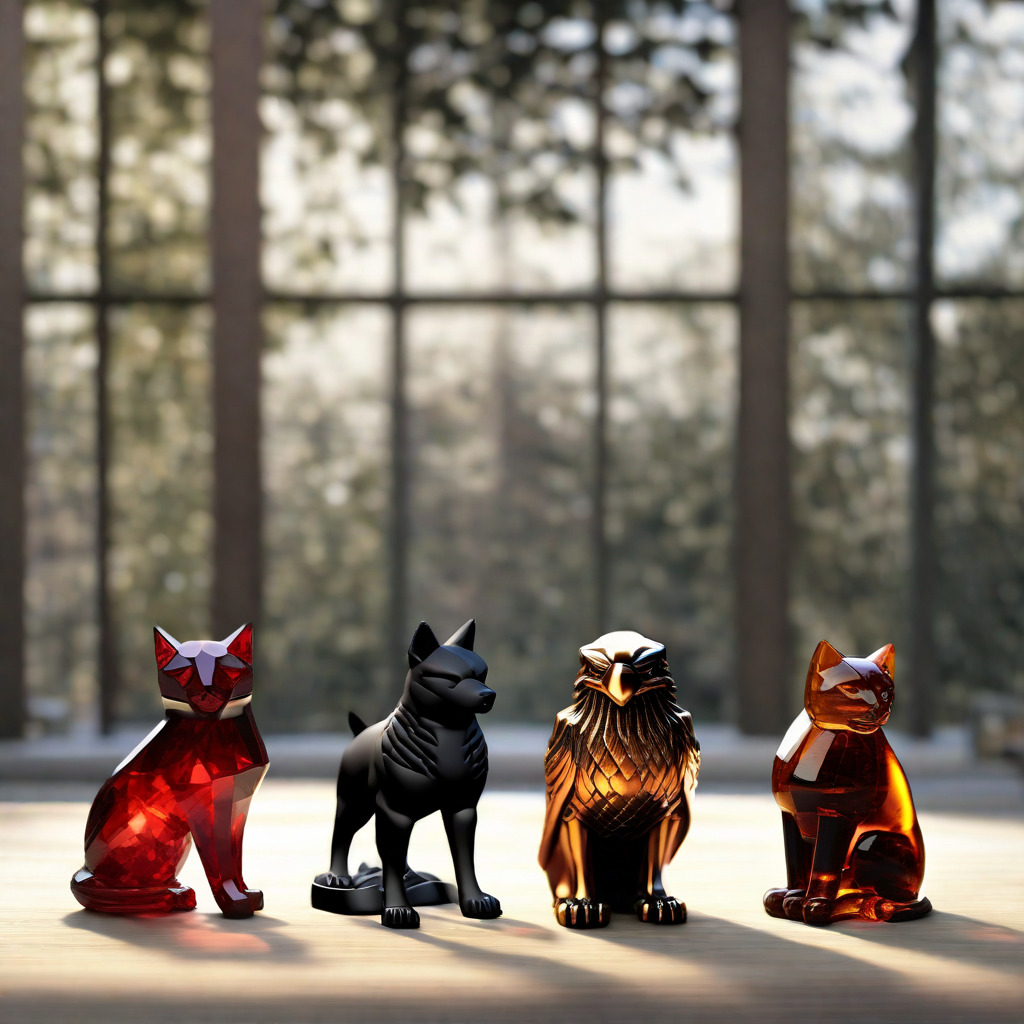}} \\
\multicolumn{1}{c}{\includegraphics[width=0.133\linewidth]{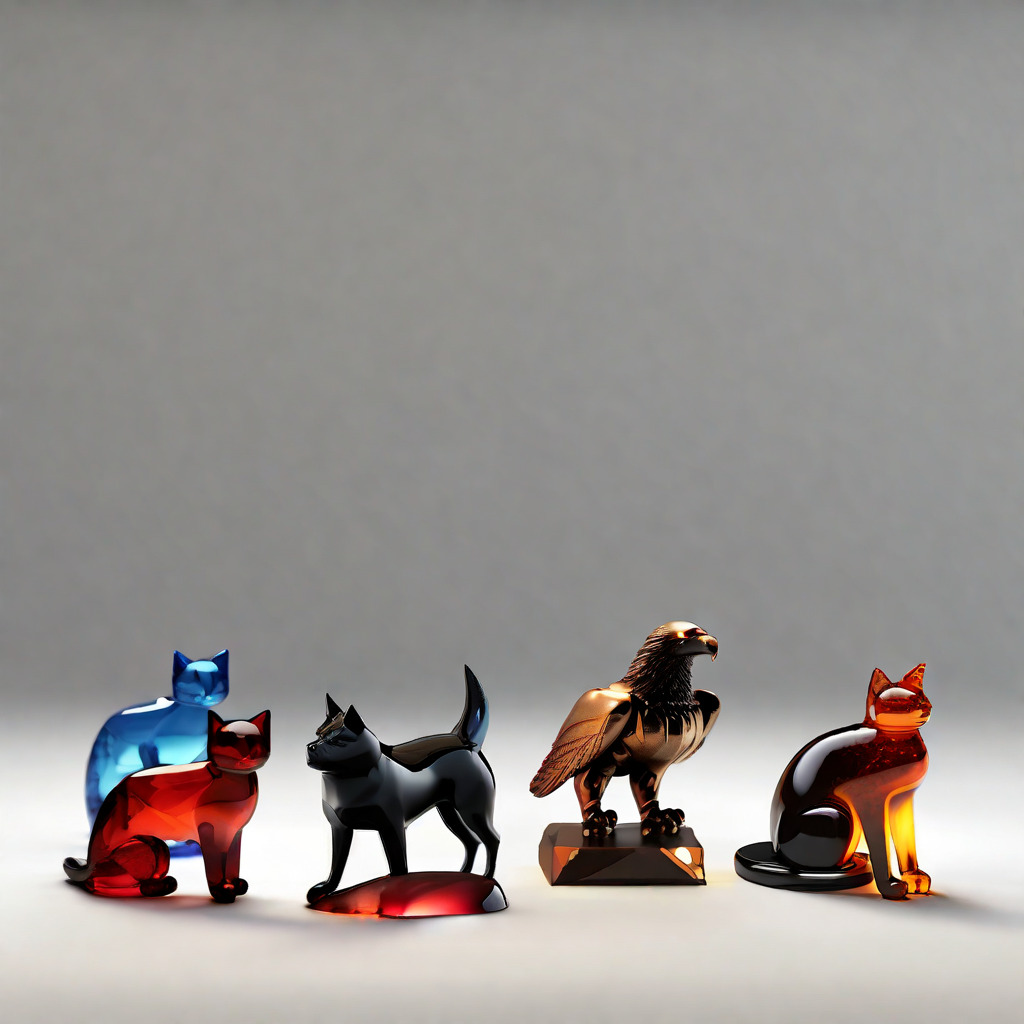}}
\end{tabular} &

\begin{tabular}{p{5cm}}
\multicolumn{1}{c}{\includegraphics[width=0.133\linewidth]{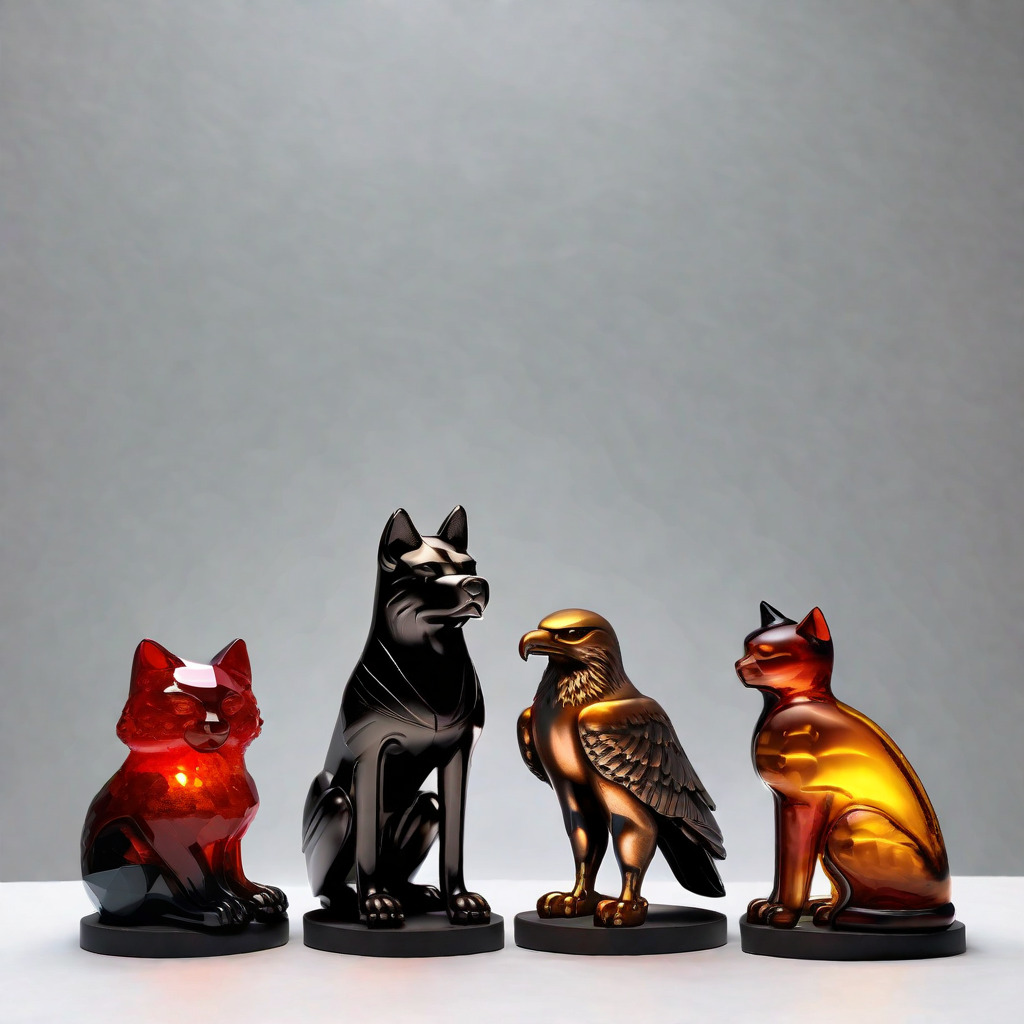}} \\
\multicolumn{1}{c}{\includegraphics[width=0.133\linewidth]{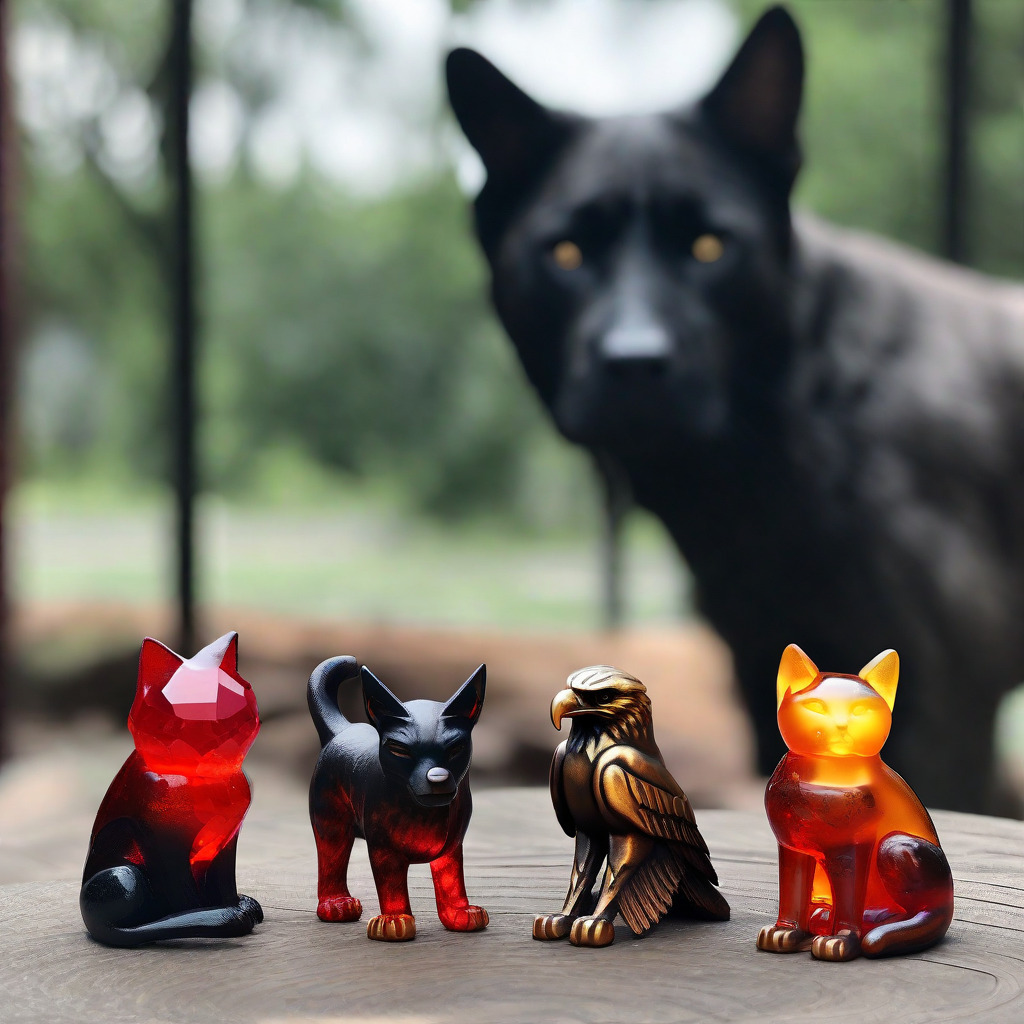}}
\end{tabular} &

\begin{tabular}{p{5cm}}
\multicolumn{1}{c}{\includegraphics[width=0.133\linewidth]{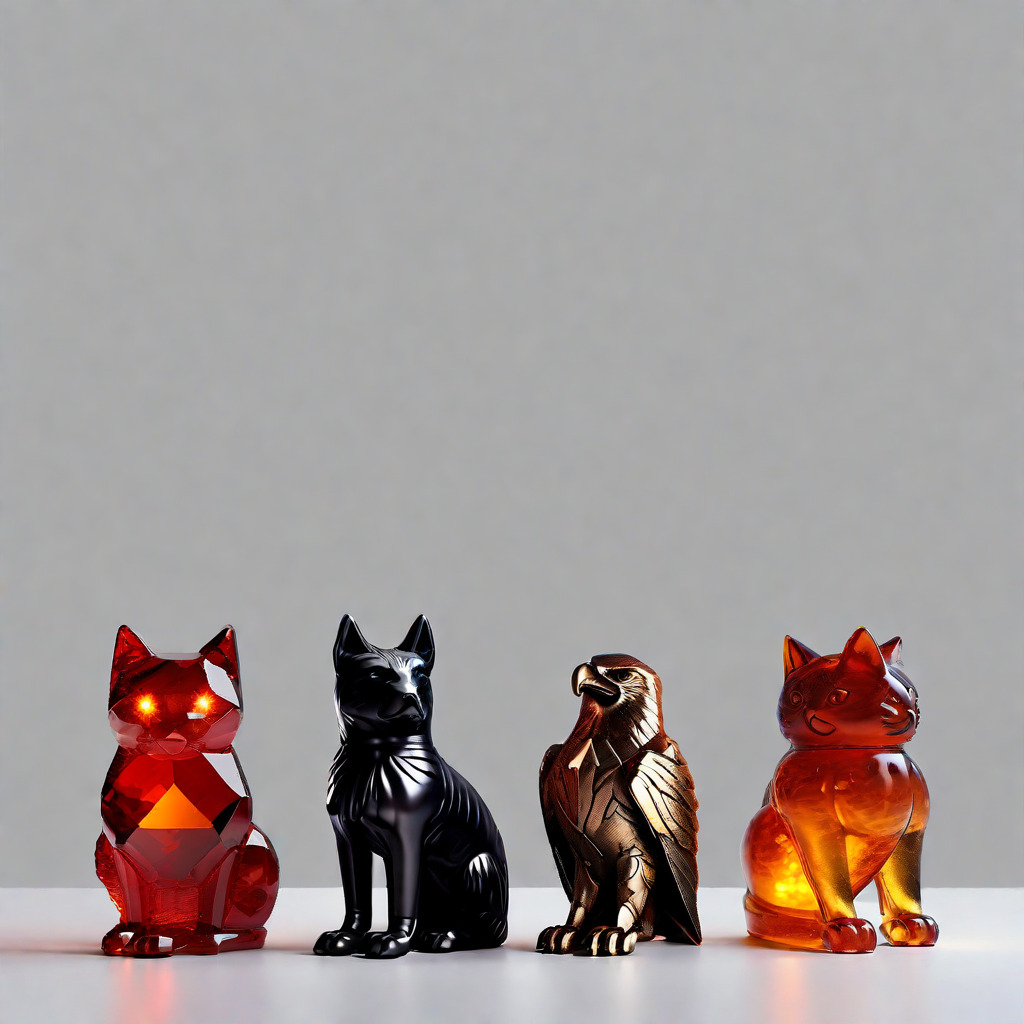}} \\
\multicolumn{1}{c}{\includegraphics[width=0.133\linewidth]{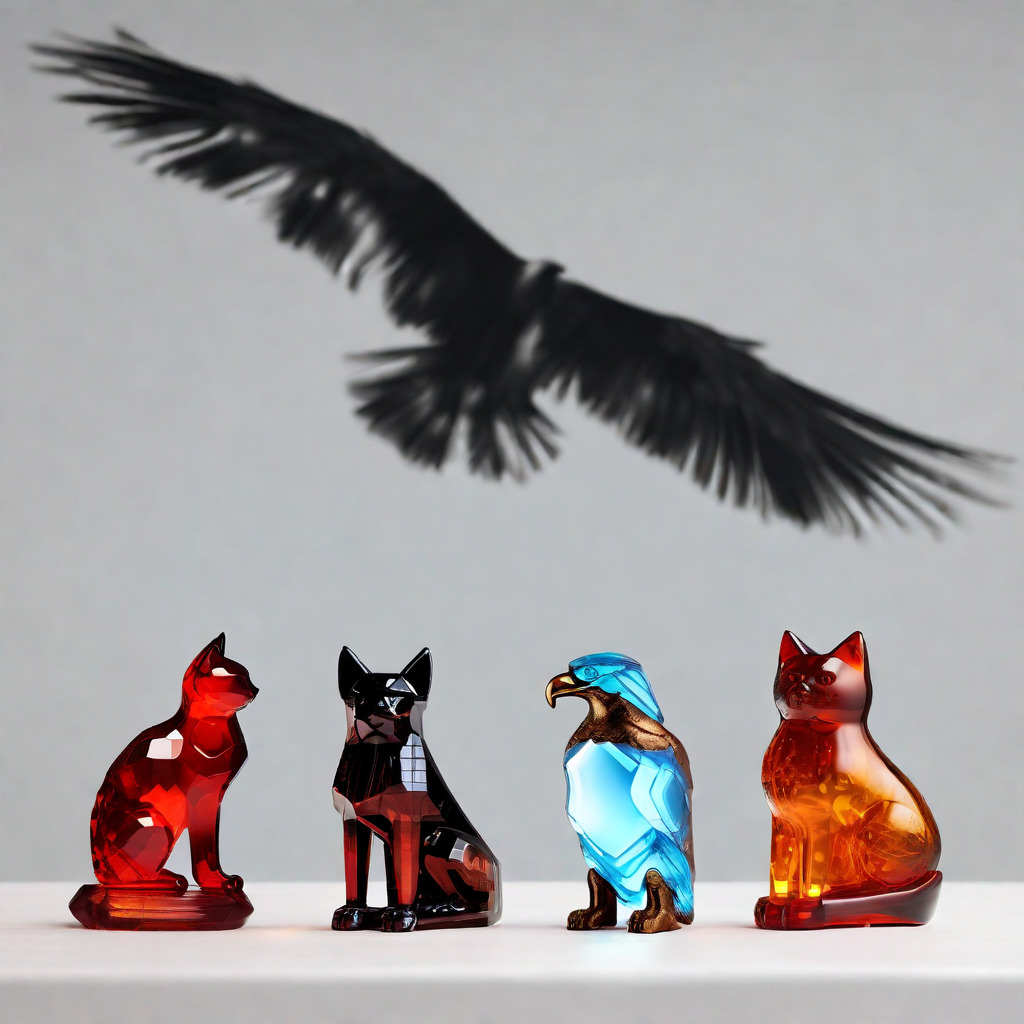}}
\end{tabular} &

\begin{tabular}{p{5cm}}
\multicolumn{1}{c}{\includegraphics[width=0.133\linewidth]{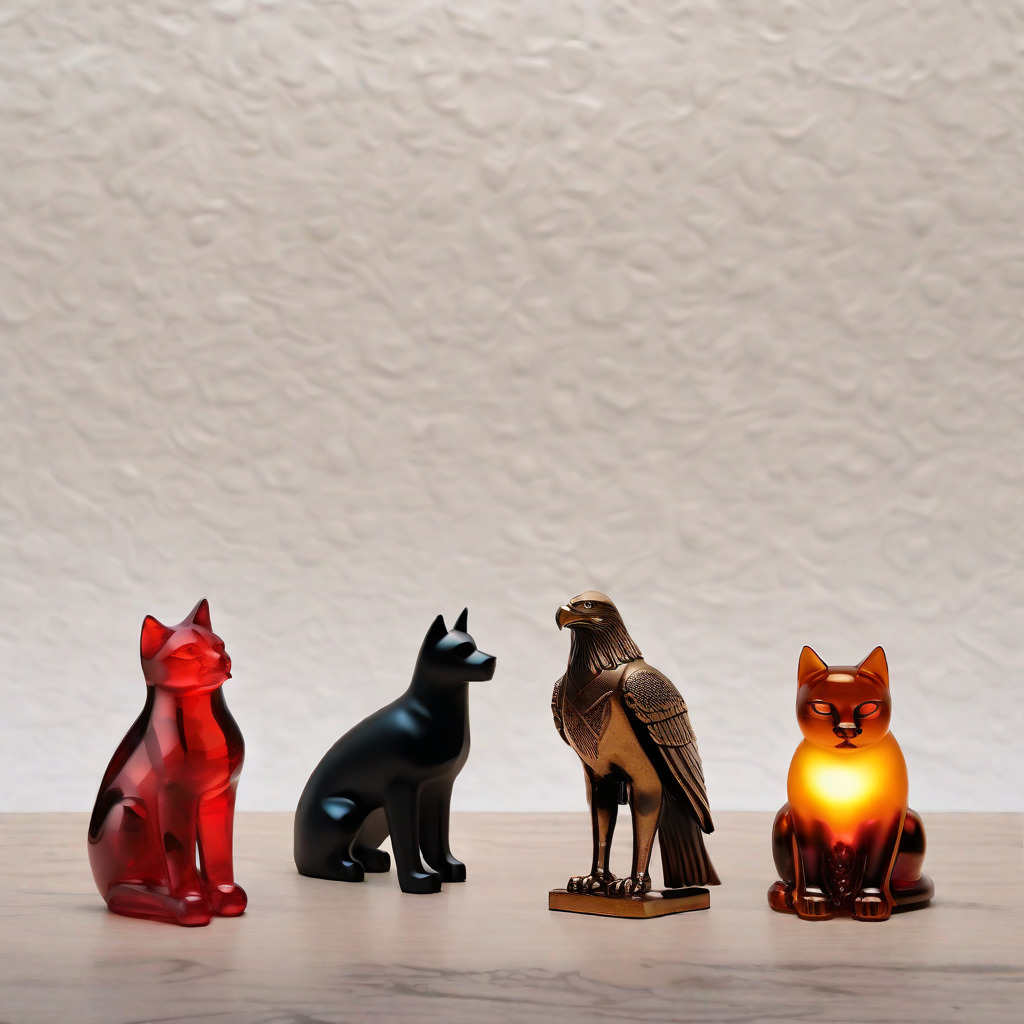}} \\
\multicolumn{1}{c}{\includegraphics[width=0.133\linewidth]{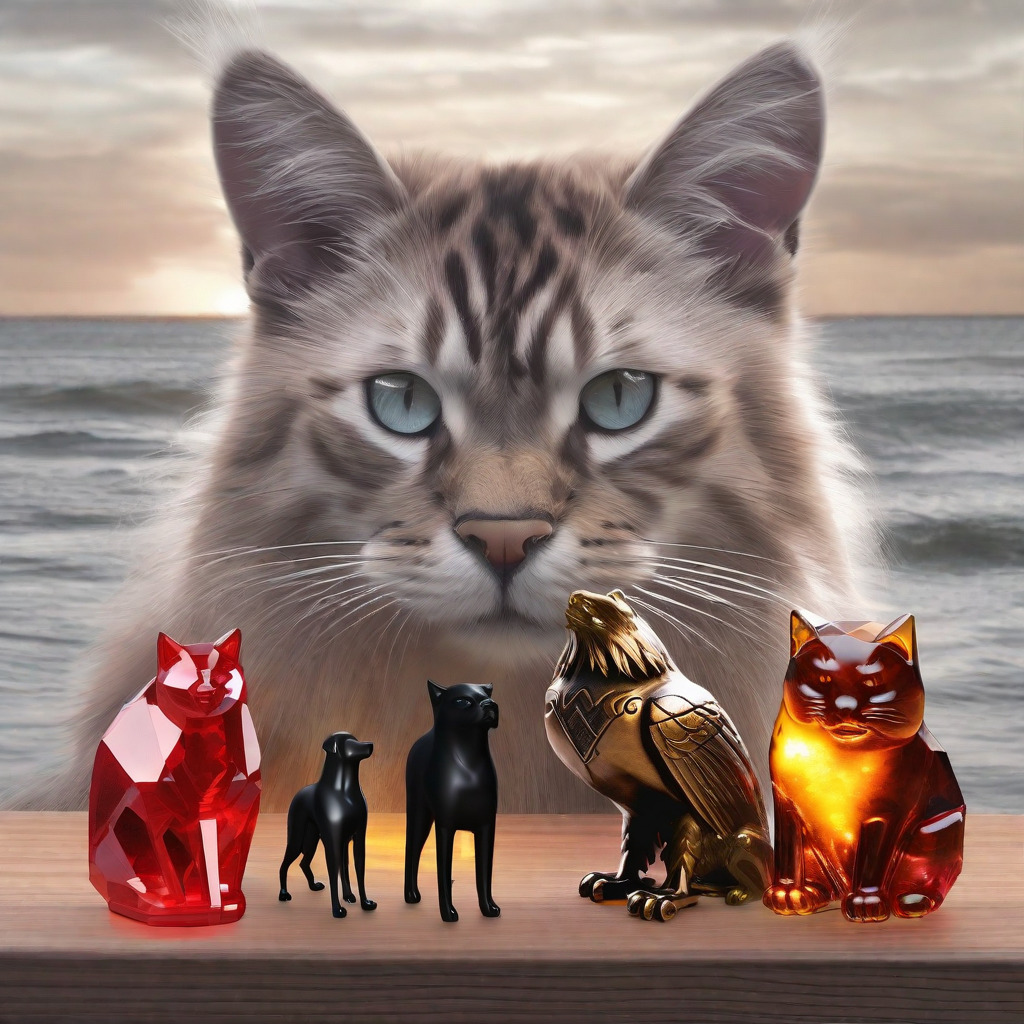}}
\end{tabular} &

\begin{tabular}{p{5cm}}
\multicolumn{1}{p{3.5cm}}{\centering 2. A realistic photo of a \textbf{\color{red} shiny red crystal cat} and a \textbf{\color{black} black matte plastic dog} and a \textbf{\color{brown} rustic bronze eagle} and a \textbf{\color{orange} glowing amber cat}.} \\
\multicolumn{1}{c}{\includegraphics[width=0.133\linewidth]{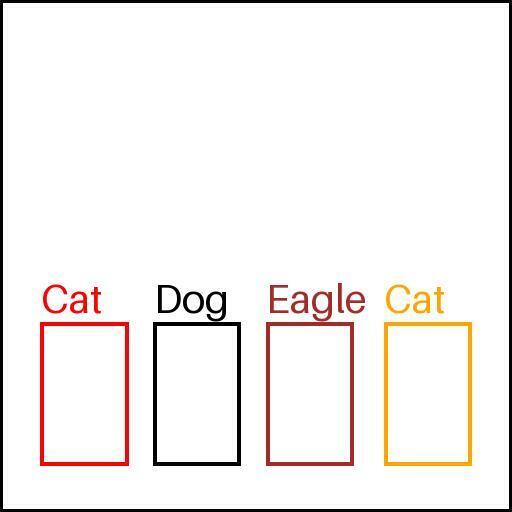}}
\end{tabular}

\\
\midrule

\begin{tabular}{p{0.2cm}}
\raisebox{0.15cm}[\height][\depth]{\rotatebox{90}{\modelshort{}}} \\
\raisebox{-0.85cm}[\height][\depth]{\rotatebox{90}{Bounded Attention}}
\end{tabular} &
\begin{tabular}{p{5cm}}
\multicolumn{1}{c}{\includegraphics[width=0.133\linewidth]{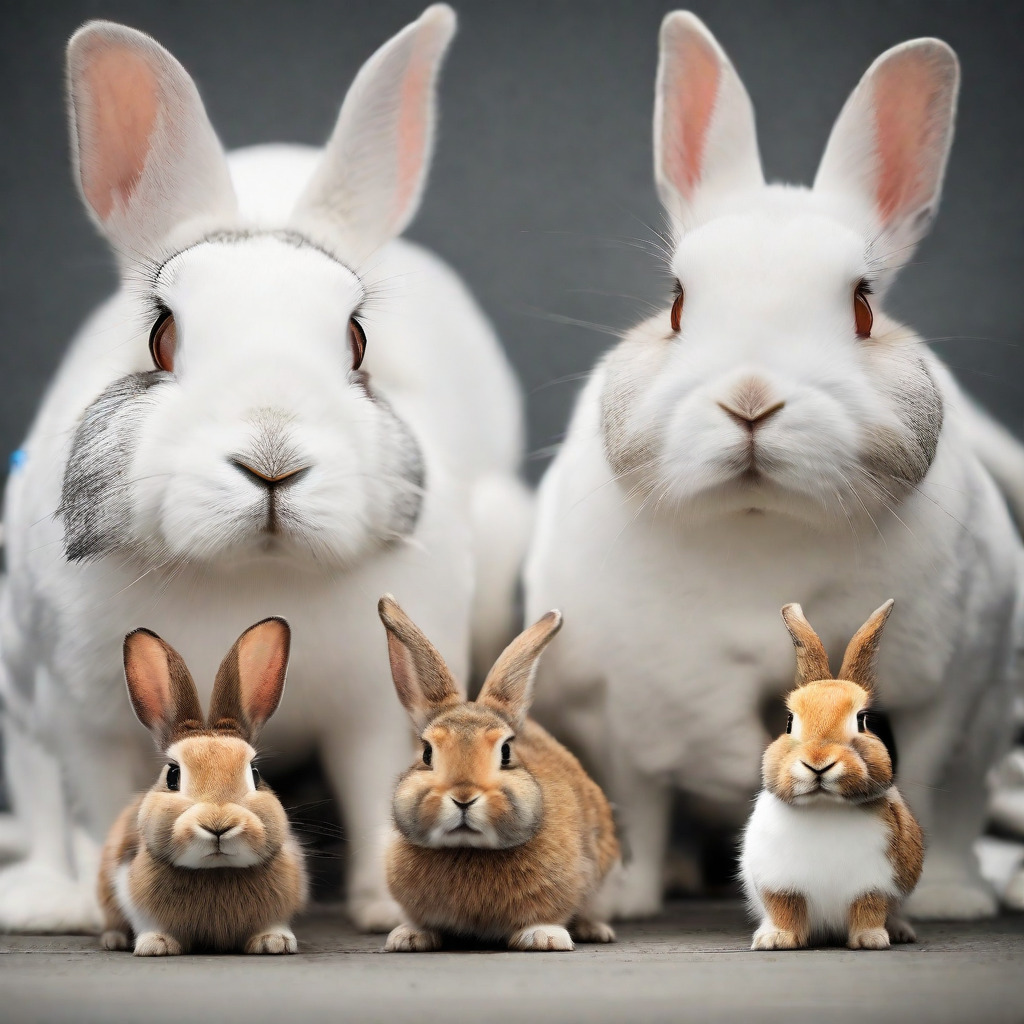}} \\
\multicolumn{1}{c}{\includegraphics[width=0.133\linewidth]{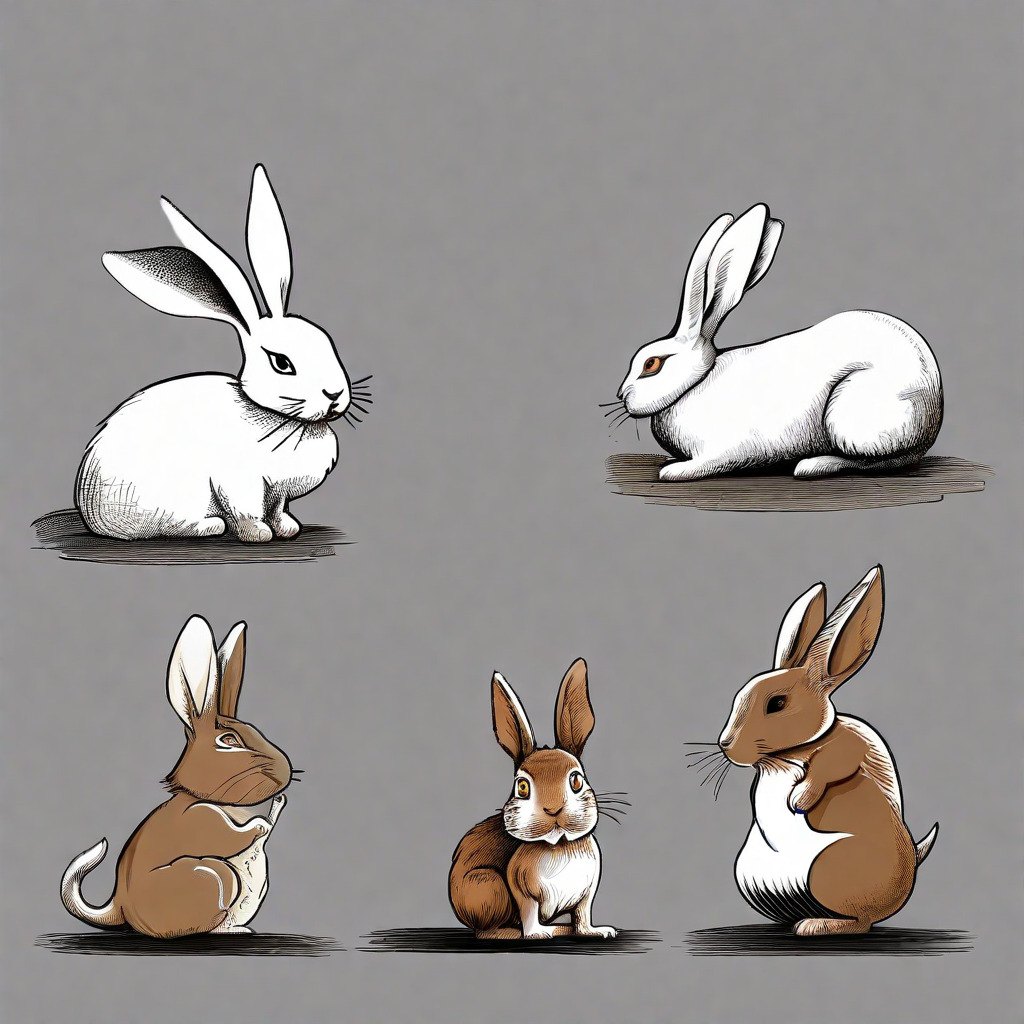}}
\end{tabular} &

\begin{tabular}{p{5cm}}
\multicolumn{1}{c}{\includegraphics[width=0.133\linewidth]{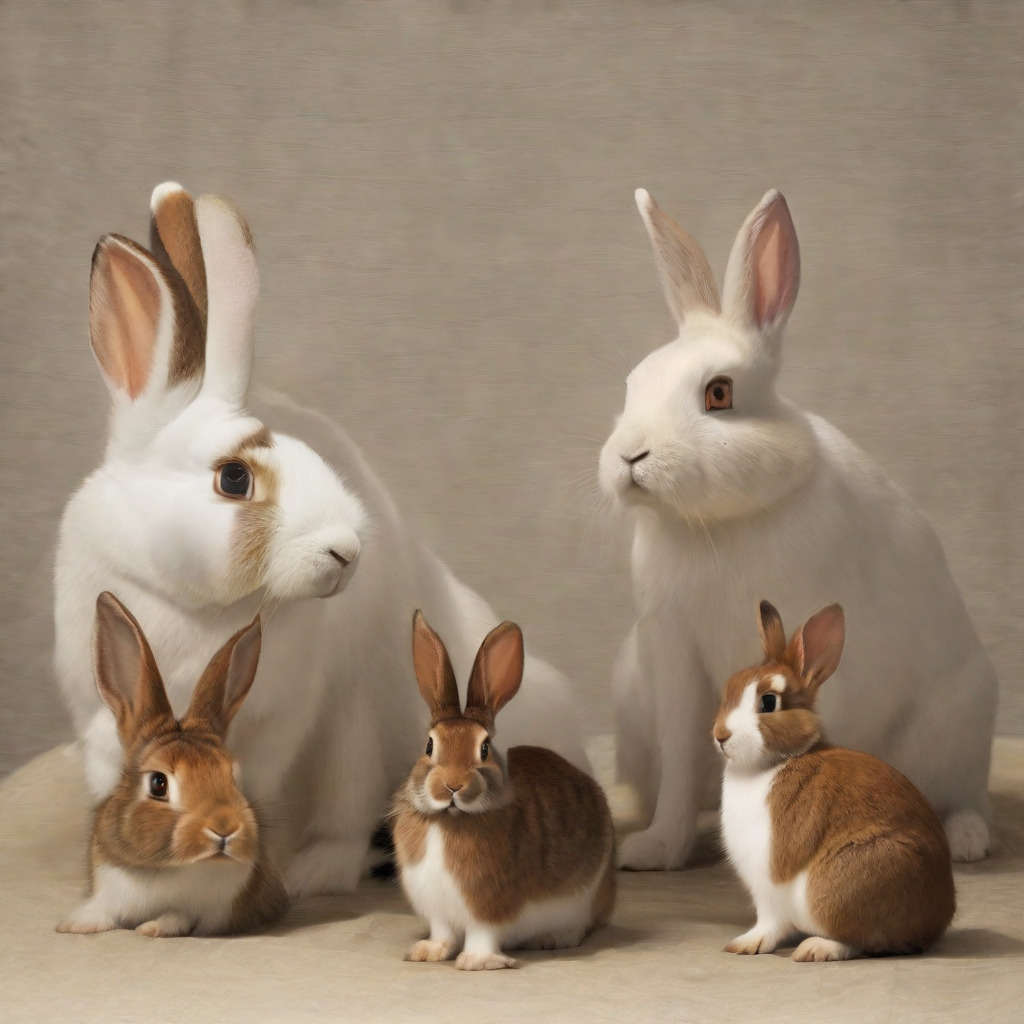}} \\
\multicolumn{1}{c}{\includegraphics[width=0.133\linewidth]{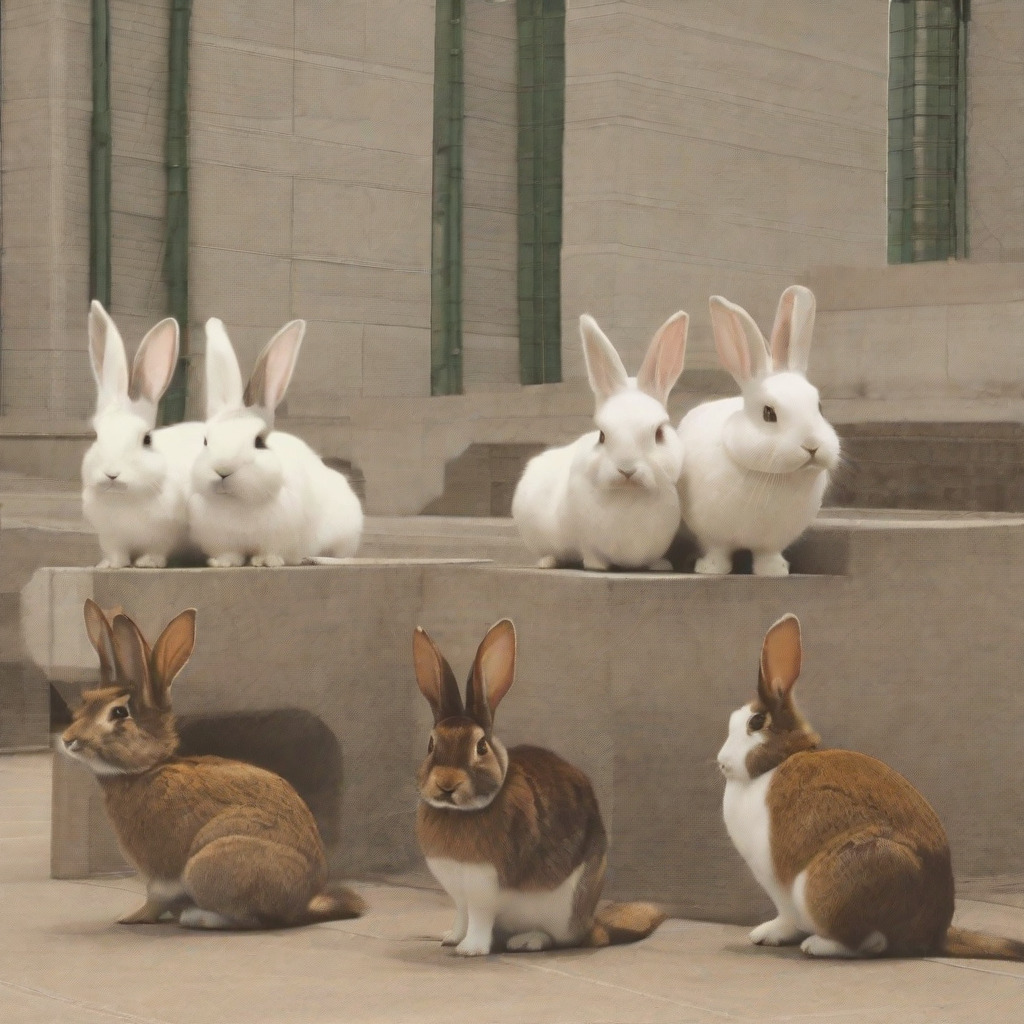}}
\end{tabular} &

\begin{tabular}{p{5cm}}
\multicolumn{1}{c}{\includegraphics[width=0.133\linewidth]{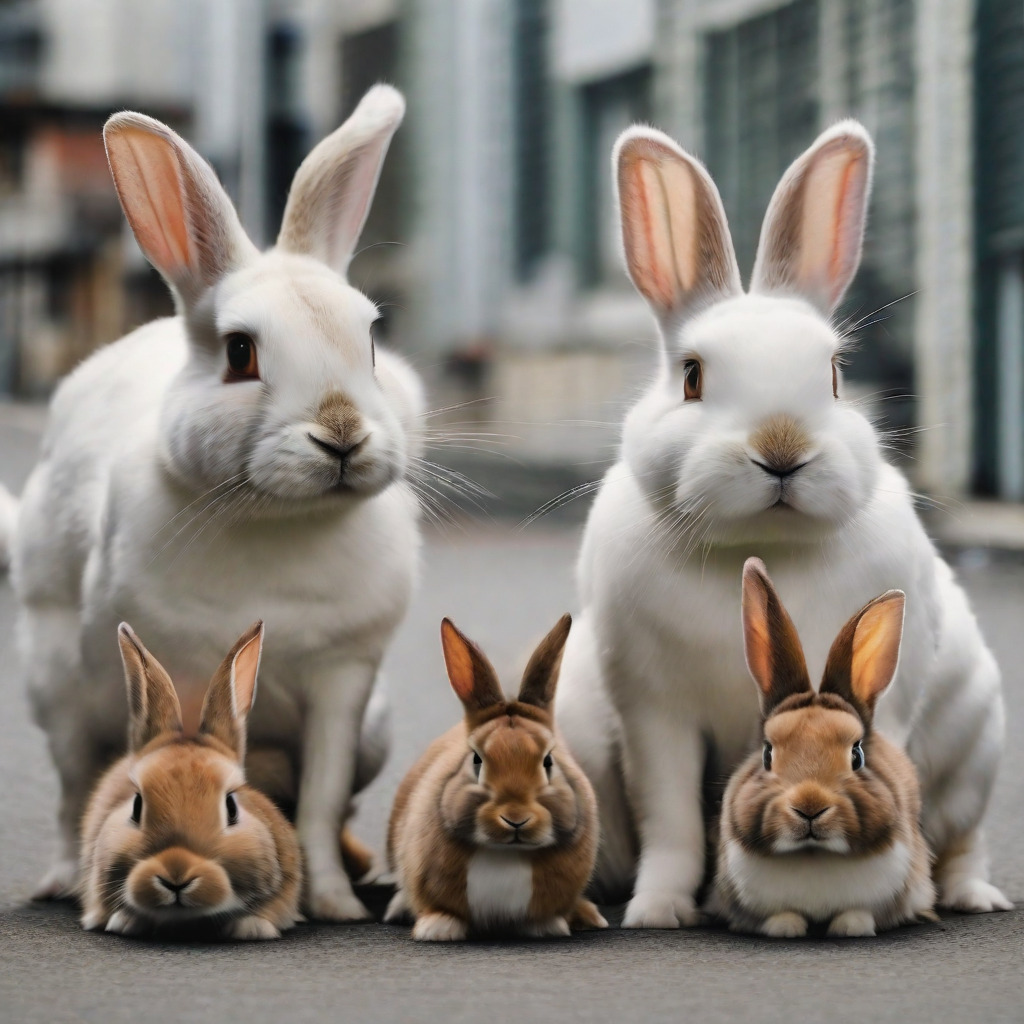}} \\
\multicolumn{1}{c}{\includegraphics[width=0.133\linewidth]{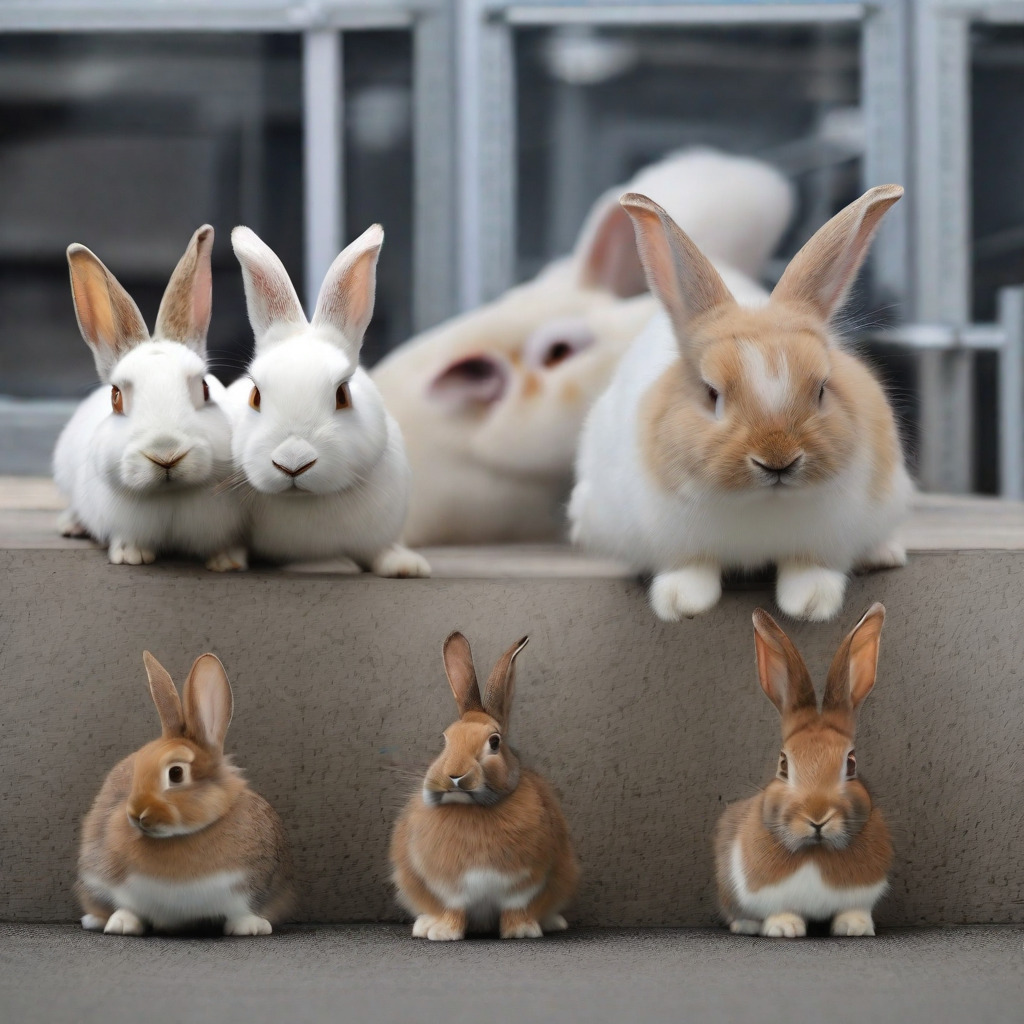}}
\end{tabular} &

\begin{tabular}{p{5cm}}
\multicolumn{1}{c}{\includegraphics[width=0.133\linewidth]{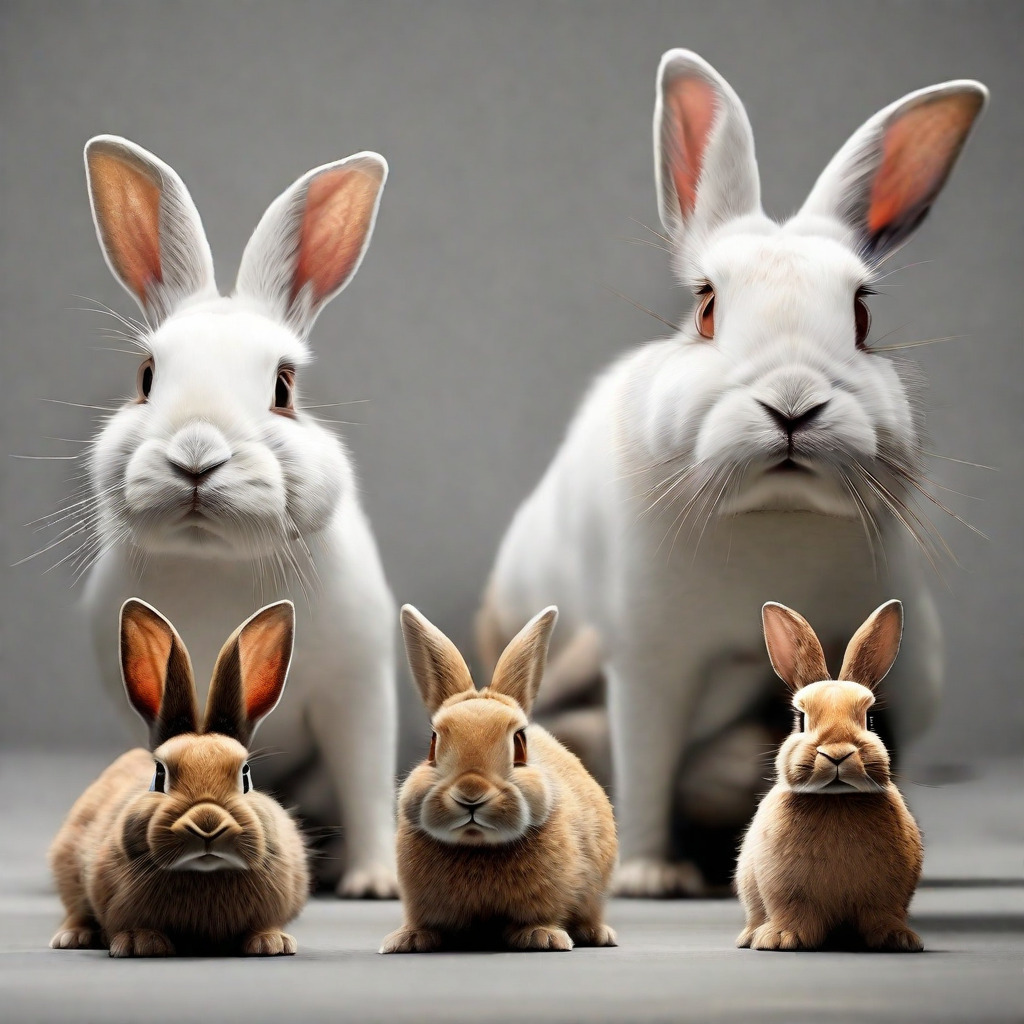}} \\
\multicolumn{1}{c}{\includegraphics[width=0.133\linewidth]{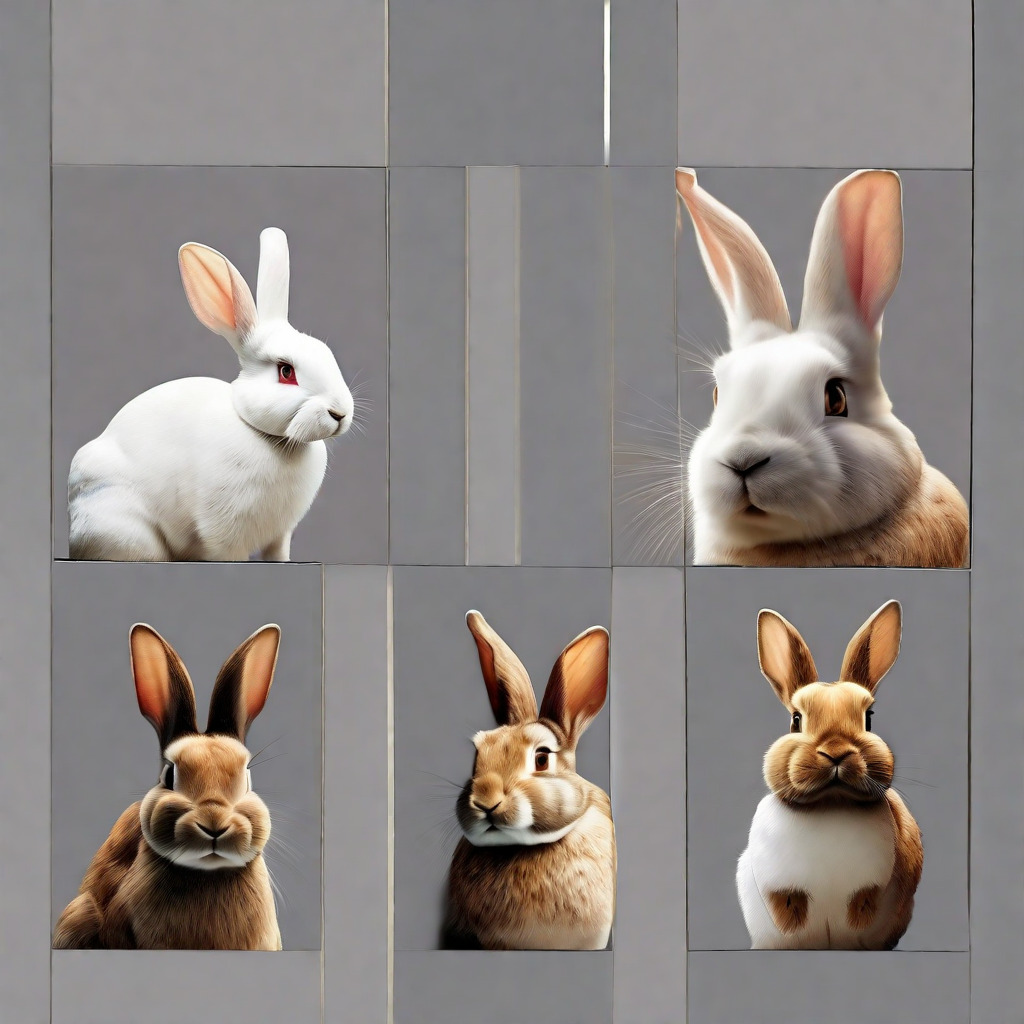}}
\end{tabular} &

\begin{tabular}{p{5cm}}
\multicolumn{1}{p{3.5cm}}{\centering 3. \textbf{\color{brown} Three brown rabbits} and \textbf{\color{gray} two white rabbits}}. \\
\multicolumn{1}{c}{\includegraphics[width=0.133\linewidth]{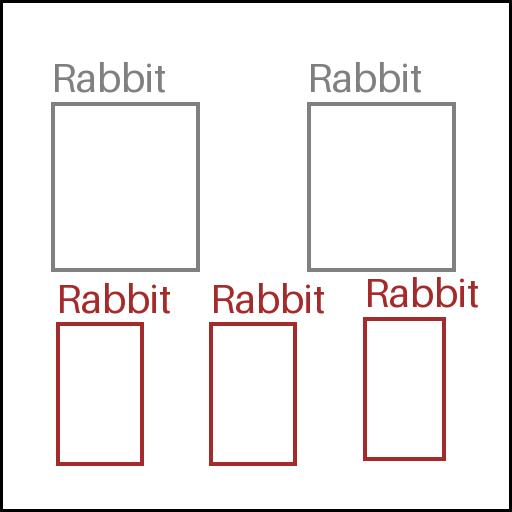}}
\end{tabular}

\\
\midrule

\begin{tabular}{p{0.2cm}}
\raisebox{0.15cm}[\height][\depth]{\rotatebox{90}{\modelshort{}}} \\
\raisebox{-0.85cm}[\height][\depth]{\rotatebox{90}{Bounded Attention}}
\end{tabular} &

\begin{tabular}{p{5cm}}
\multicolumn{1}{c}{\includegraphics[width=0.133\linewidth]{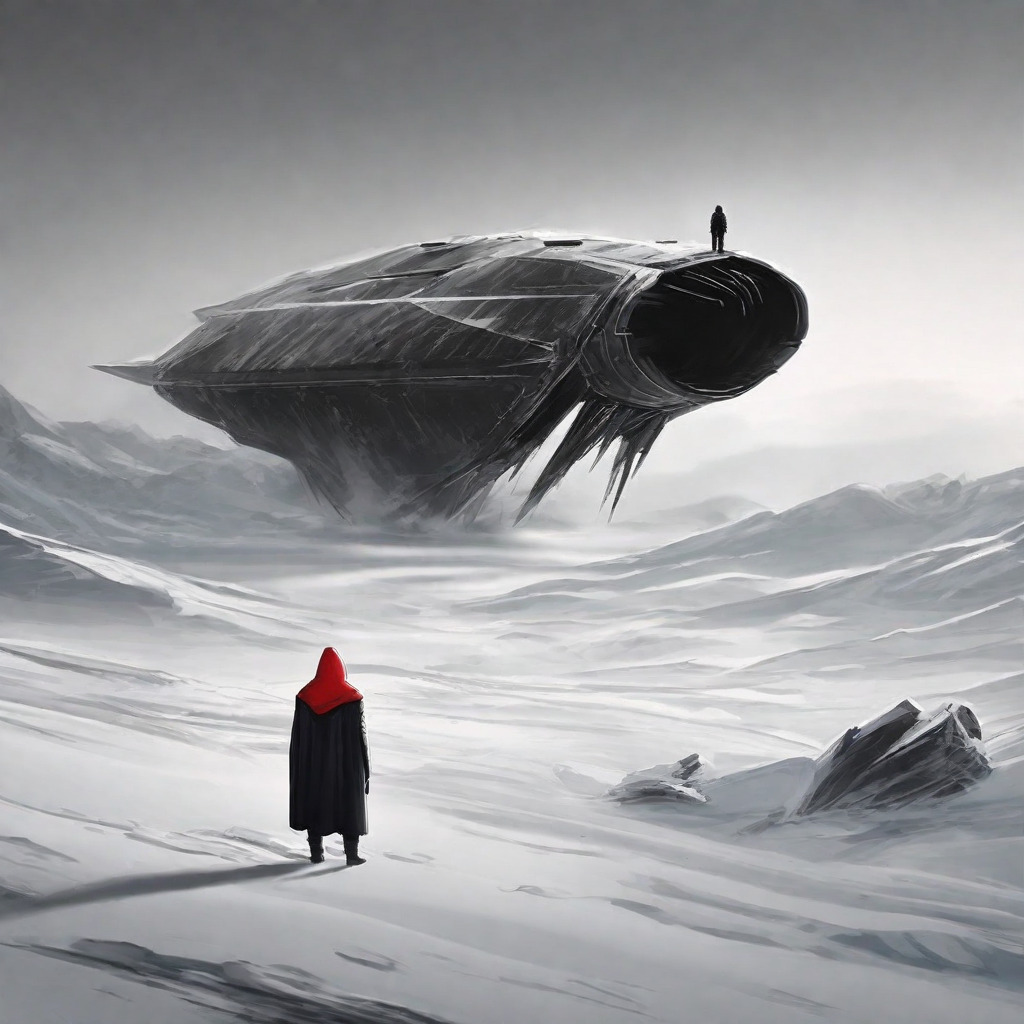}} \\
\multicolumn{1}{c}{\includegraphics[width=0.133\linewidth]{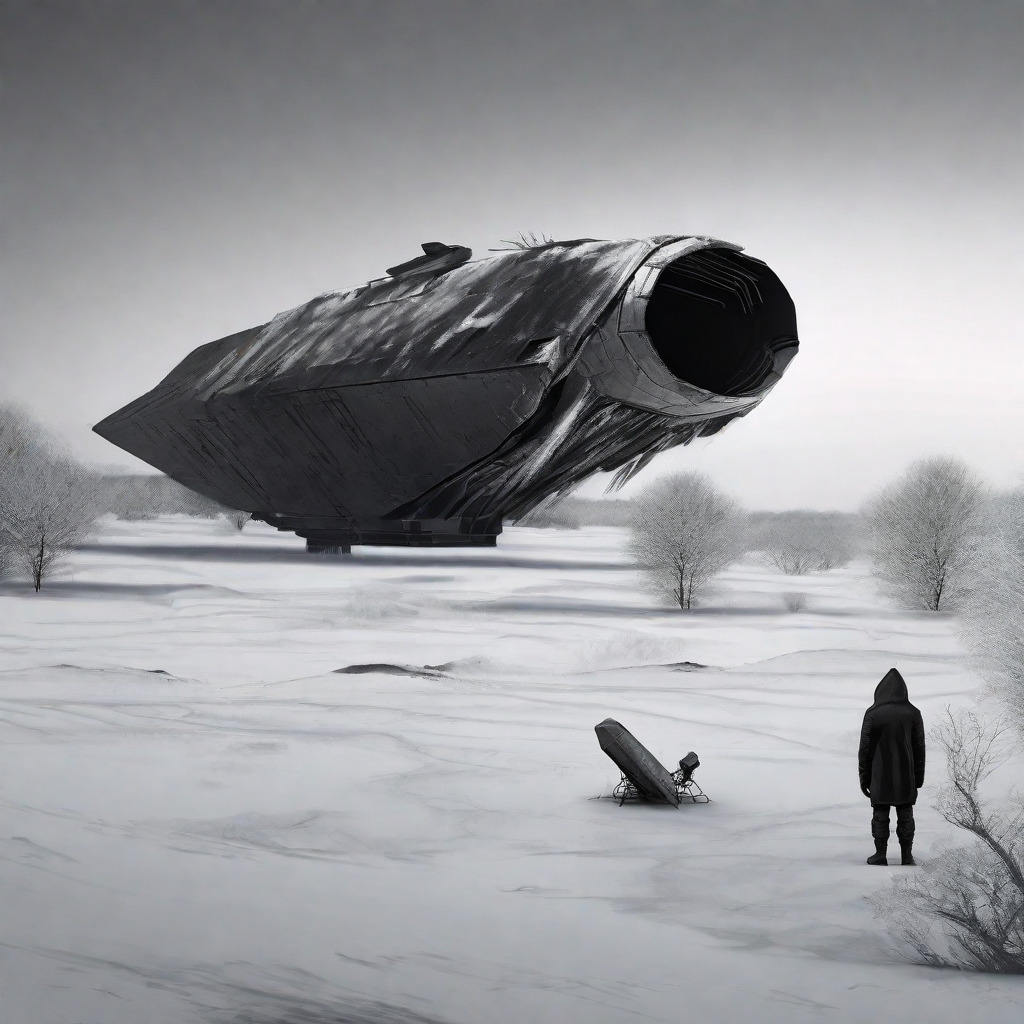}}
\end{tabular} &

\begin{tabular}{p{5cm}}
\multicolumn{1}{c}{\includegraphics[width=0.133\linewidth]{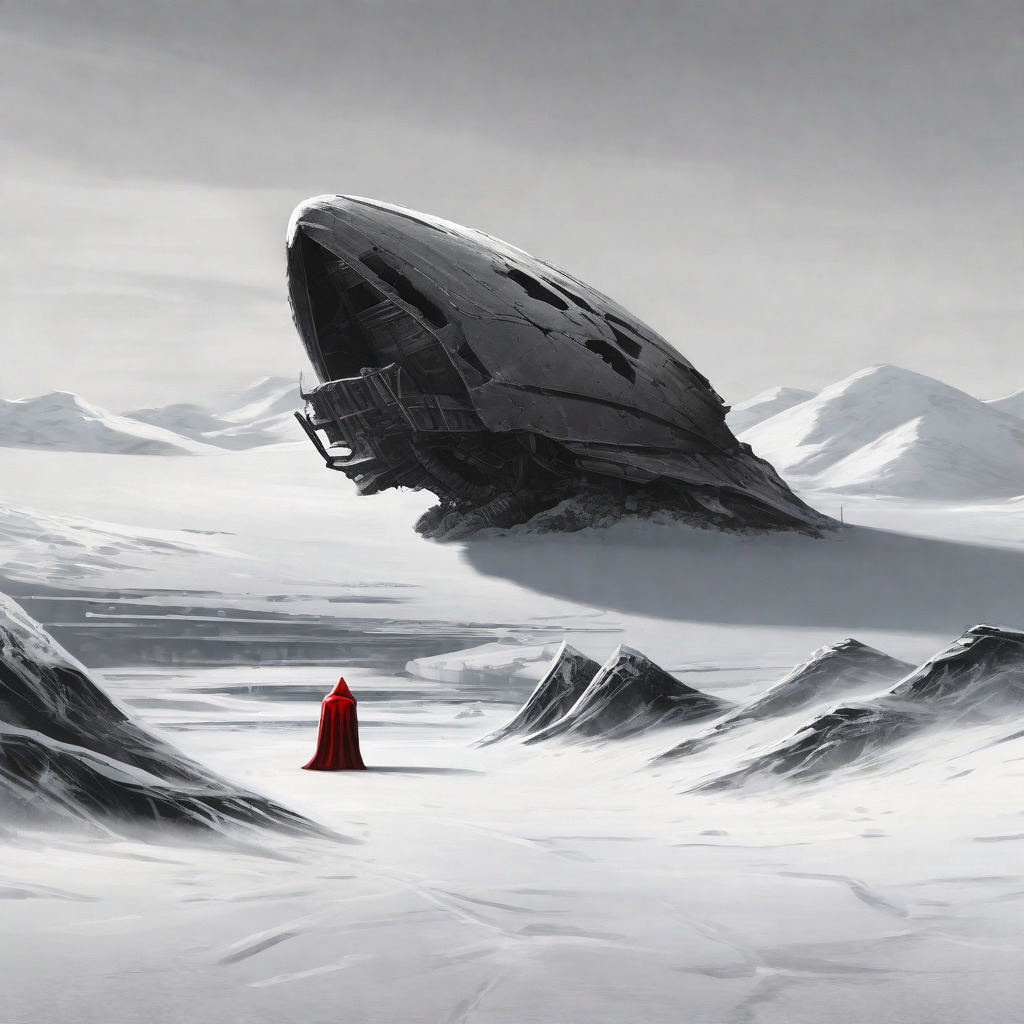}} \\
\multicolumn{1}{c}{\includegraphics[width=0.133\linewidth]{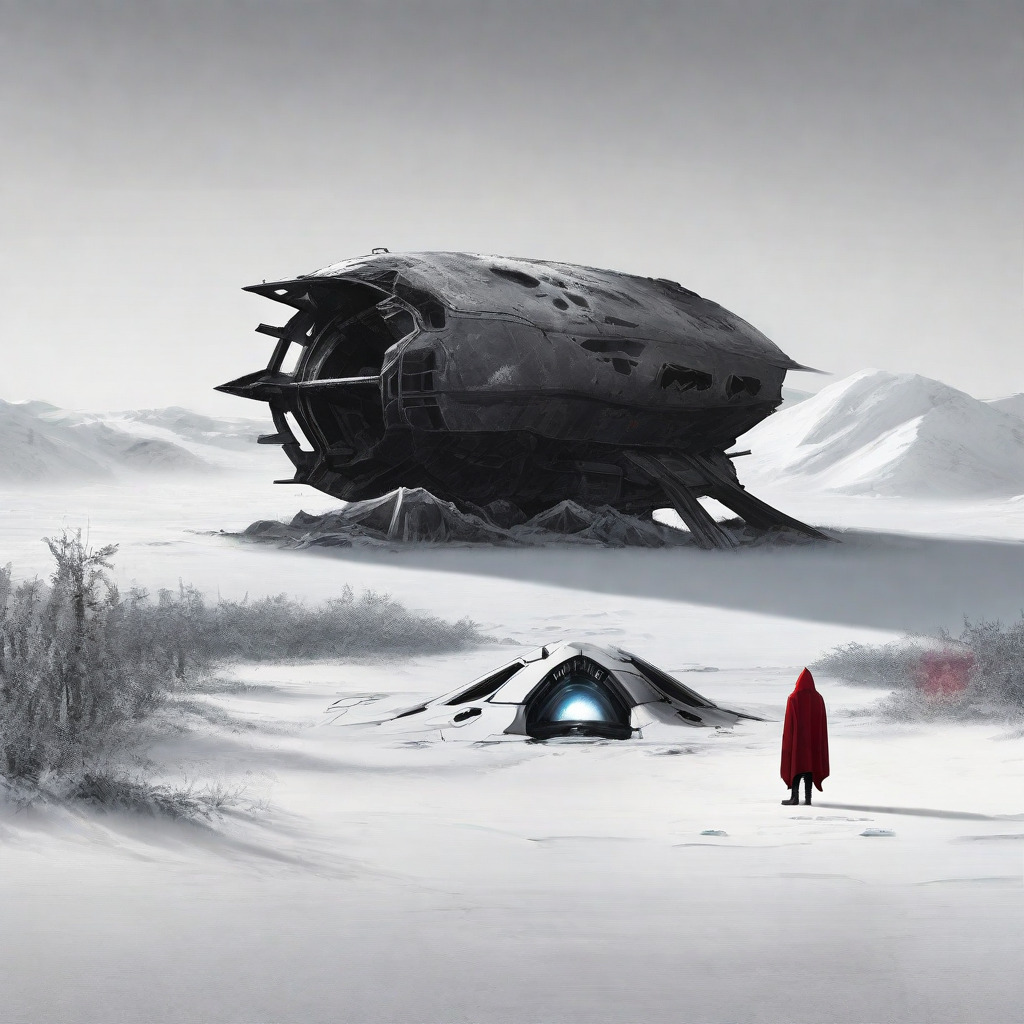}}
\end{tabular} &

\begin{tabular}{p{5cm}}
\multicolumn{1}{c}{\includegraphics[width=0.133\linewidth]{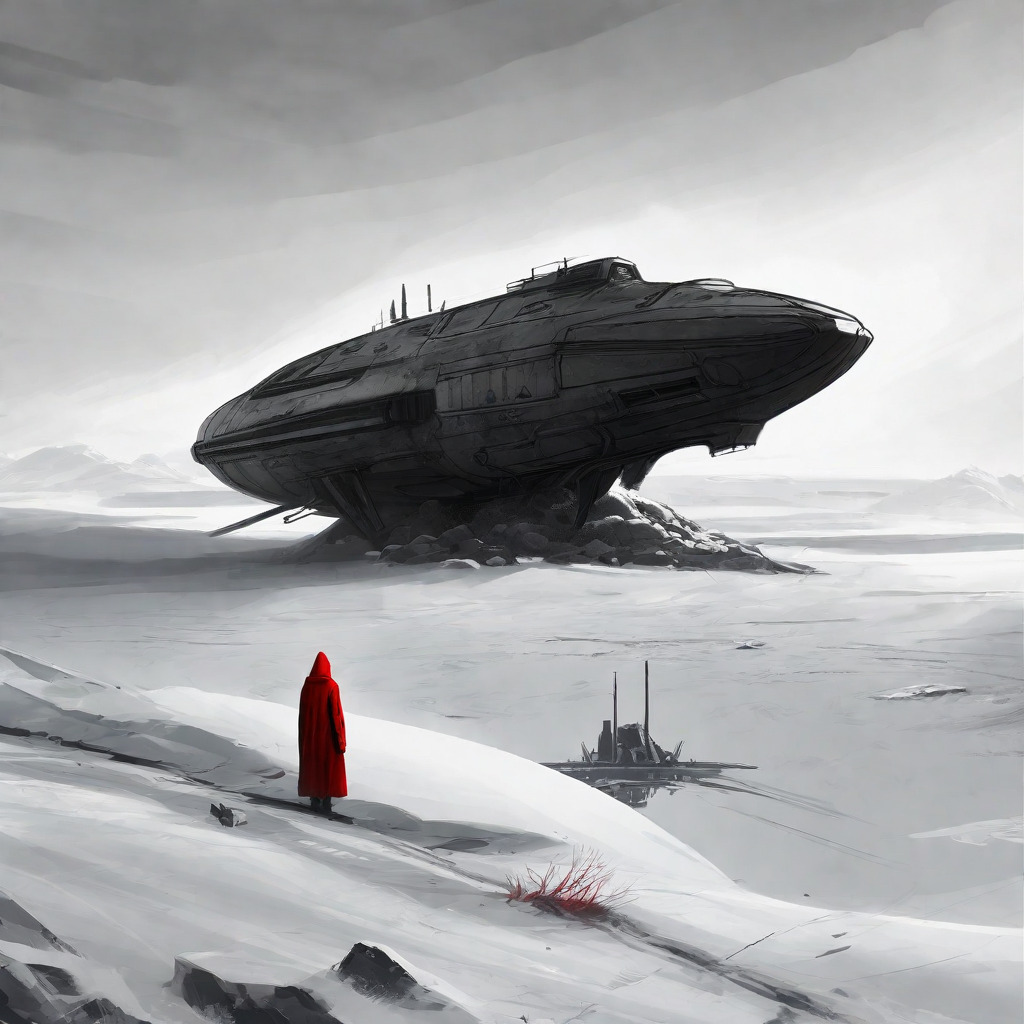}} \\
\multicolumn{1}{c}{\includegraphics[width=0.133\linewidth]{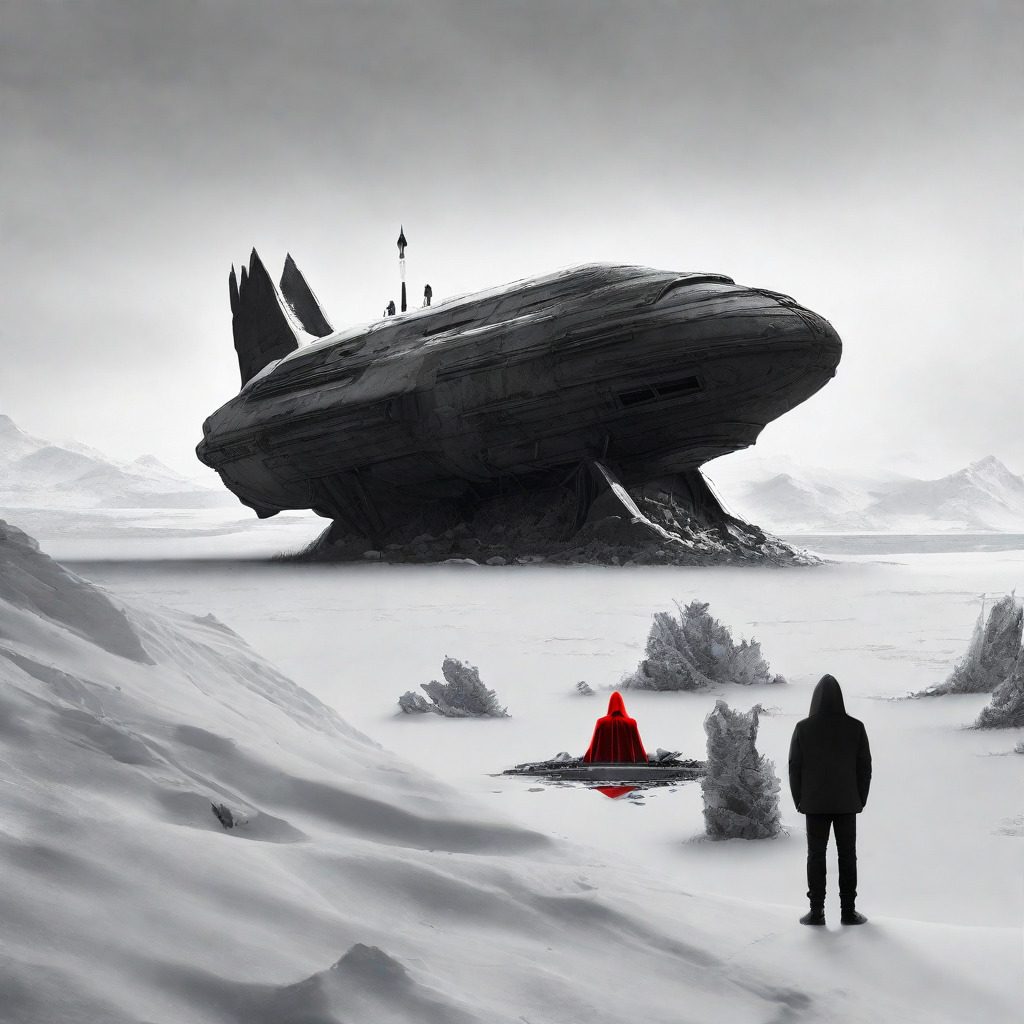}}
\end{tabular} &

\begin{tabular}{p{5cm}}
\multicolumn{1}{c}{\includegraphics[width=0.133\linewidth]{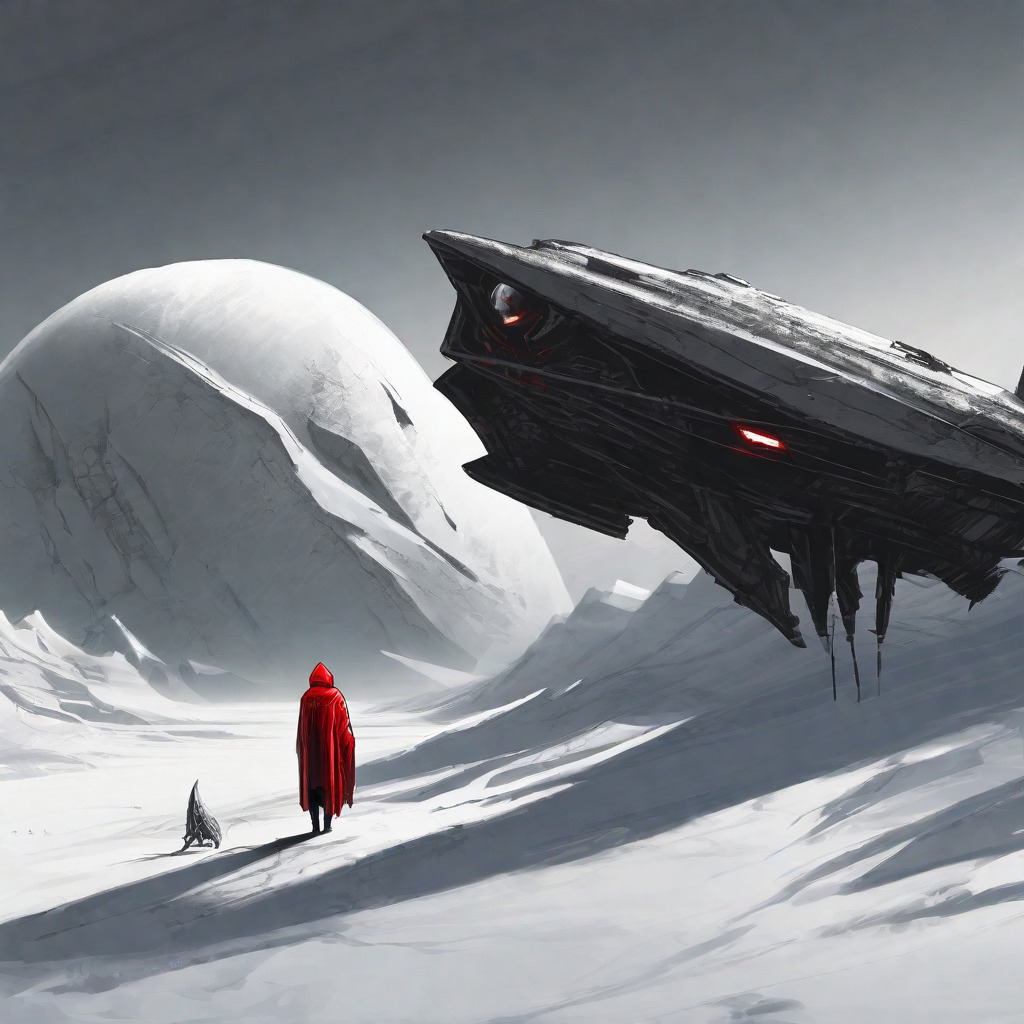}} \\
\multicolumn{1}{c}{\includegraphics[width=0.133\linewidth]{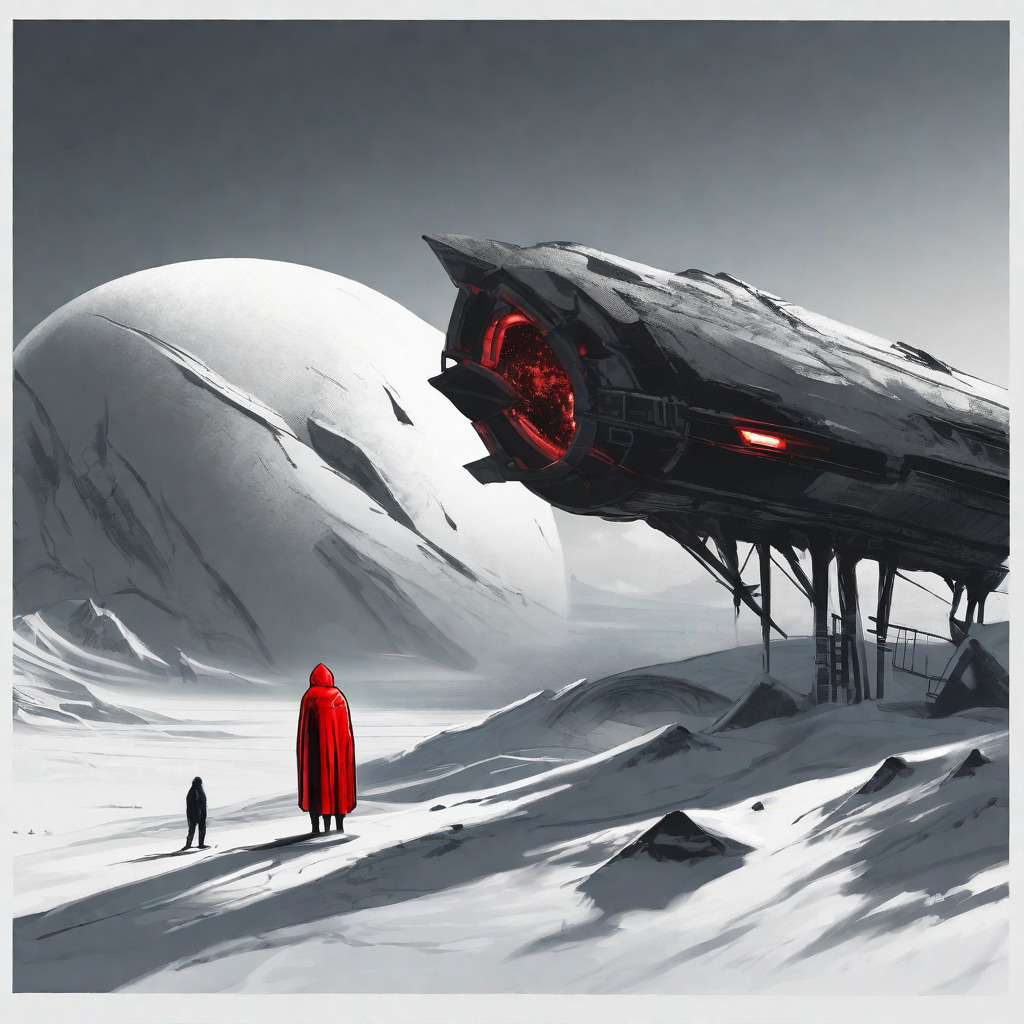}}
\end{tabular} &

\begin{tabular}{p{5cm}}
\multicolumn{1}{p{3.5cm}}{\centering 4. A black and white concept art of a \textbf{\color{gray} crashed spaceship} partially buried in icy landscape and a \textbf{\color{red} red hooded person} is watching it from a distance.} \\
\multicolumn{1}{c}{\includegraphics[width=0.133\linewidth]{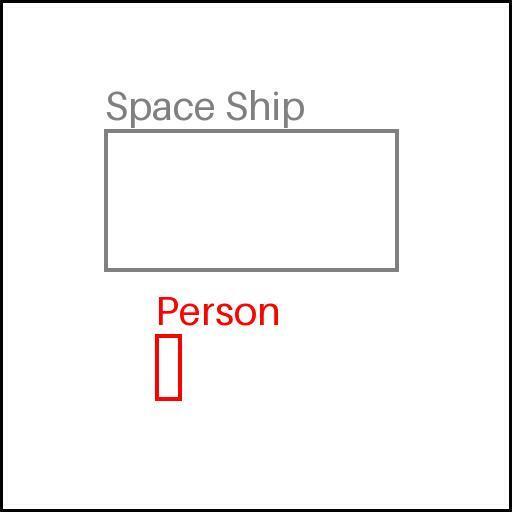}}
\end{tabular}

\\

\end{tabular}
\centering
\vspace{-3mm}
\caption{
\textbf{Effect of random seeds.}
We present 4 images for the same input text and layout prompt only differing based on the initial random seed.
Within a column, both Bounded Attention and \modelshort{} use the same seed.
We see that \modelshort{} generates more accurate images while Bounded Attention suffers many problems mentioned in the paper.
Background semantic leakage is prominently seen in 3 of 4 seeds for prompt 2,
unnatural out-of-distribution artifacts for all four generations of prompt 3 (rabbits), and
erroneous attributes are seen across multiple generations.
}
\label{fig:misc_outputs1}
\end{figure*}

%% file: figures/supp_visuals_page2.tex
\begin{figure*}[t]
\tabcolsep=0.05cm
\small
\begin{tabular}{cc:cc}

\begin{tabular}{p{5cm}}
\multicolumn{1}{p{3.5cm}}{\centering A \textbf{\color{orange} monk wearing orange robes} and and his \textbf{\color{red} dog} together on cliff side and \textbf{\color{pink} pink cherry blossoms} in japanese sumi-e ink style painting} \\
\multicolumn{1}{c}{\includegraphics[width=0.15\linewidth]{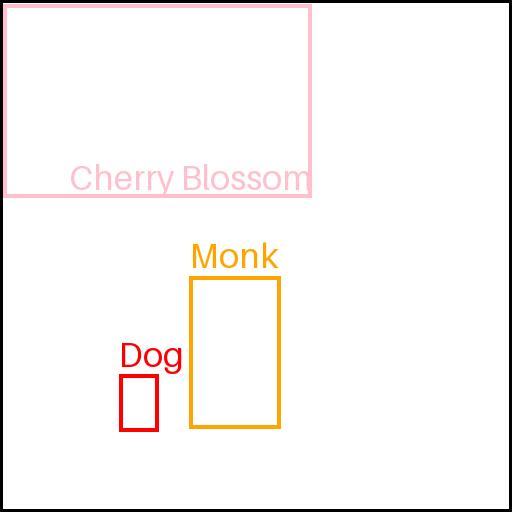}}
\end{tabular} &

\begin{tabular}{p{5cm}}
\includegraphics[width=0.9\linewidth]{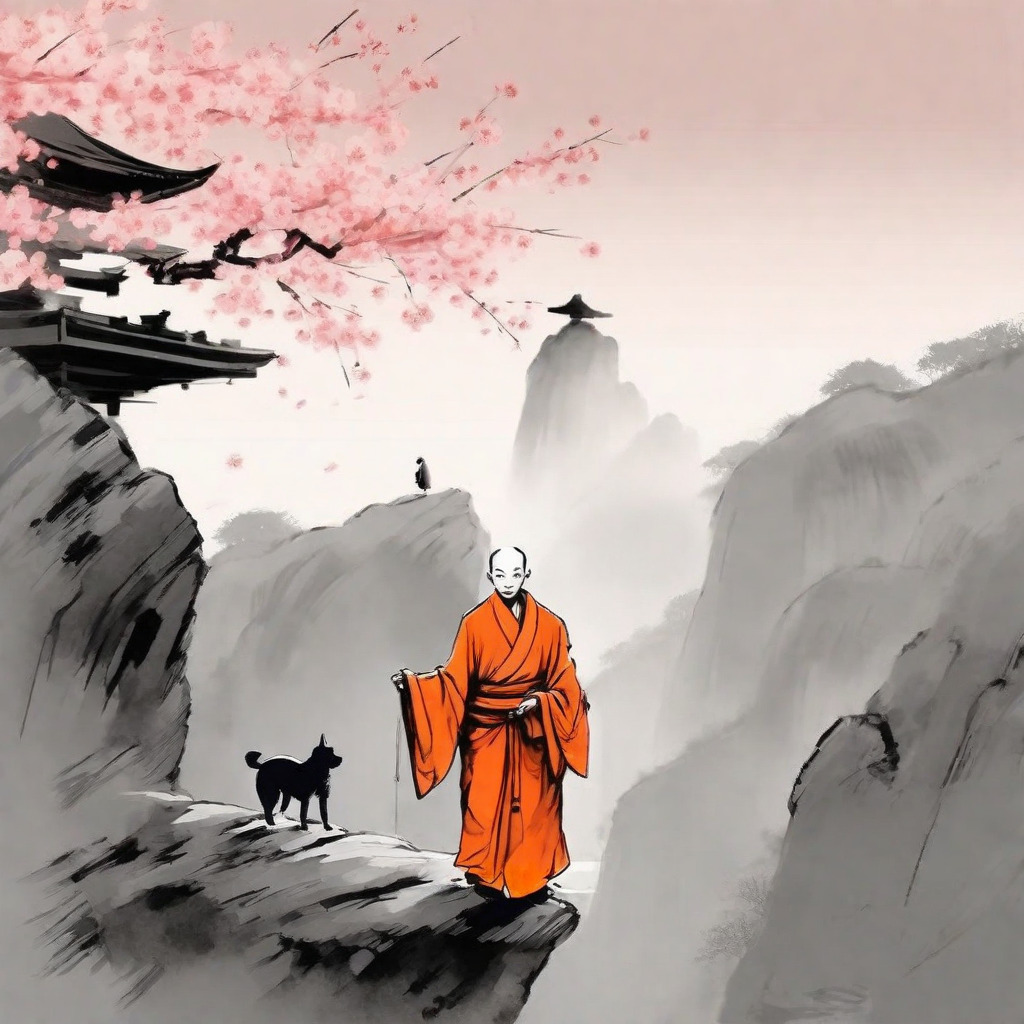}
\end{tabular} &

\begin{tabular}{p{5cm}}
\multicolumn{1}{p{4cm}}{\centering An ancient japanese calligraphy sketch of a \textbf{\color{blue} blue mystic smoke} coming out of a small \textbf{\color{gray} metallic lamp} and a \textbf{\color{orange} monk wearing orange robe} sitting nearby} \\
\multicolumn{1}{c}{\includegraphics[width=0.15\linewidth]{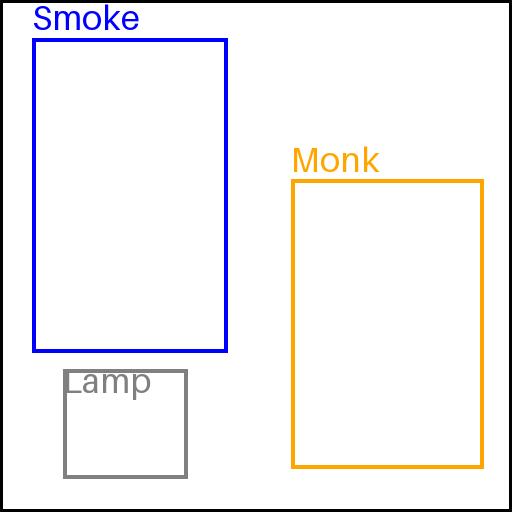}}
\end{tabular} &

\begin{tabular}{p{5cm}}
\includegraphics[width=0.9\linewidth]{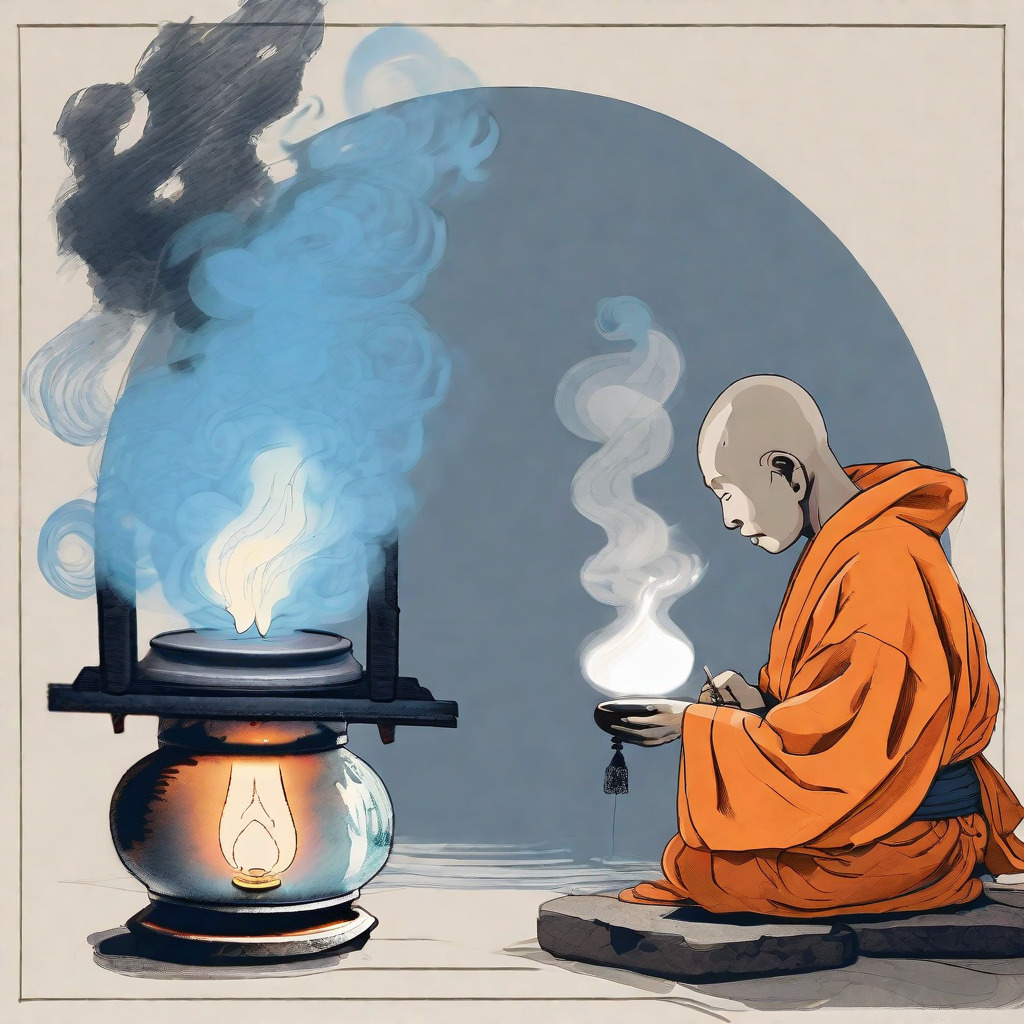}
\end{tabular}
\\
& & & \\
& & & \\

\begin{tabular}{p{5cm}}
\multicolumn{1}{p{3.5cm}}{\centering A concept art of a icy landscape with a \textbf{\color{red} robe} wizard standing far away firing his wand in air for a \textbf{\color{blue} black colored} magic thunderbolt} \\
\multicolumn{1}{c}{\includegraphics[width=0.15\linewidth]{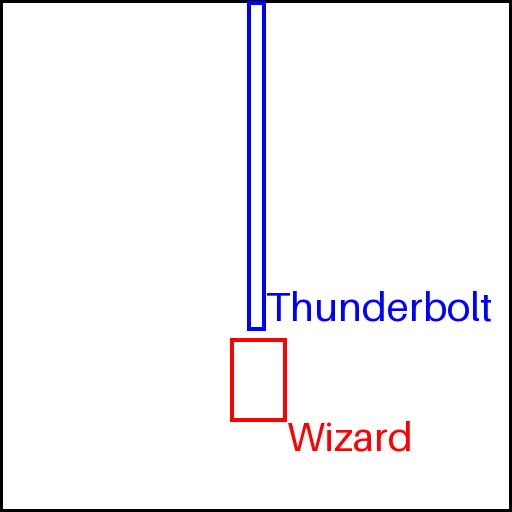}}
\end{tabular} &

\begin{tabular}{p{5cm}}
\includegraphics[width=0.9\linewidth]{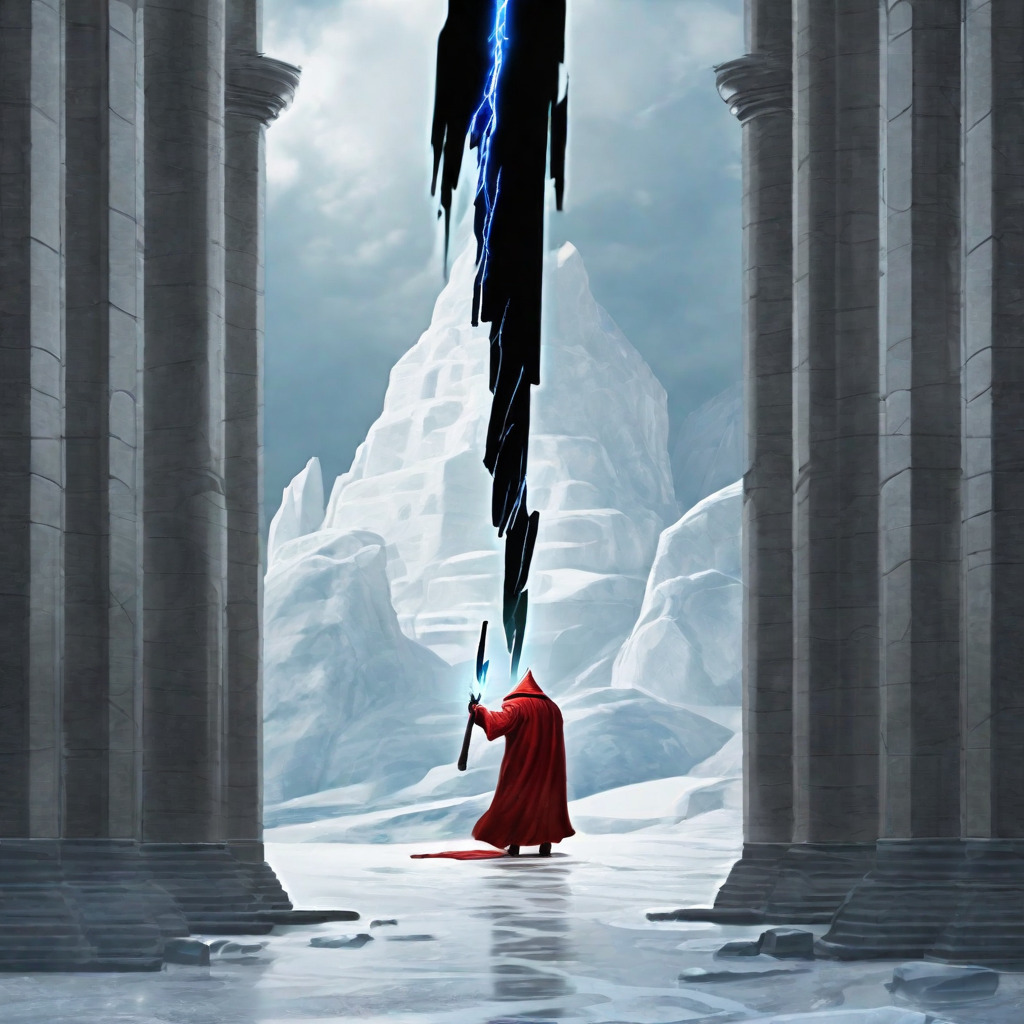}
\end{tabular} &

\begin{tabular}{p{5cm}}
\multicolumn{1}{p{3.5cm}}{\scriptsize\centering A concept art of a bright desert with \textbf{\color{blue} two magical portals} in air. One portal showing modern city with skyscrapers and another portal showing \textbf{\color{red} black and white apocalyptic city} with broken buildings} \\
\multicolumn{1}{c}{\includegraphics[width=0.15\linewidth]{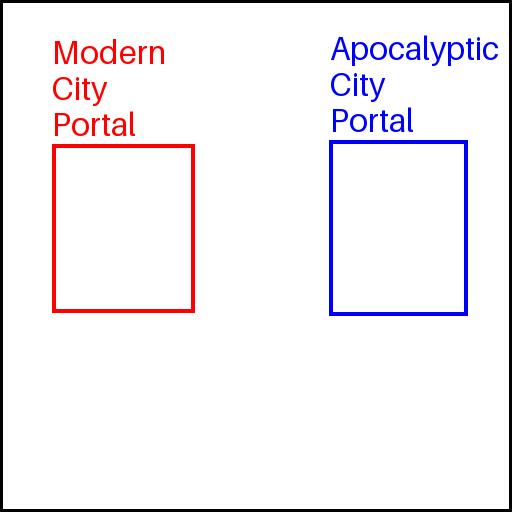}}
\end{tabular} &

\begin{tabular}{p{5cm}}
\includegraphics[width=0.9\linewidth]{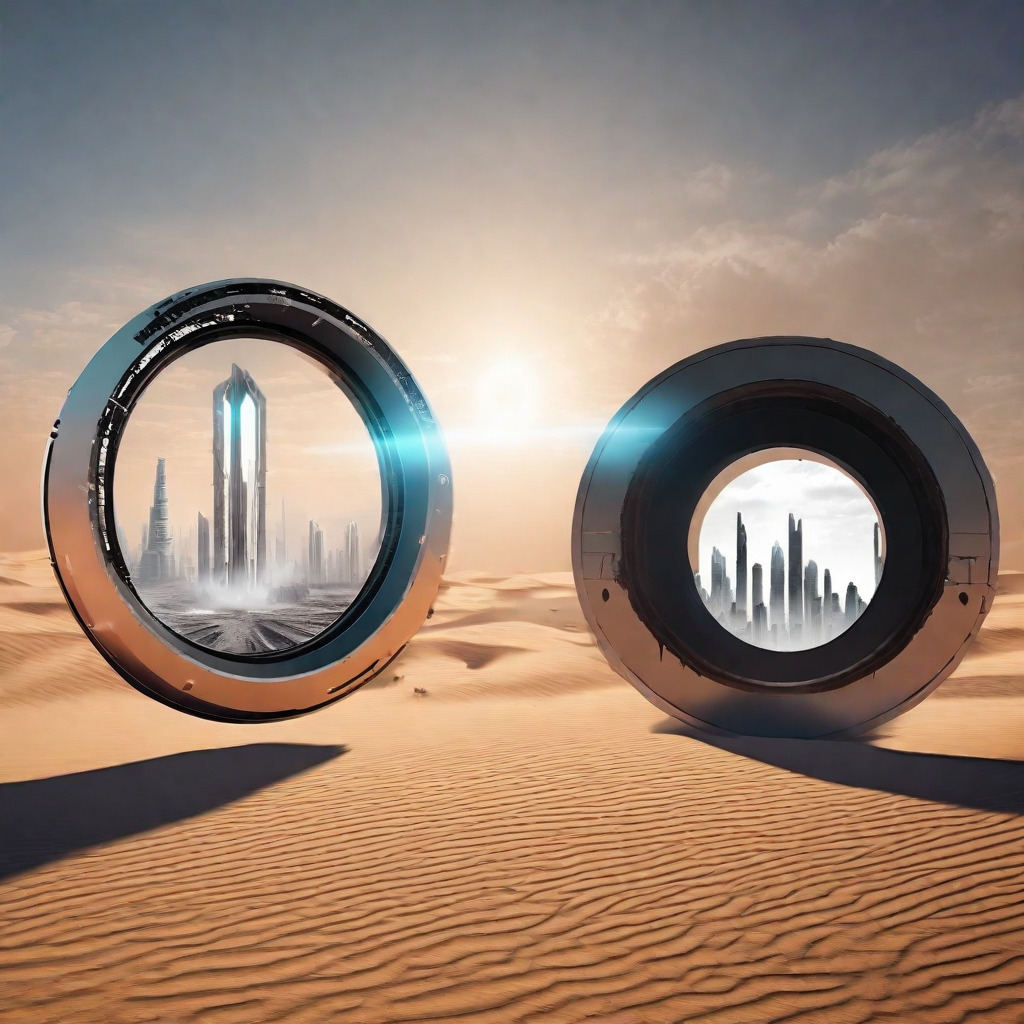}
\end{tabular}
\\
& & & \\
& & & \\

\begin{tabular}{p{5cm}}
\multicolumn{1}{p{3.5cm}}{\scriptsize\centering A concept art of a epic duel with a far away \textbf{\color{Green} green robed wizard} firing \textbf{\color{blue} magic thunderbolt} in air towards a \textbf{\color{red} wizard} protecting himself with his magical shield in an surreal icy windy landscape} \\
\multicolumn{1}{c}{\includegraphics[width=0.15\linewidth]{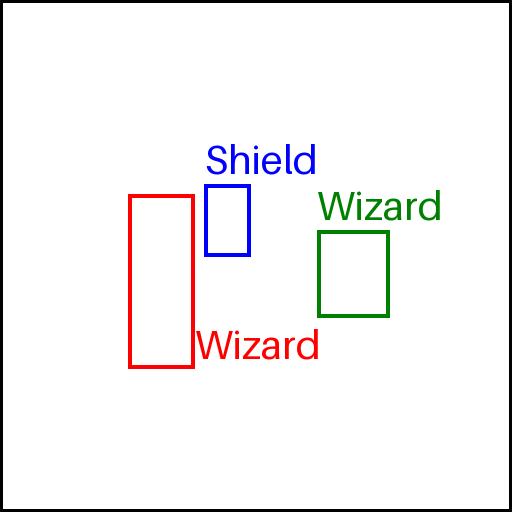}}
\end{tabular} &

\begin{tabular}{p{5cm}}
\includegraphics[width=0.9\linewidth]{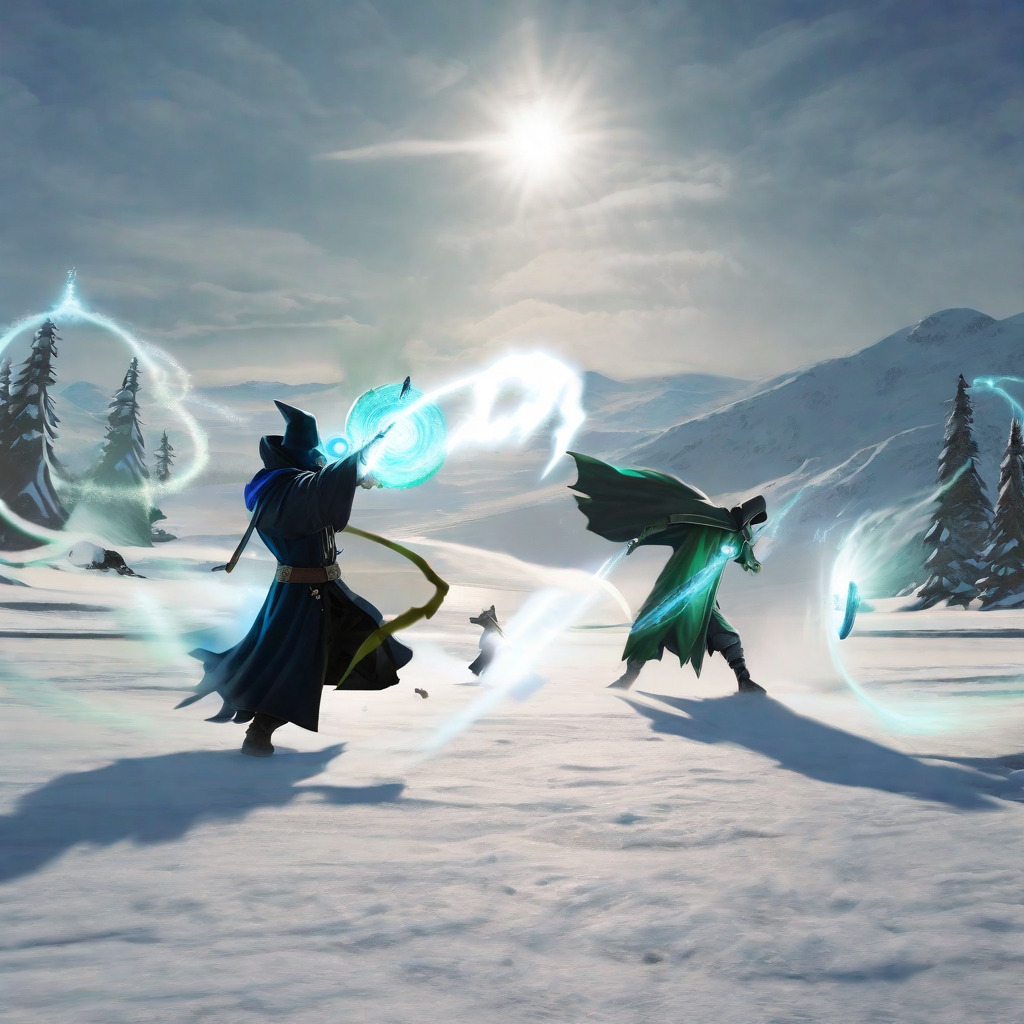}
\end{tabular} &

\begin{tabular}{p{5cm}}
\multicolumn{1}{p{4cm}}{\centering A concept art of a black and white place with a \textbf{\color{Green} green robe wizard} far away firing his wand in air for a \textbf{\color{blue} blue thunderbolt} with a sheer force} \\
\multicolumn{1}{c}{\includegraphics[width=0.15\linewidth]{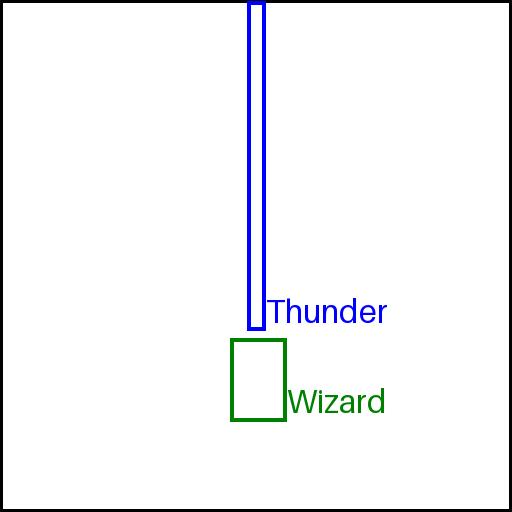}}
\end{tabular} &

\begin{tabular}{p{5cm}}
\includegraphics[width=0.9\linewidth]{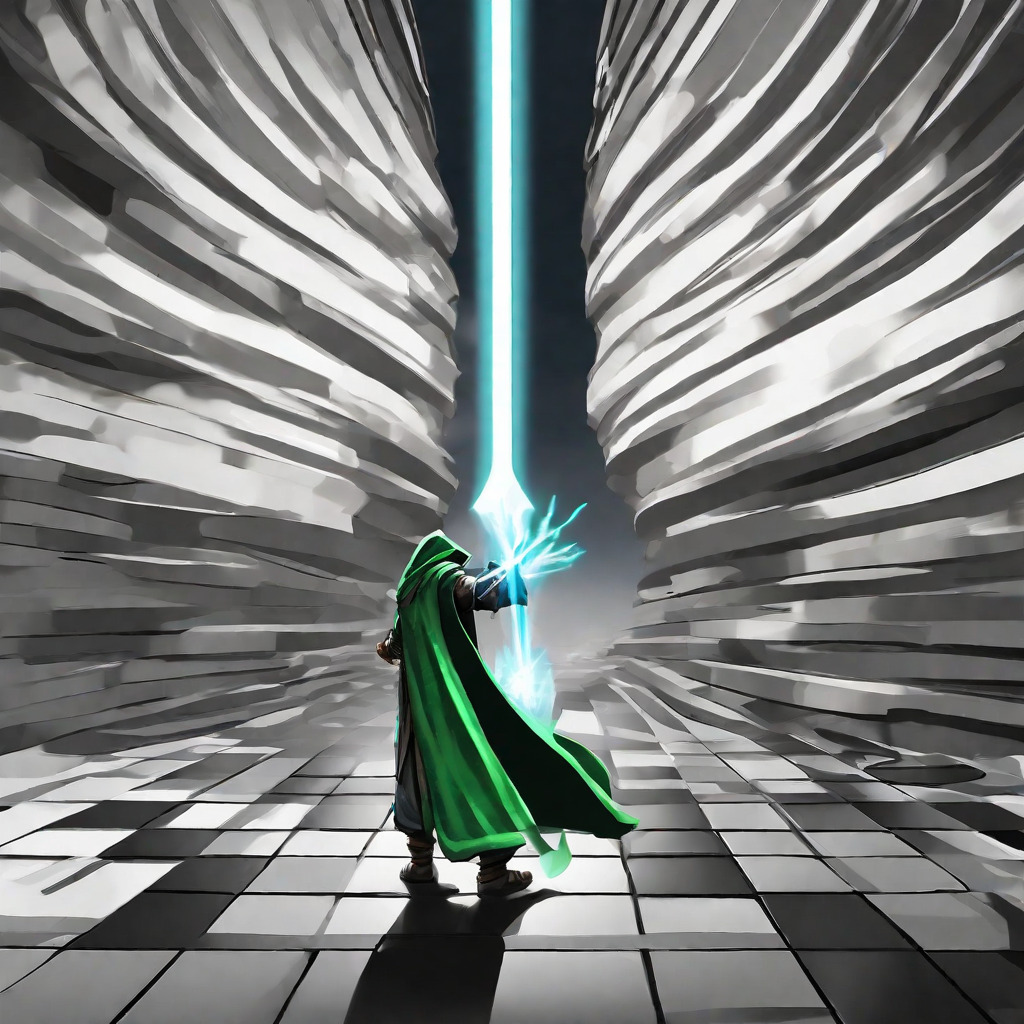}
\end{tabular}
\\
& & & \\
& & & \\

\begin{tabular}{p{5cm}}
\multicolumn{1}{p{3.5cm}}{\centering A realistic photo of a \textbf{\color{purple} purple rabbit} and a \textbf{\color{pink} pink chicken} and a \textbf{\color{yellow} yellow cat}} \\
\multicolumn{1}{c}{\includegraphics[width=0.15\linewidth]{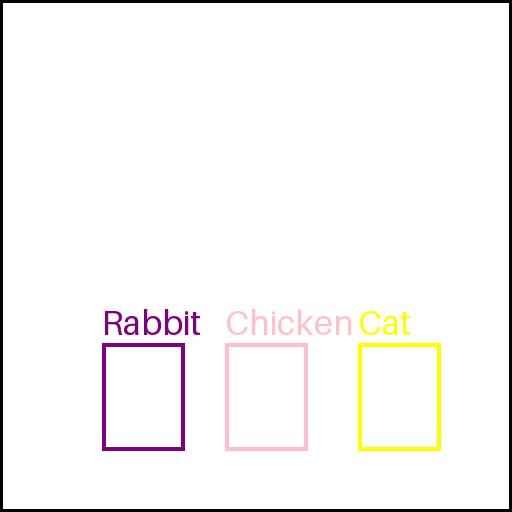}}
\end{tabular} &

\begin{tabular}{p{5cm}}
\includegraphics[width=0.9\linewidth]{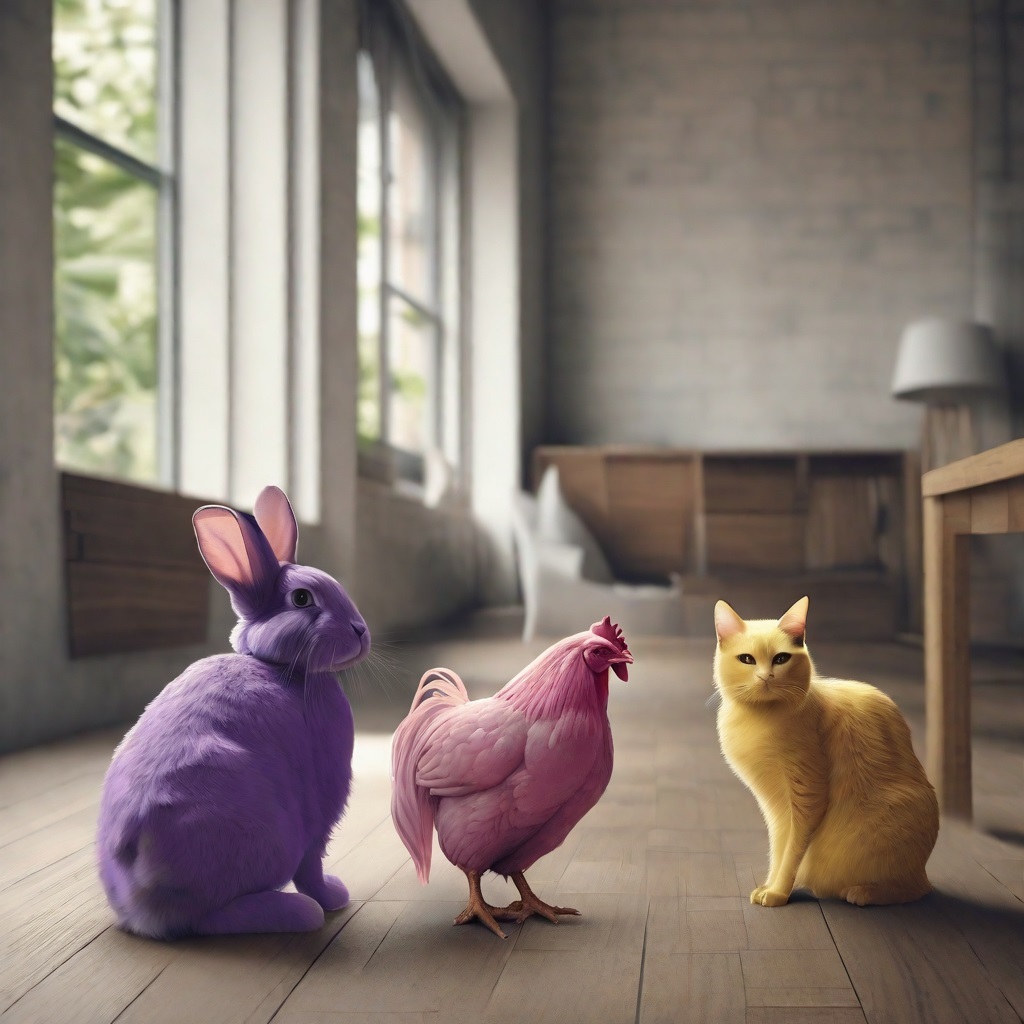}
\end{tabular} &

\begin{tabular}{p{5cm}}
\multicolumn{1}{p{3.5cm}}{\centering A realistic photo of a \textbf{\color{purple} purple cat} and a \textbf{\color{pink} pink cat} and a \textbf{\color{black} black cat}} \\
\multicolumn{1}{c}{\includegraphics[width=0.15\linewidth]{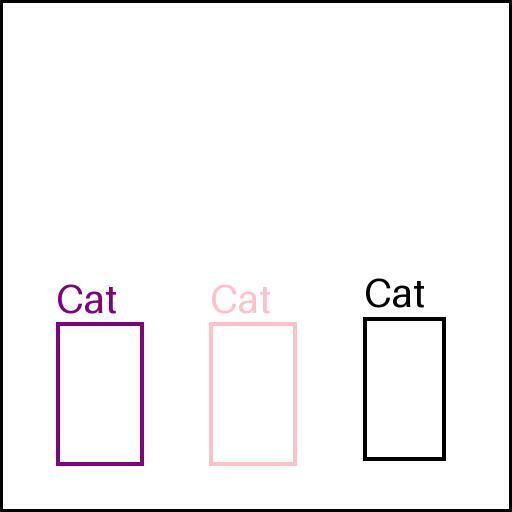}}
\end{tabular} &

\begin{tabular}{p{5cm}}
\includegraphics[width=0.9\linewidth]{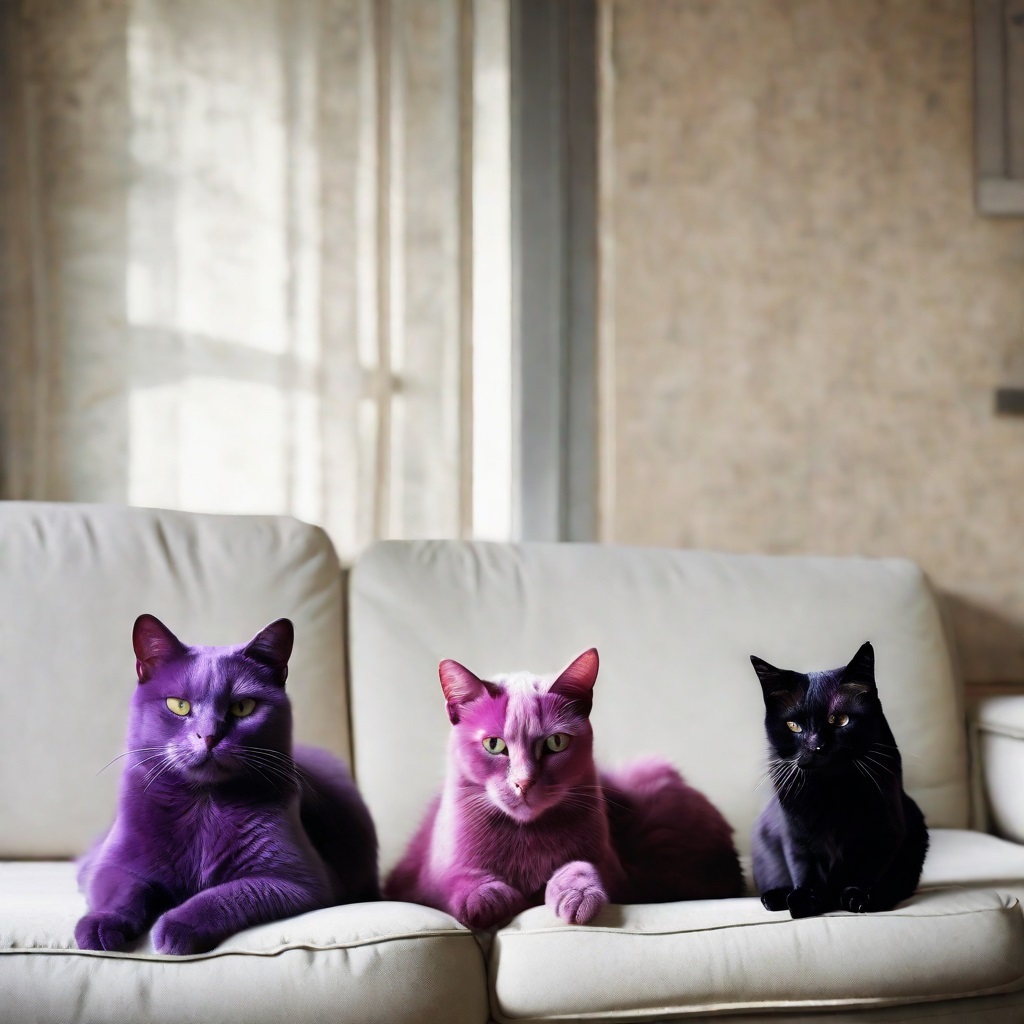}
\end{tabular}
\\

\end{tabular}
\centering
\caption{\textbf{Highlighting robustness of \modelshort{}.}
We present 8 qualitative results on diverse and challenging prompts with complex layout constraints to showcase the ability of \modelshort{} to synthesize high-quality, coherent, and spatially accurate images.}
\label{fig:misc_outputs2}
\end{figure*}